\documentclass[letterpaper]{article} 

\usepackage{styles/gintmr}
\providecolor{paperprimary}{HTML}{000000}

\usepackage{url}  
\usepackage{graphicx} 
\usepackage{natbib}  
\usepackage{caption} 

\definecolor{wxrcommentpurple}{RGB}{104, 214, 146}

\usepackage{algorithm}
\usepackage{algorithmic}

\usepackage{newfloat}
\usepackage{listings}
\DeclareCaptionStyle{ruled}{labelfont=normalfont,labelsep=colon,strut=off} 
\floatstyle{ruled}
\newfloat{listing}{tb}{lst}{}
\floatname{listing}{Listing}

\usepackage{booktabs}
\usepackage{longtable}
\usepackage{placeins}
\title{Can Image Models Imagine Time?\\
ImageTime: A Novel Benchmark for Probing Visual World Modeling Through Spatiotemporal Consistency}

\author{
    \color{paperprimary}
    Xinrui Wu\textsuperscript{\rm 1*}, Lichen Huang\textsuperscript{\rm 1*}\\
}
\affiliations{
    \color{paperprimary}
    \textsuperscript{\rm 1}University of Electronic Science and Technology of China\\
    \textsuperscript{*}Equal contribution.\\
    \textbf{Email:} xinruiwu.wxr@gmail.com, xhghlc@gmail.com
}

\newcommand{\figref}[1]{Figure~\ref{#1}}
\newcommand{\tabref}[1]{Table~\ref{#1}}

\newcommand{\FigIntroTeaser}{%
\begin{center}
    \centering
    \includegraphics[width=0.98\textwidth]{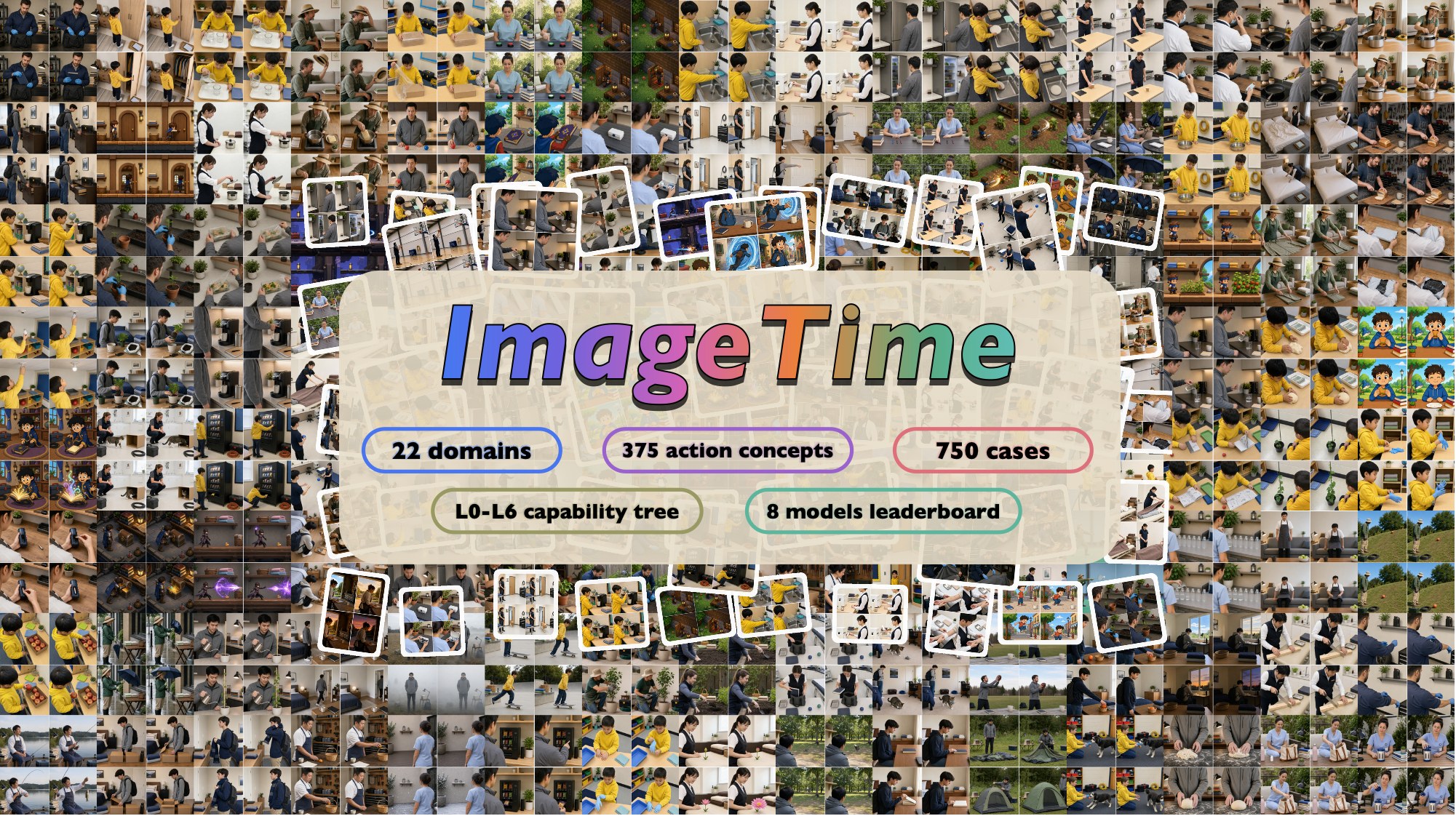}
    \captionsetup{hypcap=false}
    \captionof{figure}{Teaser overview of ImageTime. The collage summarizes the benchmark scale of 22 domains, 375 action concepts, and 750 cases, together with the L0--L6 capability tree and the eight-model prompt-only evaluation setting. A temporally coherent motion sheet must preserve identities, objects, spatial relations, and causal order across ordered key states rather than only render a plausible final frame.}
    \label{fig:intro-teaser}
\end{center}
}

\newcommand{\FigOverview}{%
\begin{figure*}[t]
    \centering
    \includegraphics[width=0.95\textwidth]{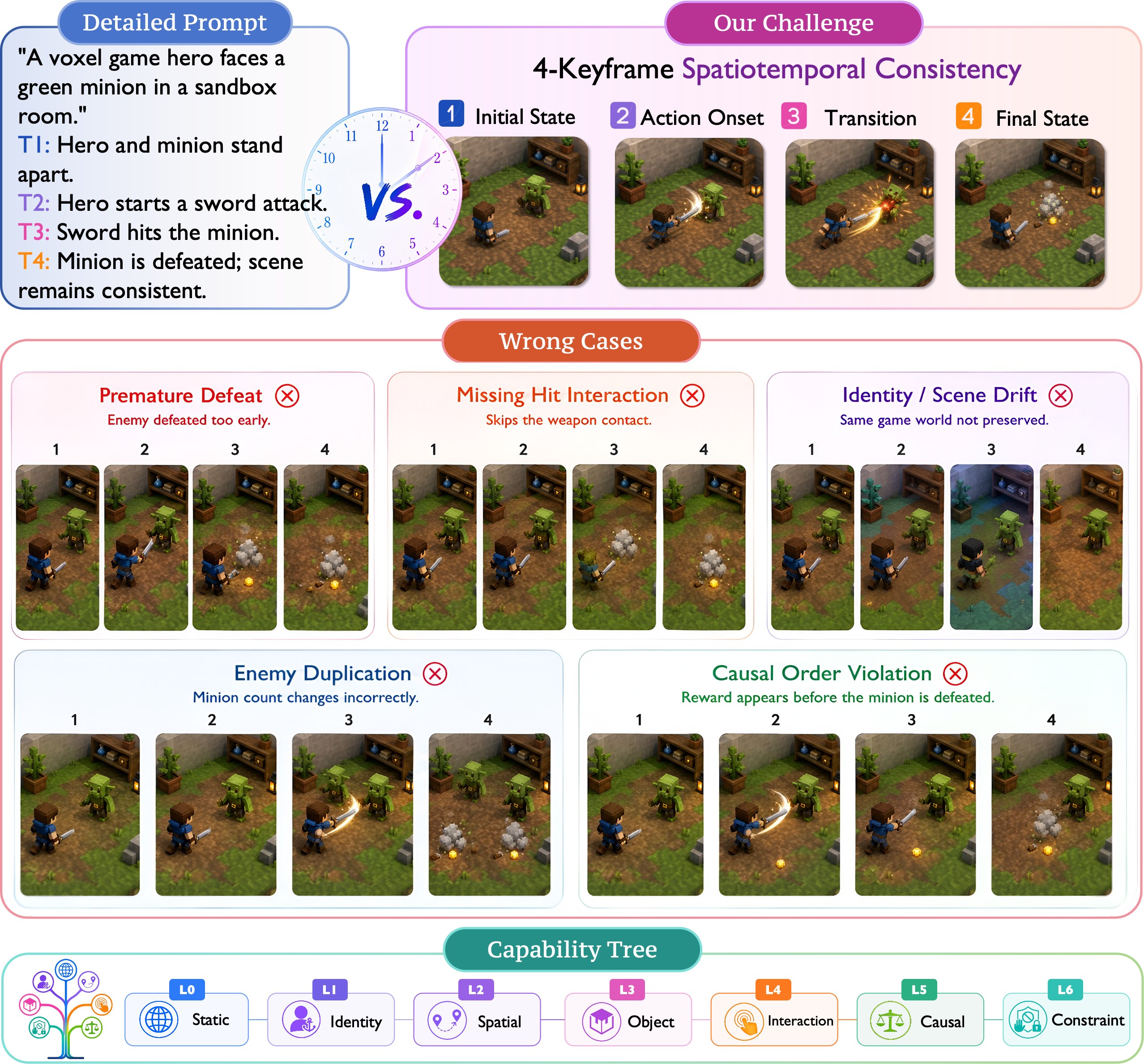}
    \caption{Overview of the ImageTime task formulation and failure taxonomy. A detailed action prompt specifies four ordered key states: initial state, action onset, transition, and final state. The figure illustrates common failure modes such as premature final states, missing interactions, identity or scene drift, object duplication, and causal-order violations, and links them to the L0--L6 capability tree.}
    \label{fig:overview}
\end{figure*}
}

\newcommand{\TabCapabilityTree}{%
\begin{table}[t]
    \centering
    \small
    \renewcommand{\arraystretch}{1.08}
    \begin{tabular}{clp{0.53\linewidth}}
        \toprule
        Level & Capability & Diagnostic focus \\
        \midrule
        0 & Static grounding & Instantiate the requested subject, scene/domain, visual style, camera/view, objects, and initial state. \\
        1 & Identity and anchors & Preserve actors, targets, non-target objects, clothing, materials, background anchors, lighting, and viewpoint across panels. \\
        2 & Spatial transition & Represent objective changes in position, containment, distance, orientation, support, or relative layout. \\
        3 & Object-state transition & Track object states, selected sets, counts, attributes, fill levels, and source--target conservation. \\
        4 & Interaction and affordance & Depict plausible contact, tool use, human/object biomechanics, garment orientation, gaze, attention, and affordance geometry. \\
        5 & Causal process & Preserve multi-stage preconditions, visible triggers, intermediate states, conservation, and causal order. \\
        6 & Constraint/counterfactual & Obey negations, blocked actions, selective operations, no-action outcomes, and counterfactual world-state conditions. \\
        \bottomrule
    \end{tabular}
    \caption{Progressive capability levels in ImageTime. Lower levels establish the stable visual world; higher levels require ordered state evolution, causal composition, and constraint-sensitive reasoning.}
    \label{tab:capability-tree}
\end{table}
}

\newcommand{\TabJudgeRubric}{%
\begin{table}[!tbp]
    \centering
    \scriptsize
    \renewcommand{\arraystretch}{0.98}
    \setlength{\tabcolsep}{3.5pt}
    \begin{tabular}{@{}p{0.055\textwidth}p{0.24\textwidth}p{0.64\textwidth}@{}}
        \toprule
        ID & Dimension & Focus \\
        \midrule
        C0 & Layout validity & Exactly four readable, self-contained panels in the expected 2$\times$2 temporal order. \\
        C1 & Reference grounding & For scaffold settings, $t_1$ preserves the reference subject, scene, style, view, layout, and state. \\
        C2 & Entity consistency & Subjects, objects, quantities, roles, orientation, materials, reflections, and target/non-target identities persist. \\
        C3 & Spatial-view consistency & Scene, relative positions, support, scale, camera, depth, lighting, shadows, reflections, and anchors remain coherent. \\
        C4 & Motion continuity & Actors or objects show task-effective displacement with plausible pose, direction, support, and path progression. \\
        C5 & Temporal ordering & Panels follow $t_1 \rightarrow t_2 \rightarrow t_3 \rightarrow t_4$ without skipped or premature states. \\
        C6 & Causal process consistency & State changes follow visible preconditions, contact, force, tool use, body mechanics, and conservation. \\
        C7 & Interaction consistency & Agents, tools, targets, recipients, garments, gaze, attention, and affordances interact plausibly. \\
        C8 & Constraint/counterfactual sensitivity & Explicit negations, selective operations, blocked actions, and counterfactual conditions are obeyed. \\
        C9 & Visual quality & The full motion sheet is clear, coherent, artifact-free enough, style-matched, and informative for judging. \\
        \bottomrule
    \end{tabular}
    \caption{Capability-score rubric C0--C9. These scores summarize the main evaluation axes used to judge whether a generated motion sheet is valid, consistent, temporally ordered, causally coherent, constraint-sensitive, and visually readable.}
    \label{tab:judge-rubric-c}
\end{table}

\begin{table}[!tbp]
    \centering
    \scriptsize
    \renewcommand{\arraystretch}{0.98}
    \setlength{\tabcolsep}{3.2pt}
    \begin{tabular}{@{}p{0.12\textwidth}p{0.05\textwidth}p{0.25\textwidth}p{0.53\textwidth}@{}}
        \toprule
        Group & ID & Dimension & Focus \\
        \midrule
        Layout & D0 & Panel parse and crop integrity & Panels are complete, separated, ordered, readable, and not stitched, merged, polluted, or cropped. \\
        Entity & D1 & Task-entity binding & Correct actor, target, recipient, tool, container, obstacle, non-target object, and final object. \\
        Entity & D2 & Object permanence and lifecycle & Objects and actors persist, transform, transfer, hide, appear, or disappear only for task-valid reasons. \\
        Entity & D3 & Background anchor persistence & Stable anchors, surface textures, shadows, reflections, windows, roads, signs, and background objects remain consistent. \\
        Spatial & D4 & Spatial coordinate and relation consistency & Left/right, front/back, inside/outside, support, contact, containment, distance, and anchor-relative positions are coherent. \\
        Spatial & D5 & Camera, scale, and depth continuity & Viewpoint, crop, scale, perspective, lighting, shadows, reflections, and depth stay stable unless justified. \\
        Motion & D6 & Trajectory and pose-position coupling & Pose, position, facing direction, stride phase, support, and displacement change together. \\
        Motion & D7 & Contact, support, and affordance geometry & Hand-object, tool-target, foot-ground, grip, pouring, insertion, hinge, sleeve, collar, handle, and support geometry is plausible. \\
        Motion & D8 & State-transition visibility & Important state changes and source/target variables are visibly decomposed, not only shown as start/end. \\
        Temporal & D9 & Phase distinctiveness & $t_1$--$t_4$ are meaningfully different ordered phases, not repeats, early final states, or unrelated snapshots. \\
        Interaction & D10 & Attention, gaze, and intent alignment & Gaze, head orientation, pointing, reaching, body facing, and attention align with the active target. \\
        Attribute & D11 & Quantity, set membership, and attribute binding & Counts, fill levels, source/target volumes, selected objects, groups, and color-object bindings are correct when applicable. \\
        Occlusion & D12 & Occlusion and reappearance consistency & Hidden, inside, behind, or reappearing objects remain consistent when occlusion is task-relevant. \\
        Constraint & D13 & Constraint execution & Explicit prohibitions, blocked actions, no-action outcomes, selective operations, and counterfactual conditions hold. \\
        Readability & D14 & Visual readability for judging & Small objects, hands, feet, contact points, faces, tools, openings, reflections, and panel boundaries are clear enough. \\
        \bottomrule
    \end{tabular}
    \caption{Diagnostic-subscore rubric D0--D14. These subscores provide fine-grained visual evidence for why each capability succeeds or fails, including entity binding, spatial anchors, motion evidence, temporal phases, interaction geometry, quantity or occlusion consistency, constraint execution, and readability.}
    \label{tab:judge-rubric-d}
\end{table}
}

\newcommand{\FigAllModelsCapabilityRadar}{%
\begin{figure}[!t]
    \centering
    \includegraphics[width=0.56\textwidth]{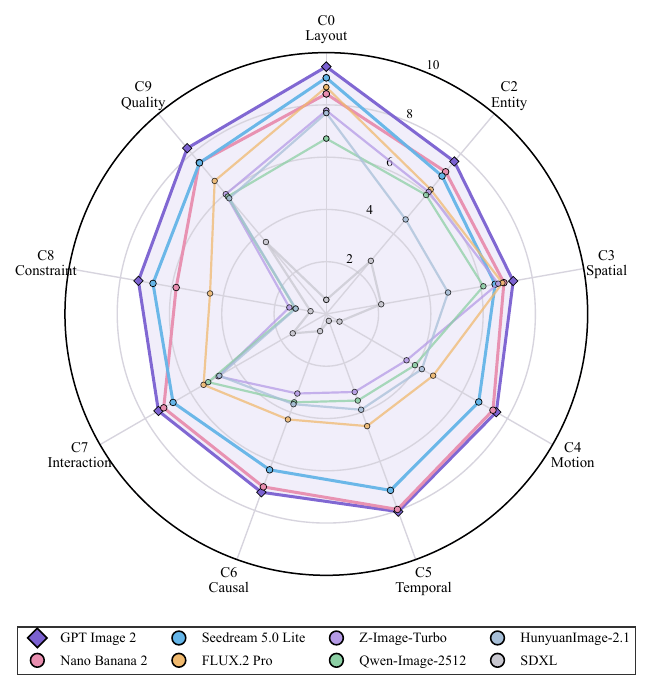}
    \caption{Radar view of prompt-only capability scores across all eight evaluated models. The axes correspond to C0 and C2--C9; C1 is omitted because reference grounding is not applicable without a reference image. The faint purple envelope highlights the GPT Image 2 capability profile, while all model contours and markers are retained for comparison.}
    \label{fig:all-models-capability-radar}
\end{figure}
}

\newcommand{\FigDiagnosticHeatmap}{%
\begin{figure*}[!t]
    \centering
    \includegraphics[width=0.92\textwidth]{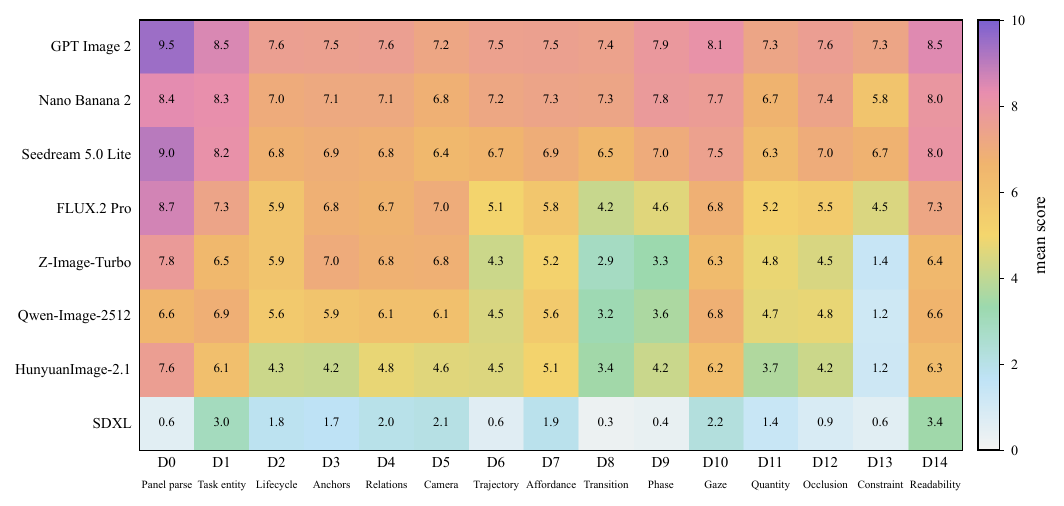}
    \caption{Fine-grained diagnostic decomposition across D0--D14. Process-sensitive diagnostics include trajectory--pose coupling (D6), state-transition visibility (D8), phase distinctiveness (D9), quantity and attribute binding (D11), occlusion and reappearance consistency (D12), and constraint execution (D13).}
    \label{fig:d-diagnostic-heatmap}
\end{figure*}
}

\newcommand{\FigScoreFamilyAndCost}{%
\begin{figure}[!t]
    \centering
    \begin{minipage}[t]{0.48\textwidth}
        \centering
        \includegraphics[width=\linewidth]{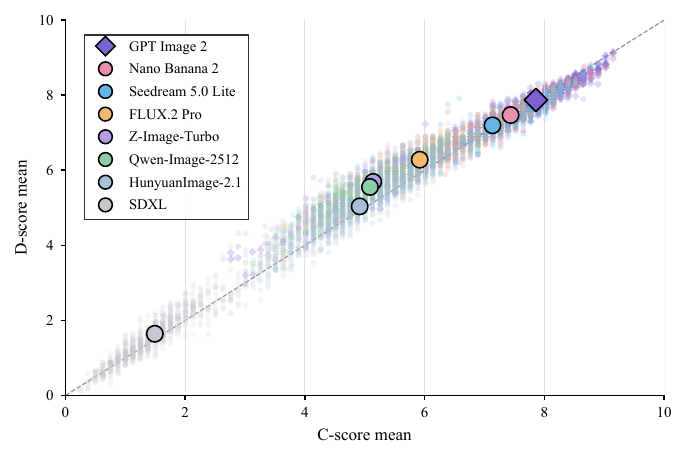}
        \caption{Capability-score mean versus diagnostic-score mean. Faint points are generated motion sheets, large markers are model means, and the diagonal indicates equal C and D averages.}
        \label{fig:quality-vs-core}
    \end{minipage}
    \hfill
    \begin{minipage}[t]{0.48\textwidth}
        \centering
        \includegraphics[width=\linewidth]{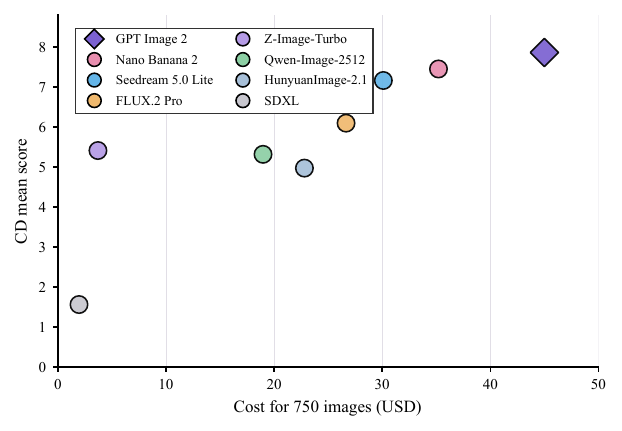}
        \caption{Cost-performance trade-off using the reported cost of 750 prompt-only generations. The vertical axis uses Overall mean.}
        \label{fig:cost-performance}
    \end{minipage}
\end{figure}
}

\newcommand{\FigDomainHeatmap}{%
\begin{figure*}[!t]
    \centering
    \includegraphics[width=0.94\textwidth,height=0.59\textheight,keepaspectratio,trim={0 72bp 0 0},clip]{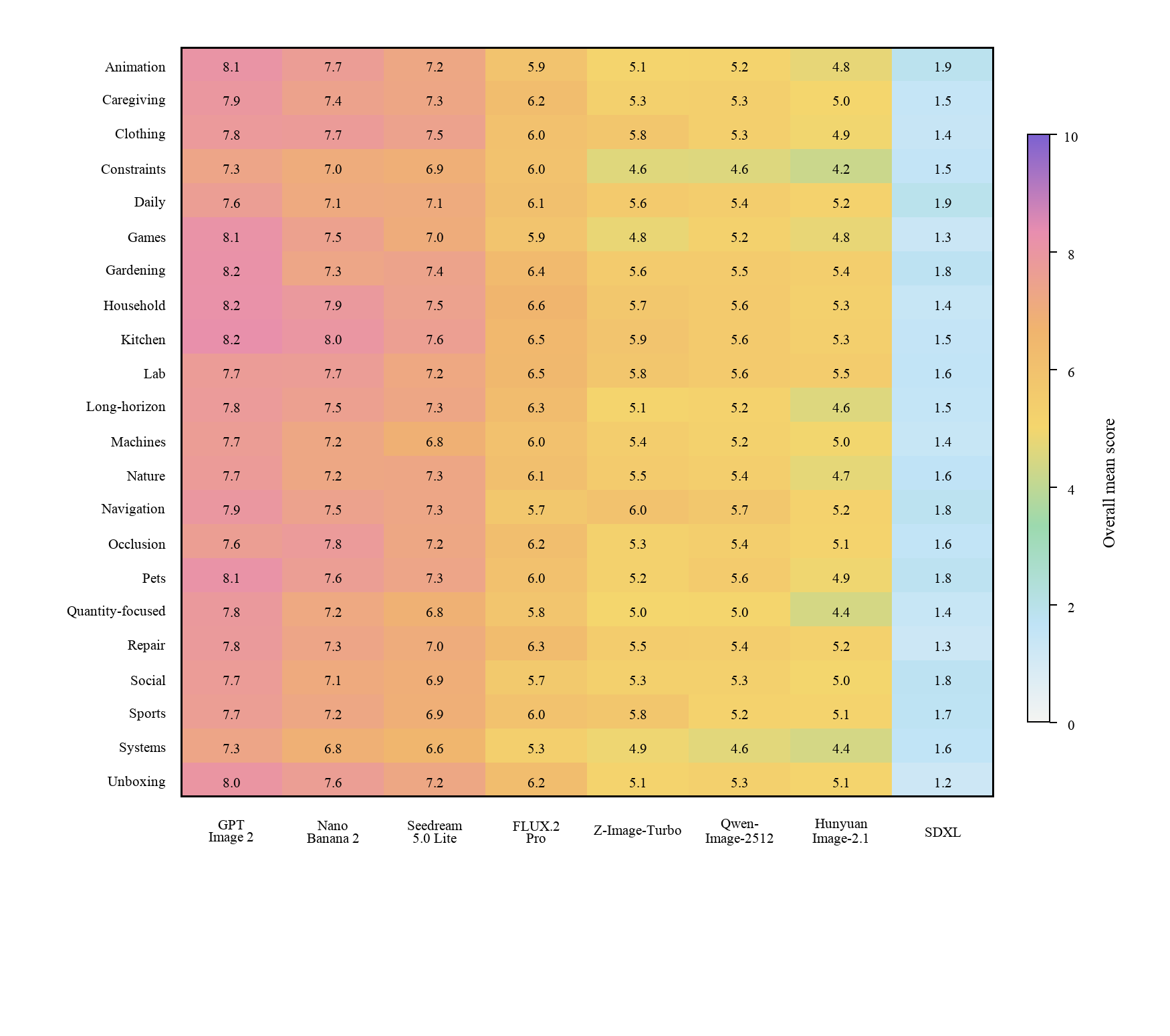}
    \caption{Fine-grained category-level prompt-only Overall mean scores. Rows cover all 22 ImageTime categories reconstructed from the case filename and case-id prefixes, and columns report the eight evaluated models.}
    \label{fig:domain-model-heatmap}
\end{figure*}
}

\newcommand{\FigTreeLevelCurve}{%
\begin{figure*}[!t]
    \centering
    \includegraphics[width=0.84\textwidth]{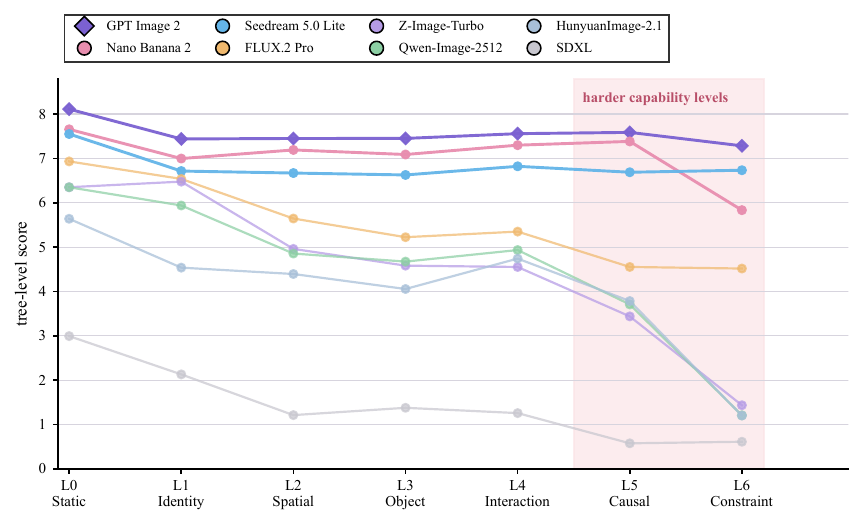}
    \caption{Tree-level scores across the Progressive Capability Tree. Lower-performing models show sharper degradation at higher capability levels, especially causal process composition and constraint/counterfactual reasoning, while stronger models remain comparatively stable across L0--L6.}
    \label{fig:tree-level-curve}
\end{figure*}
}

\begin{document}

\maketitle

\begin{abstract}

Image generation models now produce high-quality static images, yet their ability to represent how a visual world changes over time remains poorly understood. Practical workflows such as storyboarding, step-by-step illustration, reference-guided editing, and video previsualization require models to preserve identities, objects, spatial relations, and causal order across multiple visual states. Existing evaluations largely measure single-image correctness, compositional alignment, or video quality, leaving open whether an image model can coherently imagine a temporally ordered process. We introduce ImageTime, a diagnostic benchmark that uses spatiotemporal consistency as a behavioral probe of visual world modeling in image generation. Given an action instruction, and optionally a reference image specifying the initial state, a model must generate one image containing four ordered key states: initial state, action onset, transition state, and final state. This four-keyframe protocol is more temporally demanding than single-image generation while avoiding the confounds of dense video dynamics. ImageTime organizes tasks with a progressive capability hierarchy and decomposes each scenario into stage-wise state predicates, cross-frame temporal constraints, and forbidden causal violations. GPT-5.5 scores all generated images under a structured VLM-as-judge protocol, producing interpretable capability scores, diagnostic subscores, and failure labels. Through multi-family benchmarking, ImageTime reveals where current image generation systems succeed, fail, and drift when asked to maintain coherent visual world states over time.
\end{abstract}


\FigIntroTeaser

\section{Introduction}

Recent advances in artificial intelligence have increasingly suggested that visual perception is not merely a tool for surface-level pixel synthesis, but a foundational component of higher-level cognition. On the trajectory toward more general and embodied forms of intelligence, world-aware generation has emerged as an important research direction \cite{brooks2024video,duan2025worldscore,li2026worldmodelbench,yue2025simulating}: a generative model should not only render realistic visual content, but also exhibit an implicit understanding of the physical world it depicts \cite{li2026worldmodelbench,bansal2025videophy,kang2024far}. For image foundation models, this raises a fundamental question beyond static appearance modeling: can they capture the spatial, temporal, and causal regularities that underlie a coherent visual world \cite{li2026worldmodelbench,duan2025worldscore,feng2024tc,bansal2025videophy}? In this sense, the ability to maintain spatiotemporal consistency is not a peripheral visual skill, but a basic prerequisite for moving image generation models toward world-aware visual generation.

In this paper, we use \emph{visual world modeling} in an operational sense: the ability of a generation model to maintain the latent variables of a depicted scene, including entity identity, object state, spatial layout, interaction geometry, causal preconditions, and constraints, while the scene evolves under an instructed action. These variables are difficult to assess from a single image because a single snapshot can look plausible even when the underlying process is impossible. Ordered visual states make them inspectable: object identity must persist, spatial anchors must remain stable, state changes must occur only after their causes, and constrained or counterfactual conditions must prevent invalid outcomes. We therefore treat spatiotemporal consistency as a behavioral probe of visual world modeling. ImageTime does not claim to directly inspect a model's internal world model; instead, it probes whether image models externalize world-model-like behavior when asked to maintain one coherent visual world through time.

Recent image foundation models have achieved remarkable progress in static image synthesis \cite{rombach2022high,saharia2022photorealistic,podell2024sdxl,esser2024scaling,betker2023improving}, producing visually realistic and semantically aligned images that rival human-created visual assets in many scenarios. However, the physical world is not a collection of isolated frozen snapshots, but a continuous process of state changes. As image generation models are increasingly used in higher-level visual creation workflows, including multi-frame storyboarding, step-by-step visual illustration, reference-guided editing, and video previsualization \cite{dinkevich2025story2board,tian2025envision,lin2023videodirectorgpt,cheng2024autostudio,chen2025multiref}, it becomes necessary to examine whether these models can represent not only what a scene looks like, but also how it changes over time. This motivates a shift from static synthesis to process-aware image generation \cite{tian2025envision,feng2024tc}, where the key challenge is to preserve the same world across multiple ordered visual states.

Nevertheless, a substantial gap remains between this emerging requirement and current evaluation protocols. Most existing text-to-image benchmarks primarily evaluate single-state correctness \cite{ghosh2023geneval,huang2023t2i,hu2023tifa,cho2024davidsonian,meng2024phybench,wiles2025revisiting}, focusing on whether a generated image contains the correct objects, attributes, counts, spatial relations, and overall semantics \cite{ghosh2023geneval,huang2023t2i,hu2023tifa}. While such evaluation is effective for measuring static alignment, it can be misleading when the task involves state changes. A model may generate a plausible final image while skipping necessary intermediate actions, duplicating objects, drifting the scene identity, or violating causal order \cite{tian2025envision,feng2024tc,li2026worldmodelbench,huang2024vbench,liu2024evalcrafter}. For instance, in an action such as opening a drawer and taking out an object, the object should not appear in the hand before the drawer is opened, and the hand should reach into the drawer before the object is removed. These failures cannot be reliably diagnosed by judging a single image in isolation \cite{tian2025envision,feng2024tc,huang2024vbench,liu2024evalcrafter}. Therefore, evaluating image generation models under ordered multi-state settings is necessary for revealing whether they truly preserve a coherent visual world over time.

To bridge this critical evaluation gap, we introduce ImageTime, a diagnostic benchmark for evaluating spatiotemporal consistency in image generation. Rather than treating multi-panel generation as an end task \cite{tian2025envision,dinkevich2025story2board}, ImageTime uses it as a controlled evaluation interface for probing whether image generation models can maintain coherent world states under change. Specifically, given an action instruction, and optionally a reference image that specifies the initial world state \cite{gui2026image,schwarz2025recipe,duan2025worldscore}, a model is asked to generate a single image containing four temporally ordered key states: the initial state, action onset, transition state, and final state. ImageTime contains 750 benchmark cases spanning 22 domains and 375 action concepts, providing a broad evaluation standard for process-aware image generation. The main experiments in this paper use the strict prompt-only setting, while the same protocol also supports reference-conditioned edit models through a 2$\times$2 scaffold whose top-left panel contains a generated reference image. To support fine-grained and interpretable evaluation, we design a seven-level Capability Tree from L0 to L6, covering static grounding, identity and anchor preservation, spatial transition, object-state transition, interaction and affordance, causal process, and constraint/counterfactual reasoning. We further decompose each task into stage-wise state predicates, cross-frame temporal constraints, and forbidden causal violations \cite{hu2023tifa,li2026worldmodelbench,cho2024davidsonian}, and propose a structured VLM-as-judge evaluation protocol with two complementary score families: capability scores C0--C9 summarize the main evaluation axes, while diagnostic subscores D0--D14 localize the visual evidence behind each success or failure. In our reported experiments, GPT-5.5 scores all images generated by all evaluated models under this structured rubric. These scores assess dimensions such as entity persistence, spatial-view stability, motion continuity, temporal ordering, causal preconditions, interaction geometry, material and reflection stability, state conservation, biomechanical plausibility, and constraint following \cite{zhang2023gpt,wang2025lmm4lmm,wu2024gpt,sani2026imagenworld}. We apply this L0--L6 Capability Tree to evaluate multiple image generation models and use GPT-5.5-assisted multi-round quality checks to refine prompts, case specifications, and the GPT-5.5 judge rubric before final scoring. In this way, ImageTime provides a controlled probe for diagnosing whether current image generation models can move from single-state correctness toward coherent multi-state visual reasoning.

In summary, our contributions are threefold:

\begin{itemize}
    \item \textbf{A Behavioral Probe and Benchmark for Visual World Modeling.} We introduce ImageTime, a diagnostic benchmark that uses spatiotemporal consistency across ordered key states as an observable probe of whether image generation models can maintain and evolve a coherent visual world. ImageTime contributes a concrete evaluation standard with 22 domains, 375 action concepts, and 750 total cases, extending image generation evaluation from conventional single-state correctness to ordered multi-state consistency.
    \item \textbf{A Seven-Level Capability Tree \& Structured Evaluation.} We design an L0--L6 Capability Tree with seven progressive levels covering static grounding, identity and anchor preservation, spatial transition, object-state transition, interaction and affordance, causal process, and constraint/counterfactual reasoning. Based on this formulation, we use GPT-5.5 to score all generated images with a structured VLM-as-judge protocol whose C0--C9 capability scores and D0--D14 diagnostic subscores are explicitly mapped back to the hierarchy, enabling both fine-grained failure localization and tree-level analysis of progressive capability degradation.
    \item \textbf{Systematic Multi-Model Benchmarking \& Diagnostic Insights.} We apply the 750-case ImageTime benchmark and the L0--L6 Capability Tree to evaluate multiple families of image generation models. Our analysis not only quantifies their performance gaps across the seven-level hierarchy, but also reveals recurring failure modes such as identity drift, scene drift, missing intermediate states, premature final states, object duplication, impossible contact, and causal-precondition violations, providing diagnostic insights into the current limitations of image generation models in preserving coherent visual world states over time.
\end{itemize}

\FigOverview
\section{Related Work}

\subsection{Text-to-Image and Structured Evaluation}
Text-to-image evaluation has moved from global image quality and prompt alignment toward fine-grained diagnosis of compositionality, counting, spatial relations, physical commonsense, reasoning, and human preference. DrawBench and PartiPrompts exposed failures on complex prompts, text rendering, compositional scenes, and world knowledge \cite{saharia2022photorealistic,yu2022scaling}; GenEval, T2I-CompBench, GenAI-Bench, and HEIM further evaluate object-centric alignment, attribute binding, relations, numeracy, real-world compositionality, and holistic model behavior \cite{ghosh2023geneval,huang2023t2i,li2024evaluating,lee2023holistic}; and recent benchmarks such as PhyBench, Gecko, and OneIG-Bench expand evaluation toward physical commonsense, metric reliability, human ratings, text rendering, reasoning, style, and diversity \cite{meng2024phybench,wiles2025revisiting,chang2026oneig}. In parallel, structured and human-aligned metrics have become increasingly common: TIFA and DSG use question-answering and dependency structures for interpretable image-text faithfulness \cite{hu2023tifa,cho2024davidsonian}; VIEScore and VQAScore use multimodal models for explainable or VQA-style scoring \cite{ku2024viescore,lin2024evaluating}; ImageReward, Pick-a-Pic, and Gecko study human preference alignment \cite{xu2023imagereward,kirstain2023pick,wiles2025revisiting}; and LMM4LMM and ImagenWorld use multimodal judges and human annotations for broader diagnostic evaluation \cite{wang2025lmm4lmm,sani2026imagenworld}.

These works provide strong foundations for static image evaluation and structured judging. However, most are still organized around whether a single generated image satisfies a prompt, rather than whether multiple ordered visual states preserve identity, temporal order, and causal preconditions.

\subsection{Reference-conditioned and Personalized Image Generation}
Reference-conditioned generation is now a central interface for controllable visual creation. Subject-driven and personalized methods such as DreamBooth and DreamTuner adapt target subjects from one or a few references \cite{ruiz2023dreambooth,hua2023dreamtuner}; BLIP-Diffusion, IP-Adapter, and ControlNet improve subject representation, image-prompt conditioning, and structural control \cite{li2023blip,ye2023ip,zhang2023adding}; and MultiRef, DreamBench++, and DSH-Bench extend evaluation to multiple references, human-aligned personalization, and difficulty-aware subject taxonomies \cite{chen2025multiref,peng2025dreambench++,hu2026dsh}. These works improve reference fidelity and controllability, but their evaluations usually focus on target-image quality, subject preservation, or editing accuracy rather than multi-stage transformations of the same reference world.

Recent frontier image-generation systems have also moved toward more general multimodal creation interfaces, where text-to-image synthesis, image editing, reference conditioning, prompt following, text rendering, world knowledge, and low-latency deployment are increasingly integrated into a single user-facing model family. Representative systems include FLUX.2 Pro~\cite{blackforestlabs2025flux2}, which emphasizes multi-reference consistency, structured prompt following, typography, and high-resolution editing; GPT Image 2~\cite{openai2026gptimage2,openai2026chatgptimages2}, which improves instruction following, world knowledge, dense text generation, and thinking-mode tool use; Nano Banana 2~\cite{googledeepmind2026nanobanana2}, released in the Gemini Flash Image family, which targets fast image generation and editing with real-world knowledge, web grounding, and precise text rendering; Seedream 5.0 Lite~\cite{bytedance2026seedream50lite}, a unified multimodal image generation model with deep thinking and online search capabilities; Qwen-Image-2512~\cite{qwen2026qwenimage2512,wu2025qwenimagetechnicalreport}, a Qwen-family open-weight text-to-image model released as an update to Qwen-Image with stronger realism, detail rendering, and text rendering; HunyuanImage-2.1~\cite{tencent2025hunyuanimage21}, an efficient high-resolution 2K text-to-image model with prompt enhancement, distillation, and open released weights; and Z-Image-Turbo~\cite{cai2025z}, a distilled 6B-class image model designed for few-step, low-latency generation while preserving competitive photorealism and bilingual text rendering. These deployed systems motivate evaluation beyond classical prompt-image alignment, because their advertised capabilities increasingly involve reference use, reasoning, editing, grounding, and multi-step visual intent following.

Alongside these recent systems, Stable Diffusion XL (SDXL)~\cite{podell2024sdxl} remains an important open-weight diffusion baseline for high-resolution image synthesis. Including SDXL~\cite{podell2024sdxl} helps separate failures of general diffusion-based image synthesis from the additional capabilities claimed by more recent multimodal or proprietary image-generation systems.

\subsection{Multi-image, Storyboard, and Visual Process Generation}
Beyond single images, many visual creation tasks require coherence across multiple frames, panels, or interaction turns. VideoDirectorGPT uses language-model planning for consistent multi-scene video generation \cite{lin2023videodirectorgpt}, AutoStudio studies multi-turn interactive image generation with consistent subjects \cite{cheng2024autostudio}, and ConsiStory, StoryDiffusion, StoryMaker, and Infinite-Story preserve character identity, appearance, and style across long-range image sequences \cite{tewel2024training,zhou2024storydiffusion,zhou2024storymaker,park2026infinite}. Story2Board targets expressive storyboard generation with attention to spatial composition, background evolution, and narrative pacing \cite{dinkevich2025story2board}, while VinaBench evaluates faithful and consistent visual narratives through commonsense and discourse constraints \cite{gao2025vinabench}. These works highlight cross-image consistency, but they often emphasize narrative coherence, subject identity, or generation quality rather than predicate-level diagnosis of state transitions.

\subsection{Reasoning-centric and World-aware Image Generation}
Recent benchmarks increasingly ask whether image generation models can use reasoning and world knowledge rather than only render surface-level prompt content. R2I-Bench and GIR-Bench evaluate reasoning-driven image generation under implicit or inferred constraints \cite{chen2025r2i,li2025gir}; Envision studies unified understanding and generation for causal world process insights and chained text-to-multi-image generation \cite{tian2025envision}; ImagenWorld stress-tests open-ended real-world image generation and editing with explainable human annotations \cite{sani2026imagenworld}; and single-image-to-world studies reflect broader interest in generating interactive or 3D worlds from static inputs \cite{gui2026image,schwarz2025recipe,yue2025simulating}. These studies move image generation toward reasoning and world awareness, but reference-conditioned visual state transitions remain a more specific and less directly studied evaluation setting.

\subsection{Video Generation and World-model Evaluation}
Video benchmarks evaluate temporal behavior in continuous videos, including visual quality, motion smoothness, flickering, subject consistency, dynamics, physical plausibility, and text-video alignment. EvalCrafter, VBench, and FETV provide multi-dimensional or fine-grained evaluation for large video generation models \cite{liu2024evalcrafter,huang2024vbench,liu2023fetv}; T2V-CompBench and TC-Bench focus on compositional and temporal-compositional video generation \cite{sun2025t2v,feng2024tc}; and VideoPhy, WorldModelBench, WorldScore, VBench-2.0, T2VWorldBench, VideoVerse, WorldBench, and state-evolution benchmarks examine physical commonsense, instruction following, world knowledge, intrinsic faithfulness, physical disambiguation, and object-state persistence \cite{bansal2025videophy,li2026worldmodelbench,duan2025worldscore,zheng2025vbench,chen2026t2vworldbench,wang2025videoverse,upadhyay2026worldbench,ma2026out}. Earlier visual reasoning and intuitive physics benchmarks such as CLEVRER, IntPhys, and Physion provide conceptual foundations for evaluating temporal causality, object permanence, and physical prediction from vision \cite{yi2019clevrer,riochet2018intphys,bear2021physion}. These benchmarks provide important tools for temporal evaluation, but they primarily target continuous video synthesis rather than discrete key-state consistency in image generation.

\section{ImageTime Benchmark}

\subsection{Task Definition}
ImageTime evaluates whether an image generation model can depict a coherent visual world as it changes over time. Given an action instruction, and optionally a reference image specifying the initial state, the model must generate a single square image containing four temporally ordered keyframes in a 2$\times$2 layout: $t_1$ initial state, $t_2$ action onset, $t_3$ intermediate transition or interaction, and $t_4$ final state. The output is therefore a motion sheet, not a free-form collage: the same subject, objects, scene, camera/viewpoint, spatial relations, and causal order must remain inspectable across panels.

ImageTime focuses on spatiotemporal consistency rather than video realism. The four-keyframe interface is more demanding than single-image generation because the model must maintain one world over multiple ordered states, but it avoids dense-video confounds such as interpolation, optical-flow smoothness, and continuous camera motion. A successful output should preserve object identity and background anchors, show task-effective position or state changes, respect source--target conservation, maintain plausible interactions, and avoid impossible or premature final states. For example, if a character pours milk into a glass, the glass should become fuller only after a visible pouring action and the source bottle should contain less milk.

\figref{fig:overview} illustrates the ImageTime task interface: a detailed action prompt specifies four ordered key states, while the benchmark diagnoses common failures such as premature final states, missing interactions, identity or scene drift, object duplication, and causal-order violations under the L0--L6 capability tree.

\subsection{Progressive Capability Tree}
ImageTime organizes tasks with a Progressive Capability Tree rather than a flat content taxonomy. Higher-level tasks depend on lower-level visual commitments: a model cannot reliably represent causal pouring if it fails to preserve the bottle and glass, and it cannot reason about a blocked action if it cannot maintain the door, actor, and attempted interaction. The tree therefore orders capabilities from basic world-state grounding to increasingly structured temporal, causal, and counterfactual reasoning.

The tree contains seven levels. \emph{Static grounding} requires the requested subject, scene, style, camera/view, initial objects, and initial state; in scaffold-conditioned settings, $t_1$ must also preserve the reference image. \emph{Cross-frame identity and anchor consistency} requires actors, target objects, non-target objects, materials, background anchors, lighting, framing, and spatial layout to remain stable across $t_1$--$t_4$. \emph{Spatial and object-state transition} requires visible changes in position, containment, orientation, distance, support, or object state, with source/target variables such as liquid level, count, or fill amount co-varying correctly. \emph{Interaction and affordance geometry} evaluates plausible contact among agents, tools, and targets, including grip, body pose, gaze, tool use, and support. \emph{Causal process composition} requires the final state to follow visible triggers and intermediate stages rather than appearing prematurely. The highest level, \emph{constraint and counterfactual world-state reasoning}, introduces negations, selective operations, blocked actions, or conditions under which an event should not occur.

\TabCapabilityTree

The tree guides dataset construction, prompt authoring, and evaluation. Each case is assigned capability tags and written with observable $t_1$--$t_4$ states, required predicates, forbidden states, non-target objects, and special constraints. In evaluation, L-levels define the conceptual hierarchy, C0--C9 scores provide coarse judgments over ability families, and D0--D14 subscores record the visual evidence behind those judgments. This lets ImageTime aggregate model behavior back to tree-level scores under the same hierarchy used for task design.

\subsection{Data Construction}
The dataset was constructed with a human-led, LLM-assisted pipeline. We first manually collected 22 common domains and designed 200 foundational everyday and domain-specific scenario seeds covering visually concrete and physically grounded actions, including household manipulation, kitchen processes, clothing, sports, navigation, social interaction, machines and transportation, pets, gardening, laboratory operations, quantity reasoning, occlusion, long-horizon processes, and explicit constraints.

LLMs were used as constrained expansion and specification tools, not as unconstrained task inventors. Starting from the manually collected domains and seeds, the LLM expanded and normalized the inventory under fixed rules: each task had to preserve the intended domain, use plausible actors and objects, occur in a physically appropriate scene, admit a visible four-stage progression, and expose at least one diagnostic consistency requirement. This produced 375 action concepts across 22 domains; the complete action-concept inventory is reported in Appendix Section~\ref{sec:appendix-action-inventory} and Table~\ref{tab:appendix-action-concepts}. For each concept, we created two variants with semantically related but visually distinct setups, resulting in 750 benchmark cases.

For each case, an LLM-assisted drafting step produced a structured process specification and a generation prompt. The specification records the action, visual setup, expected $t_1$--$t_4$ states, key entities, stable non-target objects, state predicates, temporal constraints, forbidden premature or impossible states, and task-specific consistency requirements. The prompt verbalizes these requirements as a single-image 2$\times$2 motion-sheet instruction. We then generated pilot images with GPT Image 2~\cite{openai2026gptimage2,openai2026chatgptimages2} and manually repaired 170 cases whose prompts induced obvious factual, scale, scene, or common-sense mismatches. The repair step helps ensure that measured failures reflect model limitations rather than ambiguous or physically mismatched prompts.

\subsection{Output Settings}
ImageTime supports prompt-only and reference-conditioned settings under the same four-keyframe interface. In the prompt-only setting, used for the main comparison in this paper, the model receives only the textual motion-sheet prompt and must synthesize all four panels from scratch.

For edit-capable models, ImageTime also defines a reference-conditioned scaffold setting. For each case, we generate a square 1:1 reference image with GPT Image 2~\cite{openai2026gptimage2,openai2026chatgptimages2} to instantiate the initial world state. This image is placed in the top-left $t_1$ cell of a square 2$\times$2 scaffold, and the model completes $t_2$, $t_3$, and $t_4$ from the same action prompt. This setting tests whether a model can preserve a given reference scene while rolling it forward through ordered state changes.

Both settings are evaluated as single-pass generation tasks. For each model--case pair, the prompt is submitted once and the returned image is scored as-is. We do not use multi-turn dialogue, iterative correction, judge feedback, rejection sampling, or manual selection during evaluation. The benchmark therefore measures whether a model can construct the full ordered motion sheet in one generation attempt.

The same structured judge applies to both settings. In prompt-only evaluation, C1 reference grounding is \texttt{null}. In scaffold-conditioned evaluation, C1 measures whether the generated $t_1$ panel and subsequent panels preserve the reference subject, scene, style, camera view, layout, and initial state. C1 is therefore excluded from the prompt-only totals reported in this paper.

\section{Evaluation Protocol}
\label{sec:evaluation-protocol}

\subsection{Layout Validation}
Each generated result first passes through a layout gate. The expected output is a single 2$\times$2 motion sheet with four readable panels ordered as top-left, top-right, bottom-left, and bottom-right, corresponding to $t_1$ initial state, $t_2$ early action state, $t_3$ intermediate action state, and $t_4$ final state. The check records whether all panels are present, visually separable, complete, readable, and temporally ordered. For scaffold-conditioned settings, the top-left panel is also checked against the reference image for subject, object, scene, style, camera, and starting-state preservation.

Layout validity is reported separately from spatiotemporal consistency. If the output is not a usable four-panel sheet, C0 is set to zero and downstream temporal, causal, and interaction scores are conservatively capped.

\subsection{Structured VLM-as-Judge}
We use a structured VLM-as-judge protocol instead of a single holistic preference score. In all reported experiments, GPT-5.5 is used to score the images generated by every evaluated model. That is, each prompt-only motion sheet from GPT Image 2, Nano Banana 2, Seedream 5.0 Lite, FLUX.2 Pro, Qwen-Image-2512, Z-Image-Turbo, HunyuanImage-2.1, and SDXL is passed to the same GPT-5.5 judging pipeline with the same structured inputs and rubric. We use GPT-5.5 because it is a state-of-the-art multimodal model with strong visual parsing, instruction following, and structured reasoning ability, which are required for checking whether a generated four-panel sheet preserves entities, spatial anchors, temporal order, causal preconditions, interactions, and constraints.

The evaluation has three linked layers: the Progressive Capability Tree defines the abilities being probed, C-scores summarize the main evaluation axes, and D-scores provide fine-grained visual evidence for interpreting each C-score. For each generated image, GPT-5.5 receives the generated image, process specification, original generation prompt, C0--C9 capability rubric, D0--D14 diagnostic rubric, conservative score-capping rules, and, when applicable, the scaffold reference image. The prompt is treated as the authority for the required subject, action, scene, style, camera view, objects, constraints, and output format; GPT-5.5 is instructed not to infer the task from the generated image alone.

The judge first parses the intended process from the prompt and case specification, including the required $t_1$--$t_4$ states, actors, target objects, tools, non-target objects, temporal order, causal preconditions, quantity or occlusion constraints, and special conditions such as negation, blocked actions, gaze tracking, multi-agent interaction, or counterfactual requirements. It then parses the generated image panel by panel and tracks prompt adherence, entity identity, spatial anchors, camera scale, motion paths, interaction geometry, source--target state conservation, biomechanics, and task-specific constraints across panels.

The final judgment contains two complementary score families. Capability scores C0--C9 cover layout validity, reference grounding, entity consistency, spatial-view consistency, motion continuity, temporal ordering, causal process consistency, interaction consistency, constraint/counterfactual sensitivity, and visual quality. Diagnostic subscores D0--D14 cover panel parse integrity, task-entity binding, object permanence, background anchors, spatial coordinates, camera scale and depth, pose-position coupling, contact geometry, state-transition visibility, phase distinctiveness, gaze/intent alignment, quantity and attribute binding, occlusion consistency, constraint execution, and visual readability. Thus C-scores identify which high-level abilities succeed or fail, while D-scores explain the visual evidence behind those judgments.

The C/D rubric is defined in \tabref{tab:judge-rubric-c} and \tabref{tab:judge-rubric-d}, and mapped back to the tree in \tabref{tab:tree-judge-mapping}. Each tree node is measured by a small set of C dimensions and supporting D diagnostics; some dimensions support multiple nodes because the same evidence is shared across identity, spatial, transition, interaction, and causal reasoning. The mapping in \tabref{tab:tree-judge-mapping} is the exact mapping used for the prompt-only tree-level results.

\TabJudgeRubric
\begin{center}
    \centering
    \footnotesize
    \renewcommand{\arraystretch}{1.08}
    \setlength{\tabcolsep}{3pt}
    \begin{tabular}{@{}p{0.23\linewidth}p{0.15\linewidth}p{0.26\linewidth}p{0.30\linewidth}@{}}
\toprule
Tree node & C dimensions & D dimensions & Aggregation note \\
\midrule
Gate: Protocol & C0 & D0, D14 & Interface gate; reported separately from L0--L6. \\
L0: Static grounding & C2, C3, C9 & D1, D14 & Static world instantiation in prompt-only scoring. \\
L1: Identity and anchors & C2, C3 & D2, D3, D5 & Entity, background, and viewpoint persistence. \\
L2: Spatial transition & C3, C4 & D4, D5, D6, D8, D9 & Spatial relations, displacement, and phase evidence. \\
L3: Object-state transition & C2, C6 & D2, D8, D11, D12 & Object lifecycle, conservation, quantity, and occlusion evidence. \\
L4: Interaction and affordance & C4, C6, C7 & D6, D7, D10 & Motion, contact, tool use, and attention evidence. \\
L5: Causal process & C5, C6 & D8, D9, D11 & Ordered preconditions, visible transitions, and conservation. \\
L6: Constraint/counterfactual & C8 & D13 & Computed only on applicable constraint cases. \\
\bottomrule
\end{tabular}

    \captionof{table}{Exact prompt-only aggregation from Progressive Capability Tree nodes to structured C/D judge scores. C-scores provide coarse capability judgments, while D-scores provide the visual evidence used to interpret and aggregate each tree node. C1 reference grounding is reserved for reference-conditioned edit/scaffold settings and is not used in the prompt-only tree scores.}
    \label{tab:tree-judge-mapping}
\end{center}

\subsection{Metrics}
All capability and diagnostic scores use a discrete integer scale from 0 to 10. A score of 10 denotes a correct or near-correct result, mid-range scores indicate partial satisfaction with visible flaws, and 0 denotes a failed, contradicted, missing, unreadable, or unrelated result. When evidence is ambiguous between adjacent scores, the judge chooses the lower score. Each C-score also reports a confidence value in $[0,1]$. A score may be \texttt{null} only when the dimension is genuinely not applicable, such as C1 for prompt-only settings, C8/D13 for tasks without explicit constraints, D11 for tasks without quantity or attribute binding, and D12 for tasks without occlusion or reappearance.

For aggregate reporting, C mean averages applicable C-score dimensions except C1 in prompt-only experiments; D mean averages applicable D-score dimensions; and Overall mean is the balanced average of C mean and D mean. C9 visual quality is included as one C-score dimension rather than a separate ranking axis, so visual polish cannot compensate for missing temporal phases, incorrect causal order, or failed interaction geometry. In reference-conditioned edit/scaffold settings, C1 is reported as a dedicated reference-grounding score, but it remains null and excluded for the prompt-only comparison reported here.

For tree-level analysis, let $C_L$ and $D_L$ denote the mapped capability and diagnostic dimensions for tree node $L$ in \tabref{tab:tree-judge-mapping}. We compute
\[
T_L = \frac{1}{2}\mathrm{mean}(C_L) + \frac{1}{2}\mathrm{mean}(D_L),
\]
where null dimensions are omitted and the available family is used if only one family applies. This balanced aggregation prevents the larger diagnostic set from overwhelming the coarser capability score, while still requiring each tree-level claim to be backed by visual evidence. The layout gate is reported separately; L0--L6 are used to analyze progressive capability degradation.

The rubric also includes conservative score-capping rules. Scores are capped when the output is not a usable ordered 2$\times$2 sheet; when panels form a panorama or collage rather than four complete stages; when the requested action, actor, target, tool, recipient, scene, or style is wrong; when recurring objects disappear, duplicate, or change identity/material without task justification; when motion tasks lack objective displacement or show copied phases; when source and target states fail to co-vary in transfer, filling, emptying, or assembly tasks; or when blocked or counterfactual conditions are violated. These caps prevent high-level consistency scores when lower-level evidence contradicts the required process.

\subsection{Quality Control and Rubric Calibration}
The current large-scale results use GPT-5.5 as the primary scoring model for the structured VLM judge. Before final scoring, GPT-5.5-assisted multi-round quality checks were used to refine the prompts, case specifications, judge prompts, and rubric definitions. During dataset construction and repair, pilot outputs were inspected to identify prompts with factual, scale, scene, or common-sense mismatches; these cases were rewritten and regenerated before final prompt-only scoring.

The repeated checks also calibrated the GPT-5.5 judge prompts, capability definitions, diagnostic criteria, and conservative score-capping rules. They focused on whether GPT-5.5 reliably penalized visible failures in entity consistency, spatial-view stability, motion continuity, temporal ordering, causal process consistency, interaction geometry, and constraint sensitivity. The main reported comparison remains automated and reproducible from the released prompts, images, GPT-5.5 score files, and judge rubric.
\FloatBarrier

\section{Experiments}

\subsection{Models and Settings}
We evaluate eight text-to-image models in the prompt-only ImageTime setting: GPT Image 2~\cite{openai2026gptimage2,openai2026chatgptimages2}, Nano Banana 2~\cite{googledeepmind2026nanobanana2}, Seedream 5.0 Lite~\cite{bytedance2026seedream50lite}, FLUX.2 Pro~\cite{blackforestlabs2025flux2}, Qwen-Image-2512~\cite{qwen2026qwenimage2512,wu2025qwenimagetechnicalreport}, Z-Image-Turbo~\cite{cai2025z}, HunyuanImage-2.1~\cite{tencent2025hunyuanimage21}, and SDXL~\cite{podell2024sdxl}. Each model is evaluated on the same 750 ImageTime prompts. The prompt-only setting is intentionally strict: the model receives no reference image and must synthesize a complete ordered 2$\times$2 motion sheet from text alone. Generation is single-pass: for each prompt and model, we make one generation request, keep the first returned image, and do not run follow-up dialogue, iterative revision, regeneration-based selection, or human curation. The full prompts, generated images, judge inputs, and score files will be released with the benchmark data; representative same-prompt visual comparisons are provided in Appendix Section~\ref{sec:appendix-example-case-gallery}, such as \figref{fig:appendix-examples-animation}.

All reported scores use the structured VLM judge described in Section~\ref{sec:evaluation-protocol}. We use GPT-5.5 to score all images generated by all evaluated models. GPT-5.5 receives the generated image, the original generation prompt, the process specification, and the C/D rubric. It first checks the panel layout, then parses each panel against the required $t_1$--$t_4$ process, and finally returns C0--C9 capability scores, D0--D14 diagnostic scores, confidence values, and failure labels. For aggregate reporting, C mean averages the applicable capability scores except C1, which is not applicable in prompt-only experiments; D mean averages applicable diagnostic subscores; Overall mean is the balanced average of the two families.

\FigAllModelsCapabilityRadar

\subsection{Capability Decomposition}
\figref{fig:all-models-capability-radar} first summarizes the C-score capability profile of each model. This view is useful before looking at any aggregate ranking because it reveals which abilities form the model signature. The strongest systems do not merely score higher on visual quality or layout: GPT Image 2~\cite{openai2026gptimage2,openai2026chatgptimages2}, Nano Banana 2~\cite{googledeepmind2026nanobanana2}, and Seedream 5.0 Lite~\cite{bytedance2026seedream50lite} keep a broad envelope across entity consistency, spatial-view consistency, motion continuity, temporal ordering, causal process consistency, interaction consistency, and constraint sensitivity. By contrast, the largest separations appear on C4 motion continuity, C5 temporal ordering, C6 causal process consistency, and C8 constraint sensitivity. This already indicates that ImageTime is not mainly measuring whether a model can draw four panels, but whether it can maintain a process across them.

The radar plot also separates two kinds of progress. Open-weight or lower-cost systems such as Qwen-Image-2512~\cite{qwen2026qwenimage2512,wu2025qwenimagetechnicalreport} and Z-Image-Turbo~\cite{cai2025z} often satisfy the basic interface and preserve some static world structure, but their envelopes contract sharply on motion, temporal, causal, and constraint-sensitive axes. SDXL~\cite{podell2024sdxl} shows an earlier failure mode: its default generation behavior is poorly matched to the strict ordered motion-sheet interface, so higher-level process reasoning is often not meaningfully exposed.

\subsection{Diagnostic Evidence}
\figref{fig:d-diagnostic-heatmap} then provides the visual evidence behind the C-score gaps. The most informative diagnostic dimensions are D8 state-transition visibility, D9 phase distinctiveness, D6 trajectory--pose coupling, D11 quantity and attribute binding, D12 occlusion/reappearance consistency, and D13 constraint execution. These are precisely the dimensions where a model must do more than independently render plausible snapshots. It must show where an object moved from, how a pose supports the displacement, whether source and target states co-vary, and whether explicit constraints prevent invalid outcomes.

\FigDiagnosticHeatmap

Together, \figref{fig:all-models-capability-radar} and \figref{fig:d-diagnostic-heatmap} show why a structured judge is needed. The C-scores identify broad capability gaps, while the D-scores localize the evidence that makes those gaps interpretable. For example, a low causal-process score is not treated as an opaque failure: it can be traced to missing transition visibility, repeated phases, impossible contact, or quantity non-conservation. This decomposition is what lets ImageTime connect a model-level ranking back to concrete world-state failures.

\subsection{Tree-Level Capability Analysis}
\figref{fig:tree-level-curve} aggregates the C/D evidence back into the Progressive Capability Tree. This view changes the interpretation from ``which metric is low'' to ``where along the world-modeling progression the model begins to fail.'' Lower levels test static grounding and identity preservation; higher levels test spatial and object-state transitions, interaction geometry, causal process composition, and constraint-sensitive world-state reasoning.

\FigTreeLevelCurve

The key pattern is not strict monotonic decline for every model, but a clear upper-level degradation trend, especially among weaker systems. GPT Image 2 remains comparatively stable from L0 static grounding through L5 causal process and L6 constraint/counterfactual reasoning, suggesting that its advantage is not confined to rendering or prompt grounding. Nano Banana 2 and Seedream 5.0 Lite preserve strong lower- and mid-level scores but degrade more at the highest constraint-sensitive level. FLUX.2 Pro starts from a reasonable static-grounding profile but drops when the task requires causal process composition. Qwen-Image-2512, Z-Image-Turbo, and HunyuanImage-2.1 show larger upper-level collapses, indicating that static scene construction is easier than maintaining a causal process over ordered states. Thus, the tree analysis turns the C/D decomposition into a capability ladder: current models increasingly struggle as the task moves from scene construction to constrained world-state evolution.

\subsection{Overall Prompt-Only Results}
\begin{figure*}[!t]
    \centering
    \includegraphics[width=0.92\textwidth]{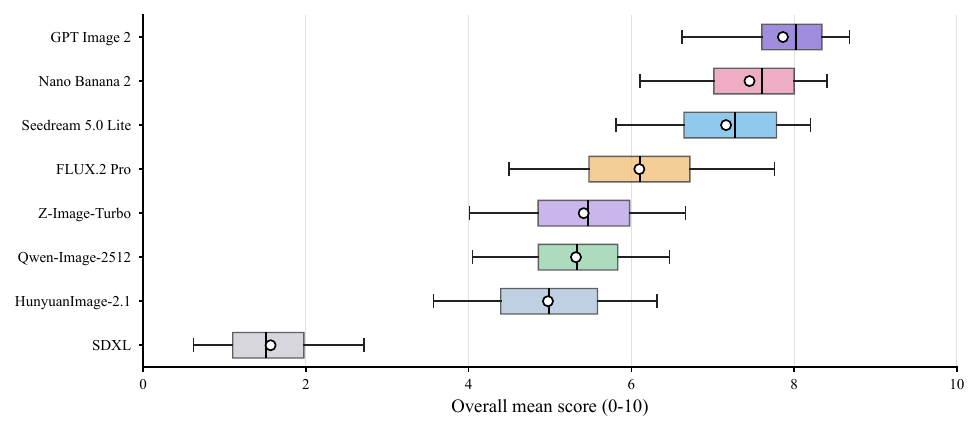}
    \caption{Distribution of Overall mean scores across the 750 prompt-only cases for each model. Boxes show the interquartile range; black lines show medians; whiskers show 5th--95th percentiles; hollow circles show means.}
    \label{fig:core-score-distribution}
\end{figure*}

\tabref{tab:all-models-main-results} gives the compact model-level summary after the capability and diagnostic evidence has been established, and \figref{fig:core-score-distribution} shows the corresponding distribution of Overall mean scores over all 750 prompt-only cases. GPT Image 2 obtains the strongest Overall mean, followed by Nano Banana 2 and Seedream 5.0 Lite. FLUX.2 Pro forms the next tier, while Z-Image-Turbo, Qwen-Image-2512, and HunyuanImage-2.1 remain below the frontier systems on the balanced aggregate. SDXL has the lowest Overall mean, mainly because many outputs fail the ordered four-panel interface before higher-level temporal reasoning can be evaluated.

\begin{table*}[!t]
    \centering
    \footnotesize
    \renewcommand{\arraystretch}{1.08}
    \setlength{\tabcolsep}{4.0pt}
    \resizebox{0.74\textwidth}{!}{\begin{tabular}{@{}lrrrr@{}}
\toprule
Model & N & C mean & D mean & Overall mean \\
\midrule
GPT Image 2~\cite{openai2026gptimage2,openai2026chatgptimages2} & 750 & \cellcolor[HTML]{EDAAAE}7.86 & \cellcolor[HTML]{EDAAAE}7.87 & \cellcolor[HTML]{EDAAAE}7.86 \\
Nano Banana 2~\cite{googledeepmind2026nanobanana2} & 750 & \cellcolor[HTML]{EFB3A3}7.43 & \cellcolor[HTML]{EFB2A4}7.47 & \cellcolor[HTML]{EFB3A3}7.45 \\
Seedream 5.0 Lite~\cite{bytedance2026seedream50lite} & 750 & \cellcolor[HTML]{F1BA9A}7.13 & \cellcolor[HTML]{F0B89C}7.20 & \cellcolor[HTML]{F1B99B}7.16 \\
FLUX.2 Pro~\cite{blackforestlabs2025flux2} & 750 & \cellcolor[HTML]{F5D499}5.92 & \cellcolor[HTML]{F4CC94}6.28 & \cellcolor[HTML]{F4D096}6.10 \\
Z-Image-Turbo~\cite{cai2025z} & 750 & \cellcolor[HTML]{F7E4A5}5.14 & \cellcolor[HTML]{F5D99D}5.69 & \cellcolor[HTML]{F6DEA1}5.41 \\
Qwen-Image-2512~\cite{qwen2026qwenimage2512,wu2025qwenimagetechnicalreport} & 750 & \cellcolor[HTML]{F7E5A6}5.09 & \cellcolor[HTML]{F6DB9E}5.55 & \cellcolor[HTML]{F6E0A2}5.32 \\
HunyuanImage-2.1~\cite{tencent2025hunyuanimage21} & 750 & \cellcolor[HTML]{F5E7A9}4.91 & \cellcolor[HTML]{F7E7A7}5.04 & \cellcolor[HTML]{F7E7A7}4.98 \\
SDXL~\cite{podell2024sdxl} & 750 & \cellcolor[HTML]{DCEEF7}1.49 & \cellcolor[HTML]{D9EDF7}1.64 & \cellcolor[HTML]{DBEEF7}1.57 \\
\bottomrule
\end{tabular}
}
    \caption{Prompt-only ImageTime results over 750 motion-sheet generations per model. C mean averages applicable capability scores except prompt-only reference grounding (C1), D mean averages applicable diagnostic subscores, and Overall mean is the balanced average of the two families. Null dimensions are omitted rather than reported as separate columns.}
    \label{tab:all-models-main-results}
\end{table*}

The table is useful not only as a leaderboard, but as a consistency check between coarse and fine-grained evaluation. C mean and D mean are close for the strongest systems, which suggests that their broad capability scores are supported by fine-grained visual evidence. For several middle-tier systems, D mean is slightly higher than C mean, reflecting that they can sometimes produce local visual evidence of objects, anchors, or readable panels even when the full process does not cohere. The Overall mean therefore serves as a compact summary, while the preceding figures explain why each model occupies its tier.

\figref{fig:core-score-distribution} asks a different question from \tabref{tab:all-models-main-results}: not only how high the model scores on average, but how reliably it maintains spatiotemporal consistency across diverse prompts. GPT Image 2, Nano Banana 2, and Seedream 5.0 Lite have higher medians and comparatively compact interquartile ranges, indicating that their advantage is not driven by a small number of easy cases. Z-Image-Turbo, Qwen-Image-2512, and HunyuanImage-2.1 show lower medians and broader low-score behavior, which is important for deployment: a model may occasionally produce a coherent motion sheet, but still be unreliable under varied process instructions.

Read together, \tabref{tab:all-models-main-results} and \figref{fig:core-score-distribution} distinguish average capability from reliability. The table gives the central tendency, while the distribution exposes whether that central tendency is stable across the benchmark or hides a heavy tail of failures.

\subsection{Category-Level Robustness}
\figref{fig:domain-model-heatmap} reports category-level Overall mean scores over all 22 fine-grained ImageTime categories, reconstructed from the filename and case-id prefixes. This view is no longer restricted to only the most discriminative rows; it shows the coverage of the full prompt set, including animation, caregiving, clothing, constraints, daily actions, games, gardening, household, kitchen, lab, long-horizon processes, machines, nature, navigation, occlusion, pets, quantity-focused tasks, repair, social interaction, sports, systems, and unboxing. The pattern reveals both category difficulty and model robustness: household, kitchen, lab, and gardening cases are comparatively high on average, while systems, constraints, and quantity-focused cases are harder because they stress stable anchors, blocked or selective actions, and source--target state conservation. These category differences matter because they show that ImageTime is not a single generic difficulty scale; different prompt families expose different weaknesses in world-state tracking.

\FigDomainHeatmap

\subsection{Score-Family and Cost Analysis}
\figref{fig:quality-vs-core} compares C mean and D mean at the per-output level, while \figref{fig:cost-performance} compares reported generation cost for 750 images with Overall mean performance.

\FigScoreFamilyAndCost

Most model means lie close to the diagonal, but the scatter reveals an important evaluation risk: some individual outputs can receive stronger coarse capability scores than fine-grained diagnostic evidence. Points below the diagonal indicate motion sheets that look broadly plausible while still lacking detailed support for object lifecycle, interaction geometry, temporal phase separation, or constraint execution. This supports the decision to report both capability and diagnostic families rather than collapsing the judge into a single holistic score.

Because public cost estimates are not a controlled economic benchmark, the cost plot should be read as a practical reference. The notable result is that low generation cost alone does not guarantee reliable temporal-process generation: inexpensive systems can be attractive for scale, but their lower Overall means indicate weaker robustness under ordered state-change tasks. At the same time, the placement of Qwen-Image-2512 and Z-Image-Turbo shows that open or low-cost systems are already entering the spatiotemporal-consistency regime, even if frontier proprietary systems still dominate the upper-right performance region.

\FloatBarrier

\section{Discussion}

\subsection{Practical Insights and Future Outlook}

ImageTime shows that temporally ordered image generation is not simply a matter of drawing four attractive panels. The strongest systems usually understand the requested scene and can produce a readable motion sheet, but they still fail on details that require a persistent world state. Common failures include copied phases, camera and background-anchor drift, target-object jumps, impossible contact geometry, unchanged source quantities after transfer, and premature final states. These errors are difficult to detect with single-image alignment metrics because each individual panel can appear plausible in isolation. The results also separate visual quality from process coherence: several models score reasonably on C9 visual quality while scoring much lower on C4--C6 motion, temporal order, and causal consistency. This suggests that current image priors are often strong enough to render a scene, but weaker at maintaining the hidden variables that make a process coherent, such as where an object came from, how much material remains, which actor is interacting with which target, and which intermediate contact must occur before the final state. The four-keyframe protocol is useful because it forces these hidden variables to become visually inspectable.

The Progressive Capability Tree provides a compact interpretation of these failures. Lower tree levels ask whether the model can instantiate a static world and preserve identities, while higher levels ask whether the same world can evolve through spatial transitions, object-state transitions, interactions, causal preconditions, and constraints. The empirical tree scores show a progressive drop for most weaker models. This pattern supports the benchmark design: failures are not random artifacts of the judge, but align with increasingly demanding visual reasoning requirements. ImageTime should therefore be interpreted as a behavioral probe rather than direct evidence of an internal world model. High scores indicate that a model can externalize world-model-like consistency in generated images, while low scores reveal failures in maintaining visually inspectable world states under temporal change.

The results carry several practical messages for both academic and industrial model development. First, open-weight or low-cost systems such as Qwen-Image-2512 and Z-Image-Turbo are not merely static-image baselines: they often follow the requested 2$\times$2 interface and exhibit nontrivial evidence of entity preservation, spatial anchoring, and ordered state change. Although they lag behind the strongest proprietary systems on causal, motion, and constraint-sensitive dimensions, their ability to produce readable motion sheets suggests that basic spatiotemporal process generation is already emerging outside closed frontier models. Second, reasoning-oriented generation is a useful and important direction. Systems such as GPT Image 2 and Nano Banana 2 achieve the strongest overall results, and their advantage is most visible on dimensions that require planning across panels, including temporal ordering, causal process consistency, motion continuity, and constraint sensitivity. At the same time, these systems still fail on object-state conservation, contact geometry, subtle occlusion, and blocked or counterfactual actions. Reasoning improves the ability to plan a process, but it does not yet guarantee robust visual state tracking.

The qualitative Dense-Grid Generative Potential study in Appendix Section~\ref{sec:appendix-dense-grid-generative-potential} pushes this observation beyond the benchmark's four-keyframe interface. When GPT Image 2 and Nano Banana 2 are asked to produce denser 4x2, 4x4, 4x6, and 4x8 temporal grids, they can often obey the more complex layout and generate visually coherent anime-style or realistic motion sheets, as illustrated by \figref{fig:appendix-dense-grid-anime-rainy-street}. This is an encouraging signal for industry: current image generators are beginning to externalize longer visual processes rather than only isolated snapshots. However, close inspection still reveals residual world-state errors, including subtle identity drift, repeated or skipped phases, inconsistent background anchors, and imperfect object-state conservation. Thus, dense-grid examples strengthen the same conclusion as the quantitative benchmark: strong models already show impressive visual world-modeling behavior, but their temporal state tracking remains fragile when the requested process becomes longer or more layout-constrained.

Overall, the benchmark suggests that current image models already exhibit basic visual world-model-like behavior: many can instantiate a coherent scene, preserve identities and anchors, and depict simple ordered transitions. The remaining gap is not the absence of world-state behavior, but its fragility under longer, more constrained, or more causally demanding processes. For deployed image-generation systems, this implies that progress should be reported not only through visual quality, prompt following, and preference scores, but also through explicit diagnostics of state persistence, transition visibility, causal preconditions, and constraint execution. For research, it suggests that image generation benchmarks can serve as controlled probes of visual world modeling, bridging static image evaluation and dense video evaluation.

Looking forward, ImageTime points to a broader agenda for improving the world-modeling ability of image generation systems. From the data-collection side, future work can build scalable pipelines that mine continuous videos for temporally grounded training examples: detect action segments, select key states, track entities and contact events, infer state predicates, and convert the resulting process traces into paired prompts, references, and supervision targets. Such pipelines would allow training data to move beyond isolated image-caption pairs toward structured examples of how the visual world changes over time. From the model-training side, image models could be explicitly trained to preserve world states across keyframes, using objectives for identity consistency, spatial anchoring, quantity conservation, contact validity, causal preconditions, and constraint satisfaction. This may improve not only spatiotemporal consistency metrics, but also the model's broader understanding of real-world objects, interactions, and physical processes. From the evaluation side, benchmarks should combine key-state probes, dense-video diagnostics, GPT-5.5-assisted rubric calibration, and judge ablations, so that improvements in visual quality are separated from improvements in state tracking and causal coherence. From the deployment side, generation systems could expose controllable intermediate states, editable process plans, and uncertainty signals when the requested action requires fragile physical reasoning. Together, these directions suggest a path from better benchmark scores toward image models that more faithfully represent how the world persists, changes, and constrains possible actions.

\subsection{Limitations}

There are several boundaries to the current study. First, ImageTime evaluates key-state consistency, not dense video motion. A model that performs well here may still fail at frame-by-frame video dynamics, and a video model may solve some interpolation problems that a static image model cannot. Second, the main cross-model experiment uses the strict prompt-only setting. This is deliberately challenging, but it also asks the model to solve layout, prompt parsing, and temporal construction at once. Reference-conditioned and scaffold-conditioned settings can isolate different sources of failure and should be expanded in future comparisons. Third, the large-scale scores rely on GPT-5.5 as a structured VLM judge. We reduced obvious judge failures by providing the original prompt, process specification, detailed C/D rubric, and conservative score-capping rules, but GPT-5.5 judging remains an imperfect proxy for human assessment, especially for small contacts, subtle reflections, occlusions, and physically plausible but unusual actions. Finally, ImageTime evaluates generated outputs rather than directly observing the internal representations of the model. The most reproducible comparison is therefore the released set of generated images and score files, together with the scripts that produced and analyzed them.

\section{Conclusion}

We introduced ImageTime, a diagnostic benchmark for evaluating whether image generation models can maintain a coherent visual world across temporally ordered key states. The benchmark combines a four-keyframe motion-sheet protocol, a progressive capability tree, structured process specifications, and GPT-5.5-based C/D scores that localize both high-level capability gaps and fine-grained visual evidence.

Across 750 prompt-only tasks and eight image generation systems, the results show that strong visual synthesis does not guarantee reliable temporal-process construction. GPT Image 2 and Nano Banana 2 perform best overall, but even high-scoring systems exhibit failures in causal preconditions, intermediate-state visibility, quantity conservation, interaction geometry, and constraints. Lower-scoring systems often fail earlier, either by not producing the required four-panel interface or by repeating near-static panels without a meaningful state transition.

ImageTime therefore complements static image benchmarks and video benchmarks. It does not claim to measure all forms of temporal understanding, but it provides a controlled and interpretable probe of key-state consistency. This makes it useful for tracking progress from static scene rendering toward image generation systems that can preserve identities, objects, spatial relations, and causal structure as a visual world changes over time. Appendix Section~\ref{sec:appendix-reproducibility-ethics} summarizes the release artifacts, GPT-5.5 scoring setup, intended use, and evaluation boundaries.

\bibliography{references/references}

\FloatBarrier
\onecolumn
\section*{Appendix Contents}
\label{sec:appendix-contents}

The appendix contains dense qualitative grids, the complete action inventory,
domain-level example galleries, and reproducibility and ethics details. The page
directory below is included to make the image-heavy appendix easier to navigate.

\vspace{1.2em}
\begin{center}
\renewcommand{\arraystretch}{1.35}
\begin{tabular}{p{0.08\textwidth}p{0.68\textwidth}r}
\toprule
 & \textbf{Appendix Section} & \textbf{Page} \\
\midrule
A &
\hyperref[sec:appendix-dense-grid-generative-potential]{Dense-Grid Generative Potential} &
\pageref{sec:appendix-dense-grid-generative-potential} \\
B &
\hyperref[sec:appendix-action-inventory]{ImageTime Action Inventory} &
\pageref{sec:appendix-action-inventory} \\
C &
\hyperref[sec:appendix-example-case-gallery]{Example Case Gallery} &
\pageref{sec:appendix-example-case-gallery} \\
D &
\hyperref[sec:appendix-reproducibility-ethics]{Reproducibility and Ethics} &
\pageref{sec:appendix-reproducibility-ethics} \\
\bottomrule
\end{tabular}
\end{center}

\clearpage


\appendix
\section{Dense-Grid Generative Potential}
\label{sec:appendix-dense-grid-generative-potential}
The preceding benchmark uses four-keyframe motion sheets to make spatiotemporal consistency measurable at scale. Here we add a qualitative stress test with denser temporal grids. We randomly sample prompts from the domain-level example pool and ask GPT Image 2 and Nano Banana 2 to generate 4x2, 4x4, 4x6, and 4x8 motion sheets, with multiple prompts per grid shape. Here 4xN denotes four temporal rows with N panels per row. In each comparison, the two raw model outputs are shown at matched display scale while preserving their original aspect ratios. These examples are not included in the quantitative scores; they illustrate how current image generators may extend a short action prompt into longer ordered visual processes.

\begin{center}
    \makebox[\textwidth][c]{\includegraphics[width=1.02\textwidth,height=0.84\textheight,keepaspectratio]{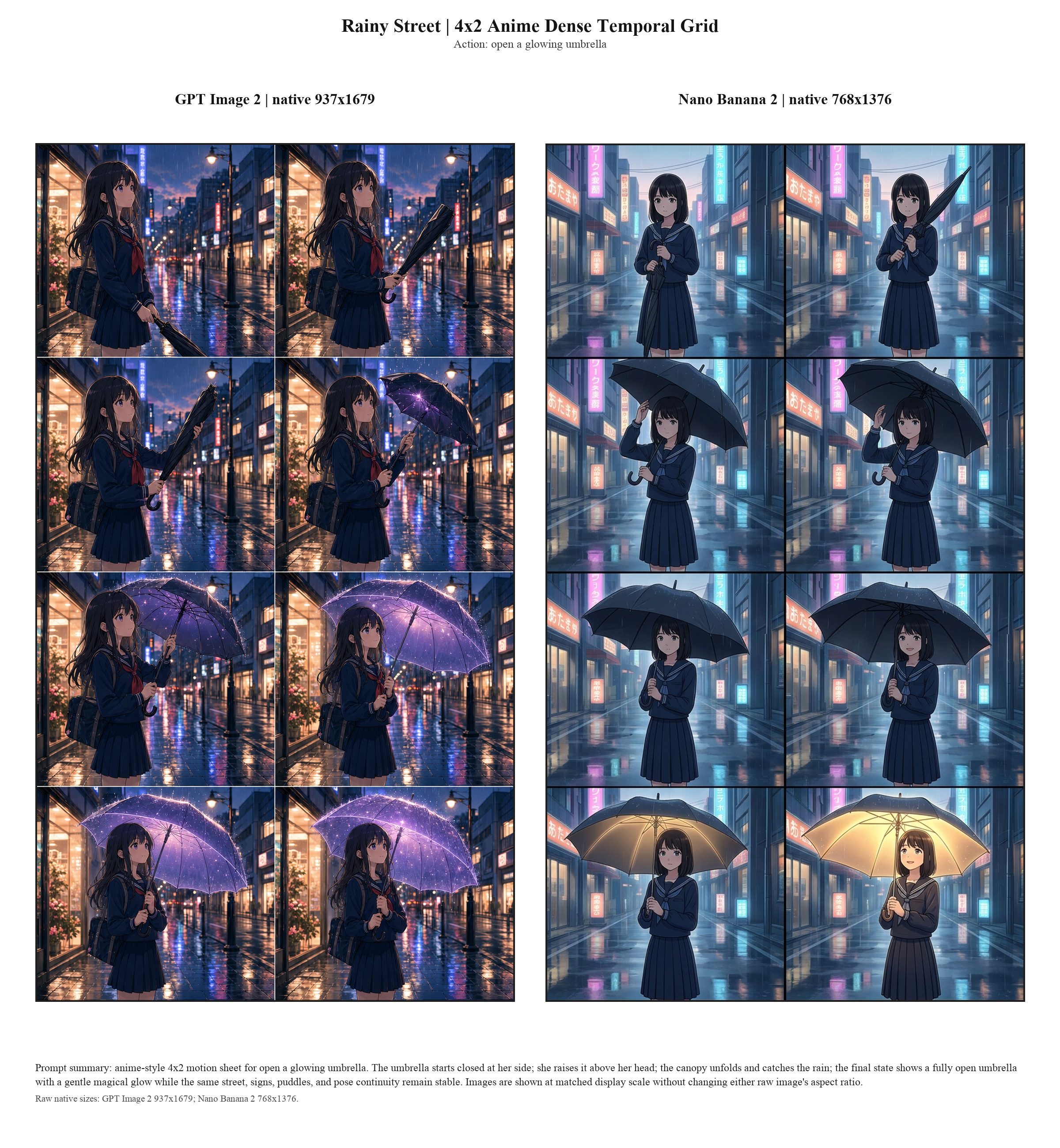}}\par
    \captionof{figure}{Anime-style dense-grid qualitative comparison for a Rainy Street prompt using a 4x2 temporal grid, i.e., four rows with two panels per row.}
    \label{fig:appendix-dense-grid-anime-rainy-street}
\end{center}
\clearpage

\begin{center}
    \makebox[\textwidth][c]{\includegraphics[width=1.02\textwidth,height=0.84\textheight,keepaspectratio]{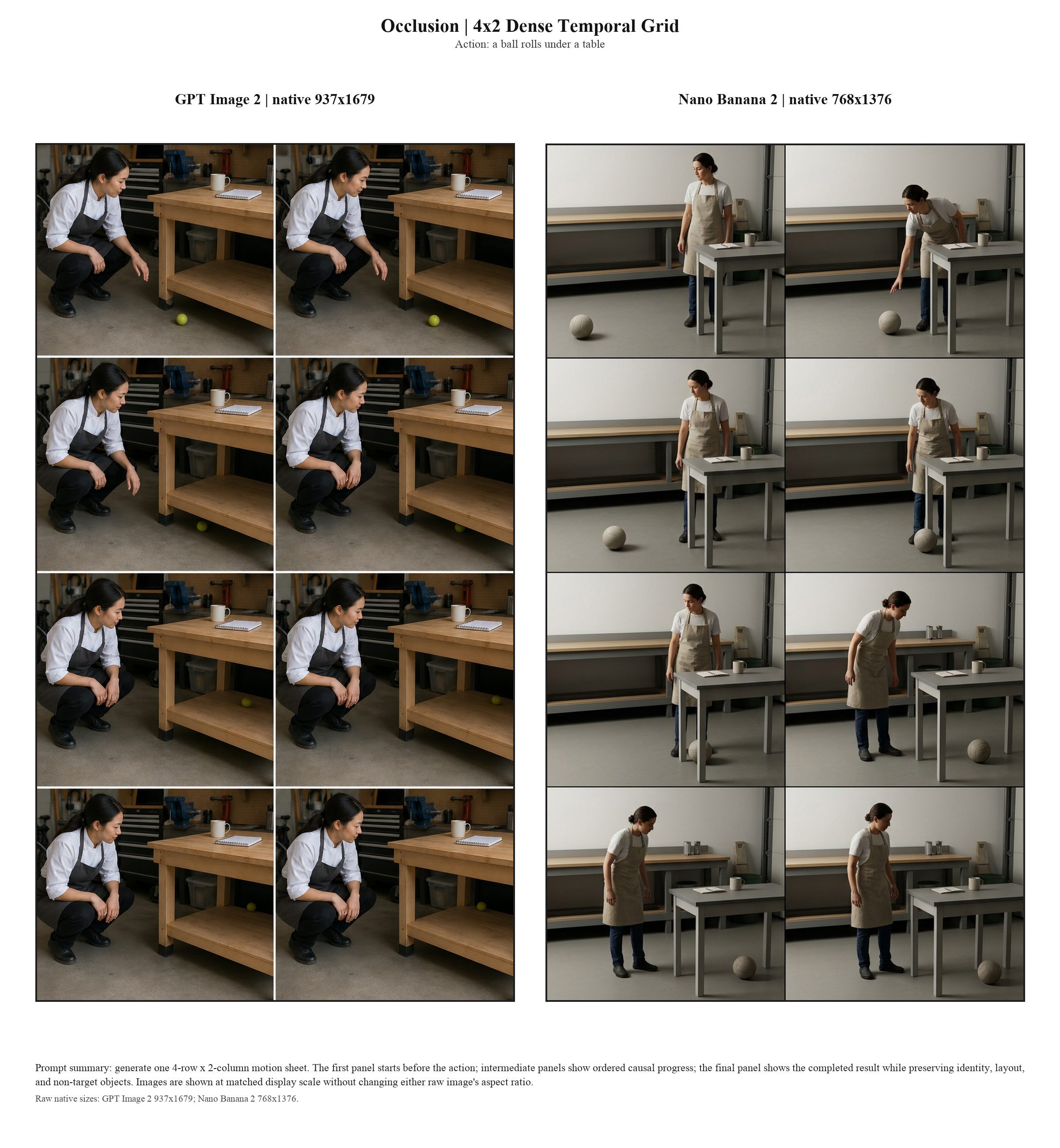}}\par
    \captionof{figure}{Dense-grid qualitative comparison for an Occlusion prompt using a 4x2 temporal grid, i.e., four rows with two panels per row.}
    \label{fig:appendix-dense-grid-occlusion}
\end{center}
\clearpage

\begin{center}
    \makebox[\textwidth][c]{\includegraphics[width=1.02\textwidth,height=0.84\textheight,keepaspectratio]{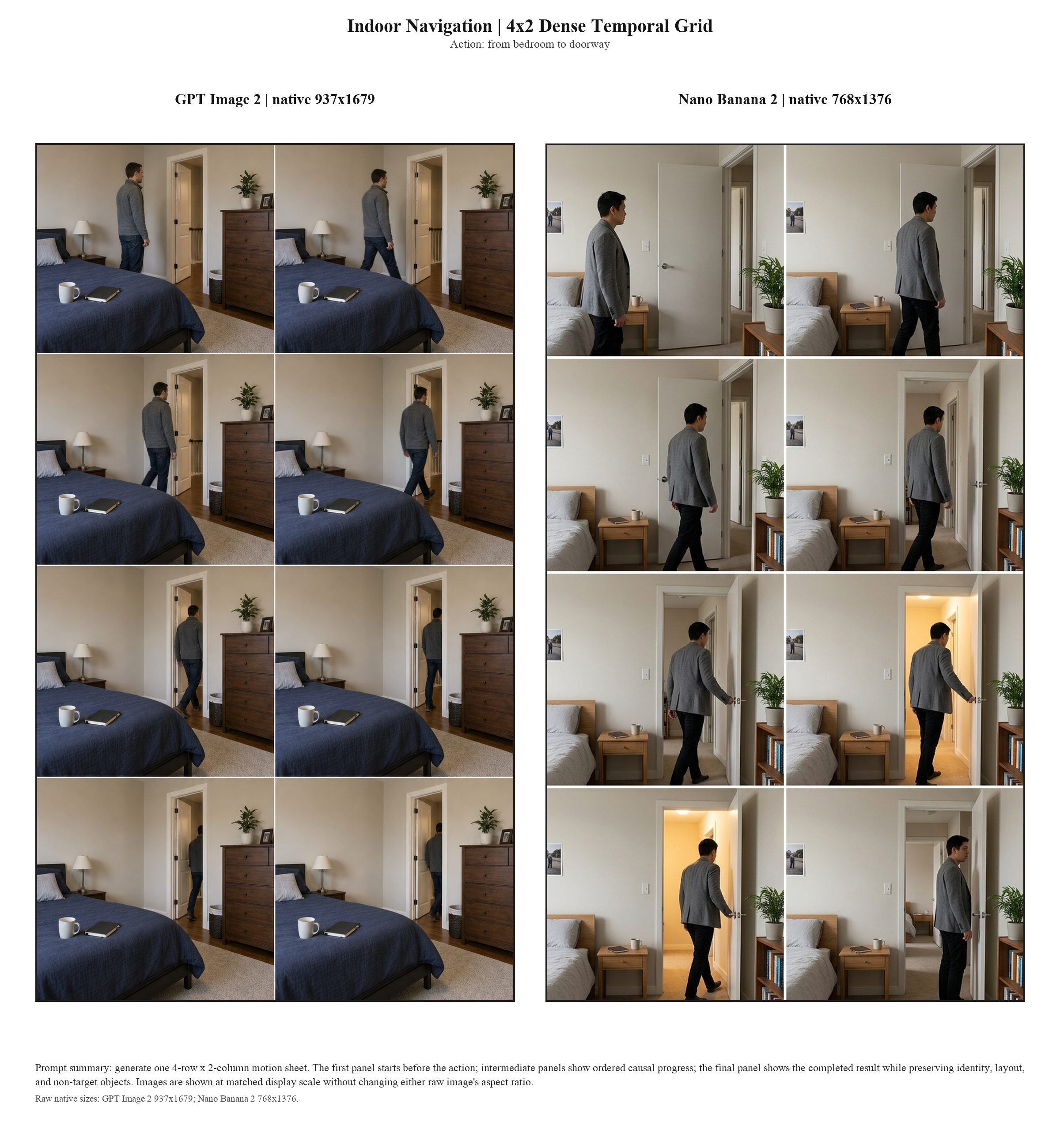}}\par
    \captionof{figure}{Dense-grid qualitative comparison for an Indoor-navigation prompt using a 4x2 temporal grid, i.e., four rows with two panels per row.}
    \label{fig:appendix-dense-grid-indoor-navigation}
\end{center}
\clearpage

\begin{center}
    \makebox[\textwidth][c]{\includegraphics[width=1.02\textwidth,height=0.84\textheight,keepaspectratio]{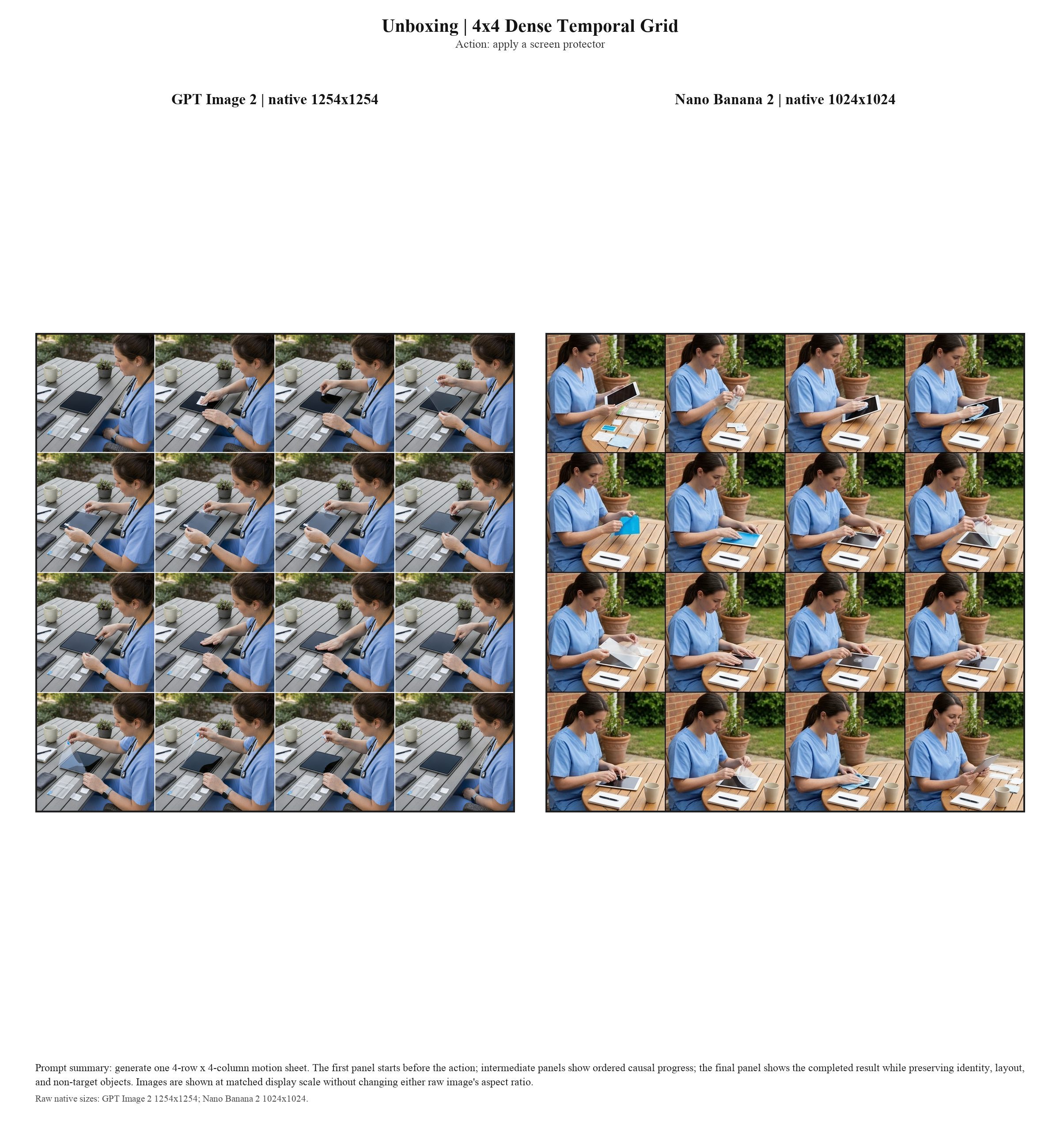}}\par
    \captionof{figure}{Dense-grid qualitative comparison for an Unboxing prompt using a 4x4 temporal grid, i.e., four rows with four panels per row.}
    \label{fig:appendix-dense-grid-unboxing}
\end{center}
\clearpage

\begin{center}
    \makebox[\textwidth][c]{\includegraphics[width=1.02\textwidth,height=0.84\textheight,keepaspectratio]{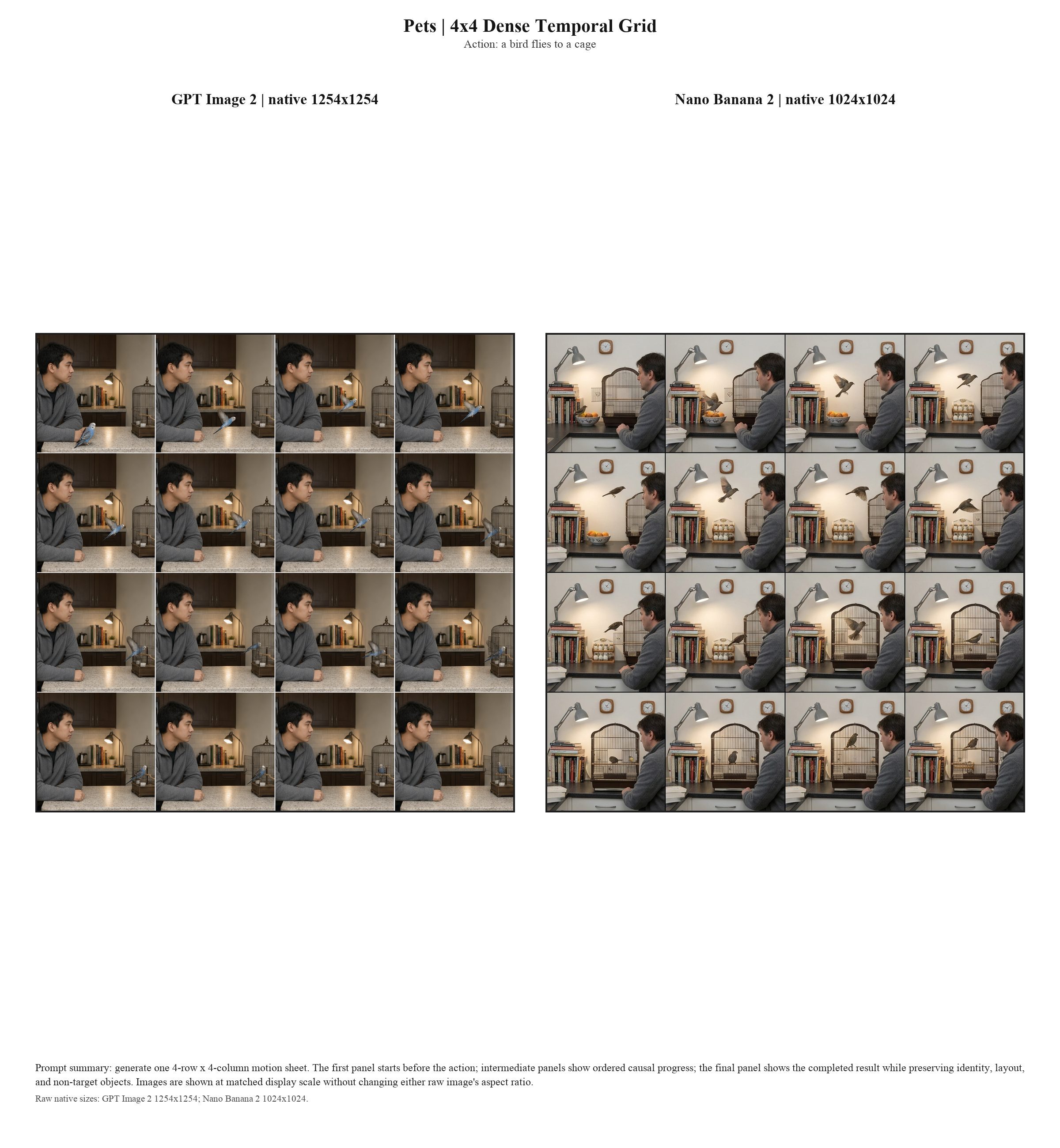}}\par
    \captionof{figure}{Dense-grid qualitative comparison for a Pets prompt using a 4x4 temporal grid, i.e., four rows with four panels per row.}
    \label{fig:appendix-dense-grid-pets}
\end{center}
\clearpage

\begin{center}
    \makebox[\textwidth][c]{\includegraphics[width=1.02\textwidth,height=0.84\textheight,keepaspectratio]{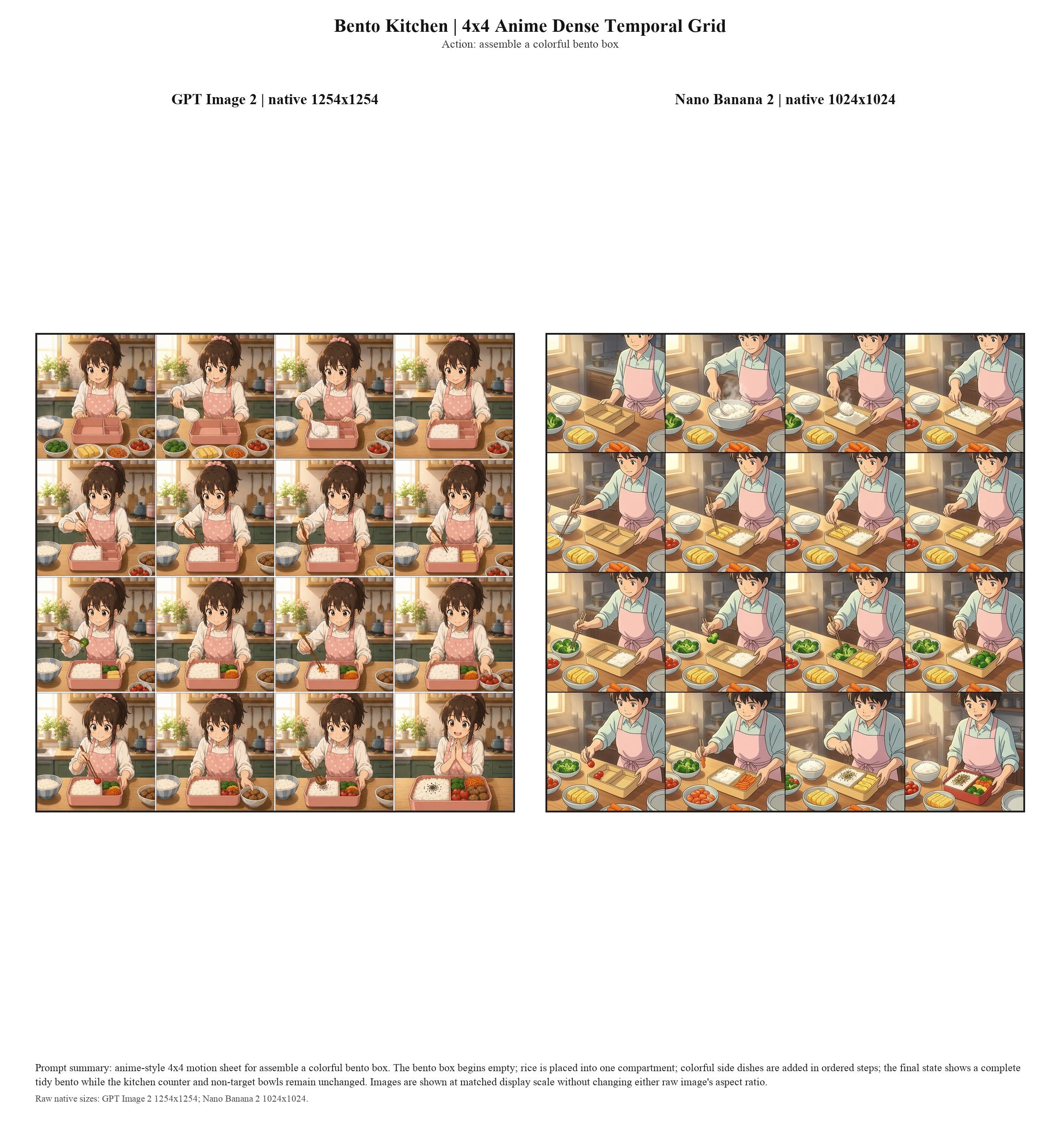}}\par
    \captionof{figure}{Anime-style dense-grid qualitative comparison for a Bento Kitchen prompt using a 4x4 temporal grid, i.e., four rows with four panels per row.}
    \label{fig:appendix-dense-grid-anime-bento-kitchen}
\end{center}
\clearpage

\begin{center}
    \makebox[\textwidth][c]{\includegraphics[width=1.02\textwidth,height=0.84\textheight,keepaspectratio]{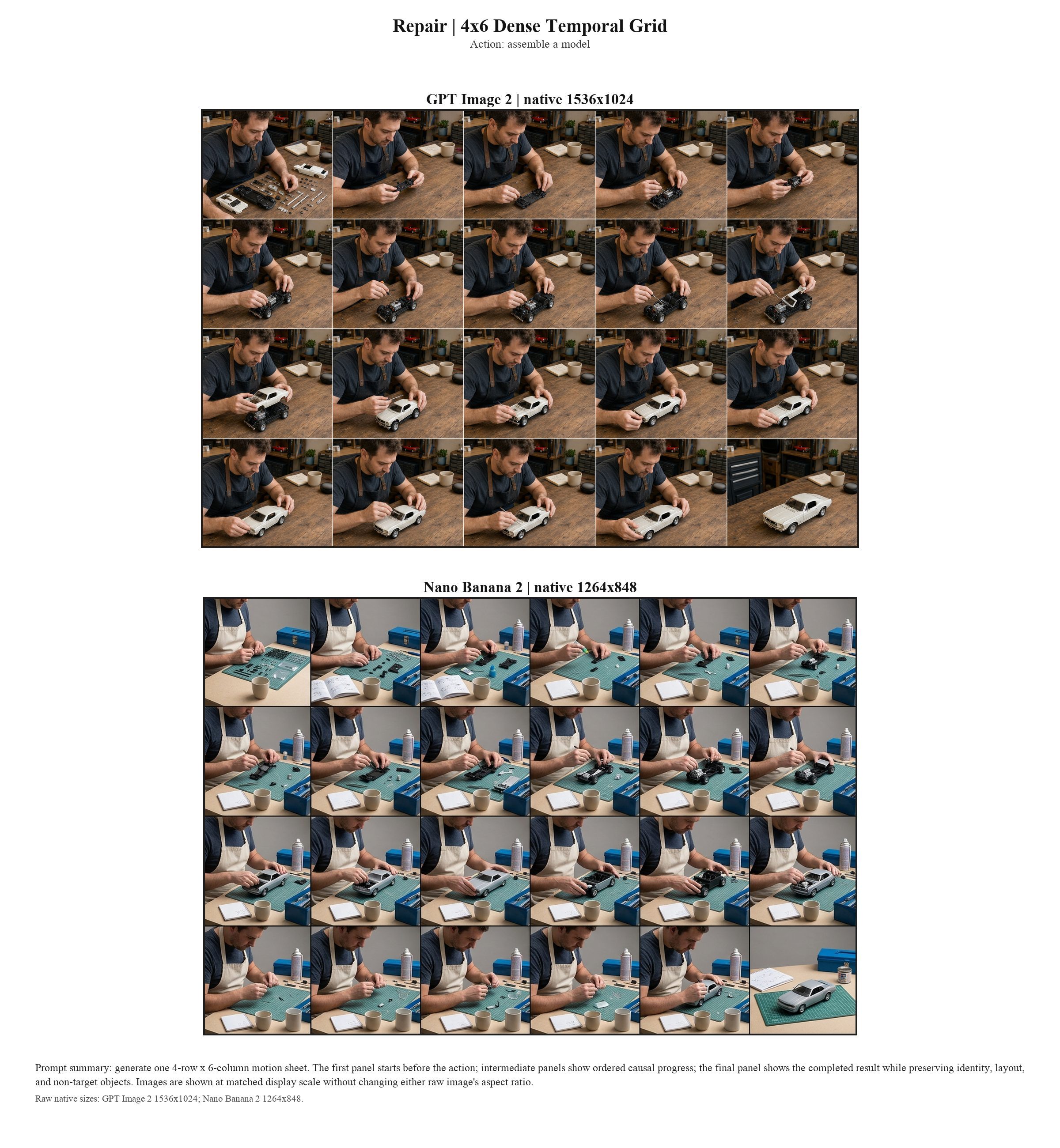}}\par
    \captionof{figure}{Dense-grid qualitative comparison for a Repair prompt using a 4x6 temporal grid, i.e., four rows with six panels per row.}
    \label{fig:appendix-dense-grid-repair}
\end{center}
\clearpage

\begin{center}
    \makebox[\textwidth][c]{\includegraphics[width=1.02\textwidth,height=0.84\textheight,keepaspectratio]{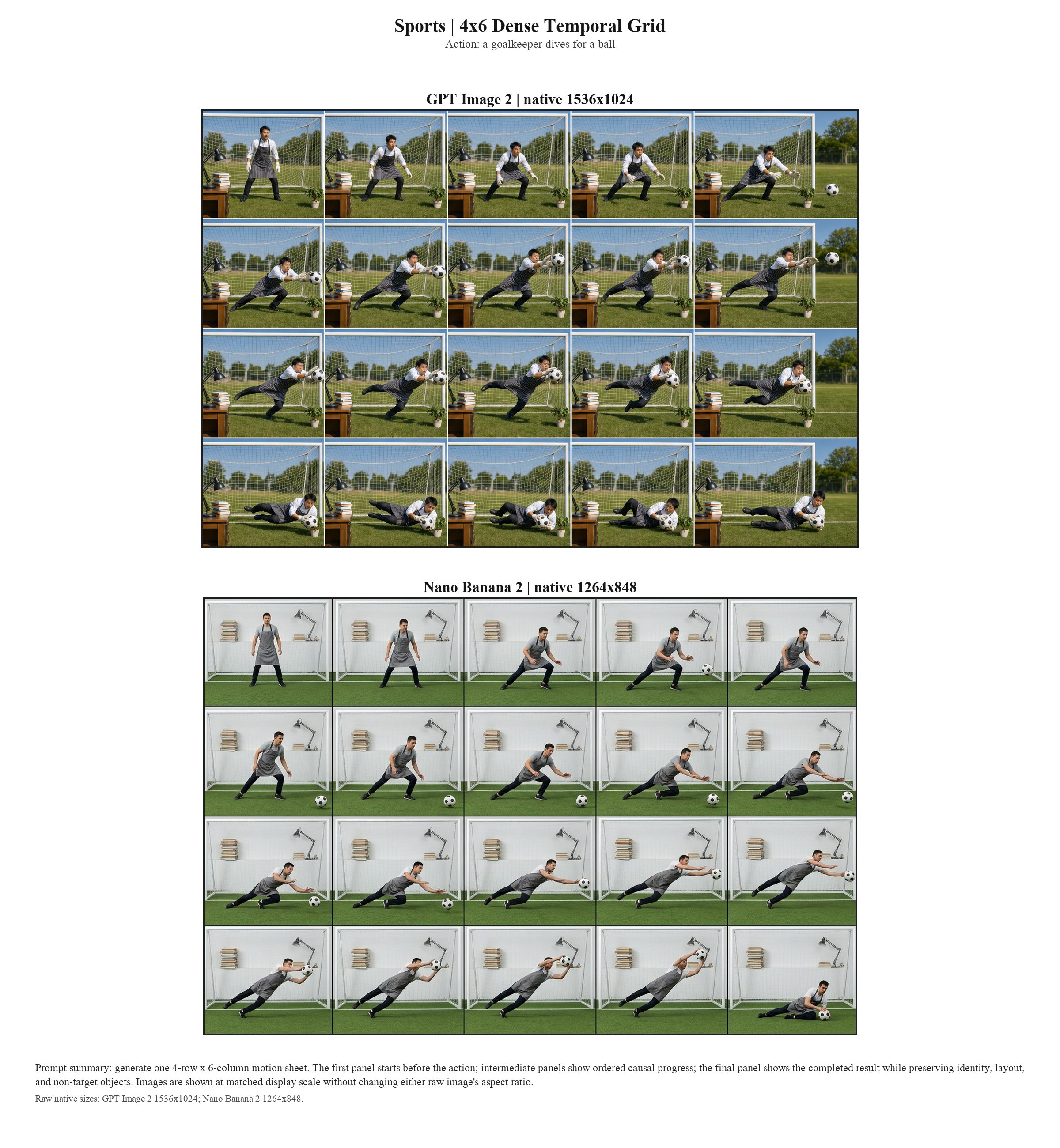}}\par
    \captionof{figure}{Dense-grid qualitative comparison for a Sports prompt using a 4x6 temporal grid, i.e., four rows with six panels per row.}
    \label{fig:appendix-dense-grid-sports}
\end{center}
\clearpage

\begin{center}
    \makebox[\textwidth][c]{\includegraphics[width=1.02\textwidth,height=0.84\textheight,keepaspectratio]{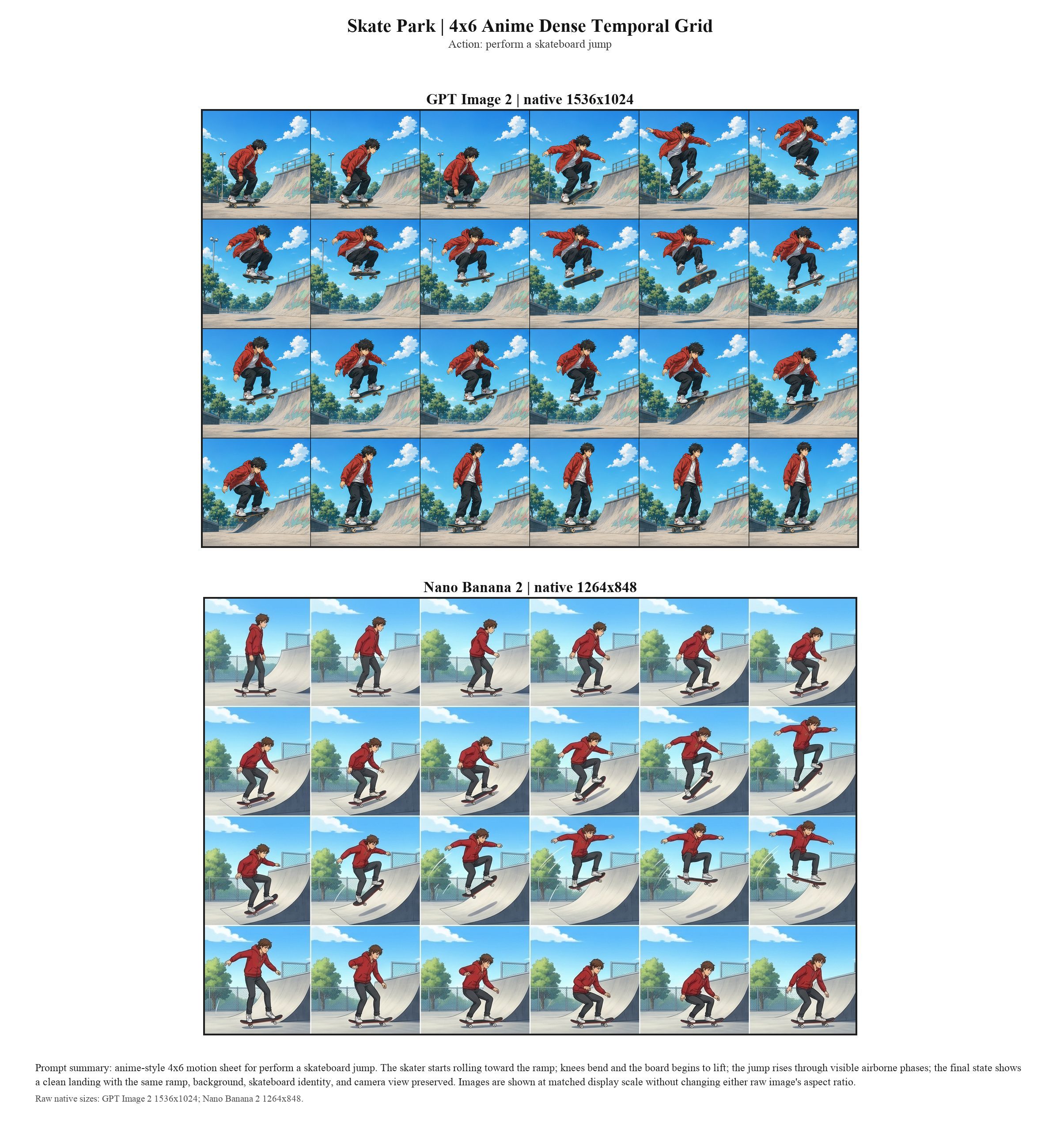}}\par
    \captionof{figure}{Anime-style dense-grid qualitative comparison for a Skate Park prompt using a 4x6 temporal grid, i.e., four rows with six panels per row.}
    \label{fig:appendix-dense-grid-anime-skate-park}
\end{center}
\clearpage

\begin{center}
    \makebox[\textwidth][c]{\includegraphics[width=1.02\textwidth,height=0.84\textheight,keepaspectratio]{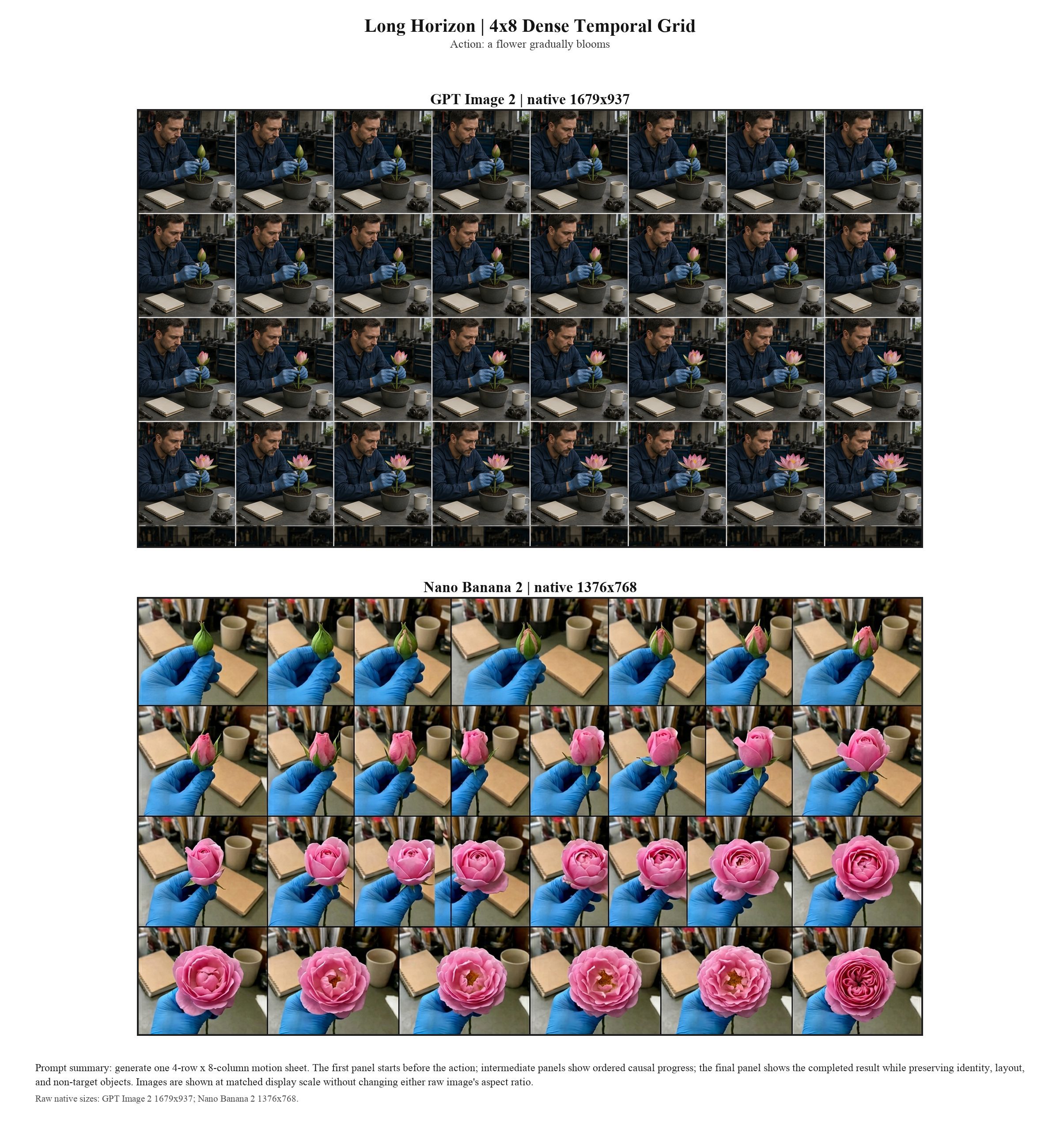}}\par
    \captionof{figure}{Dense-grid qualitative comparison for a Long-horizon prompt using a 4x8 temporal grid, i.e., four rows with eight panels per row.}
    \label{fig:appendix-dense-grid-long-horizon}
\end{center}
\clearpage

\begin{center}
    \makebox[\textwidth][c]{\includegraphics[width=1.02\textwidth,height=0.84\textheight,keepaspectratio]{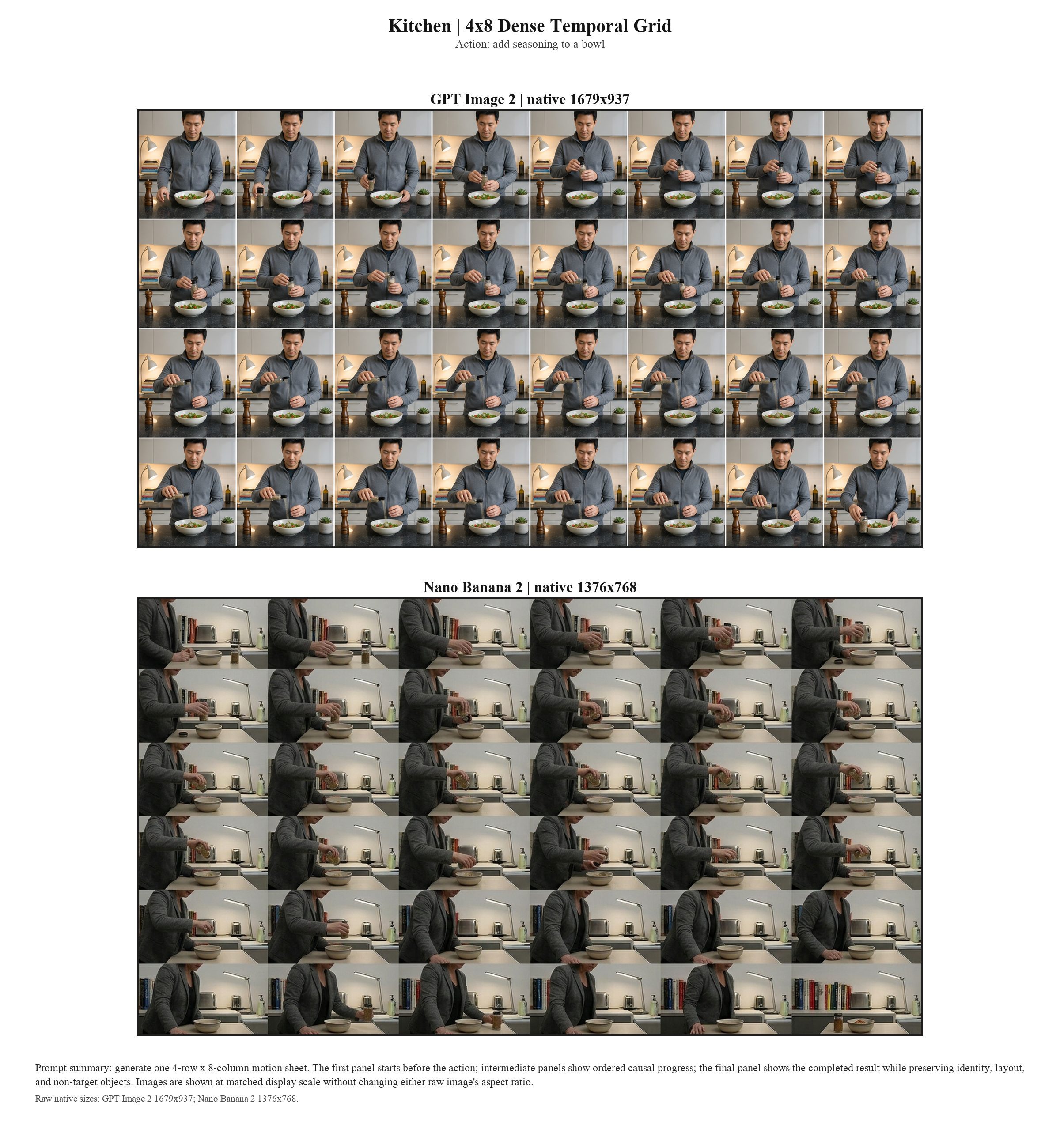}}\par
    \captionof{figure}{Dense-grid qualitative comparison for a Kitchen prompt using a 4x8 temporal grid, i.e., four rows with eight panels per row.}
    \label{fig:appendix-dense-grid-kitchen}
\end{center}
\clearpage

\begin{center}
    \makebox[\textwidth][c]{\includegraphics[width=1.02\textwidth,height=0.84\textheight,keepaspectratio]{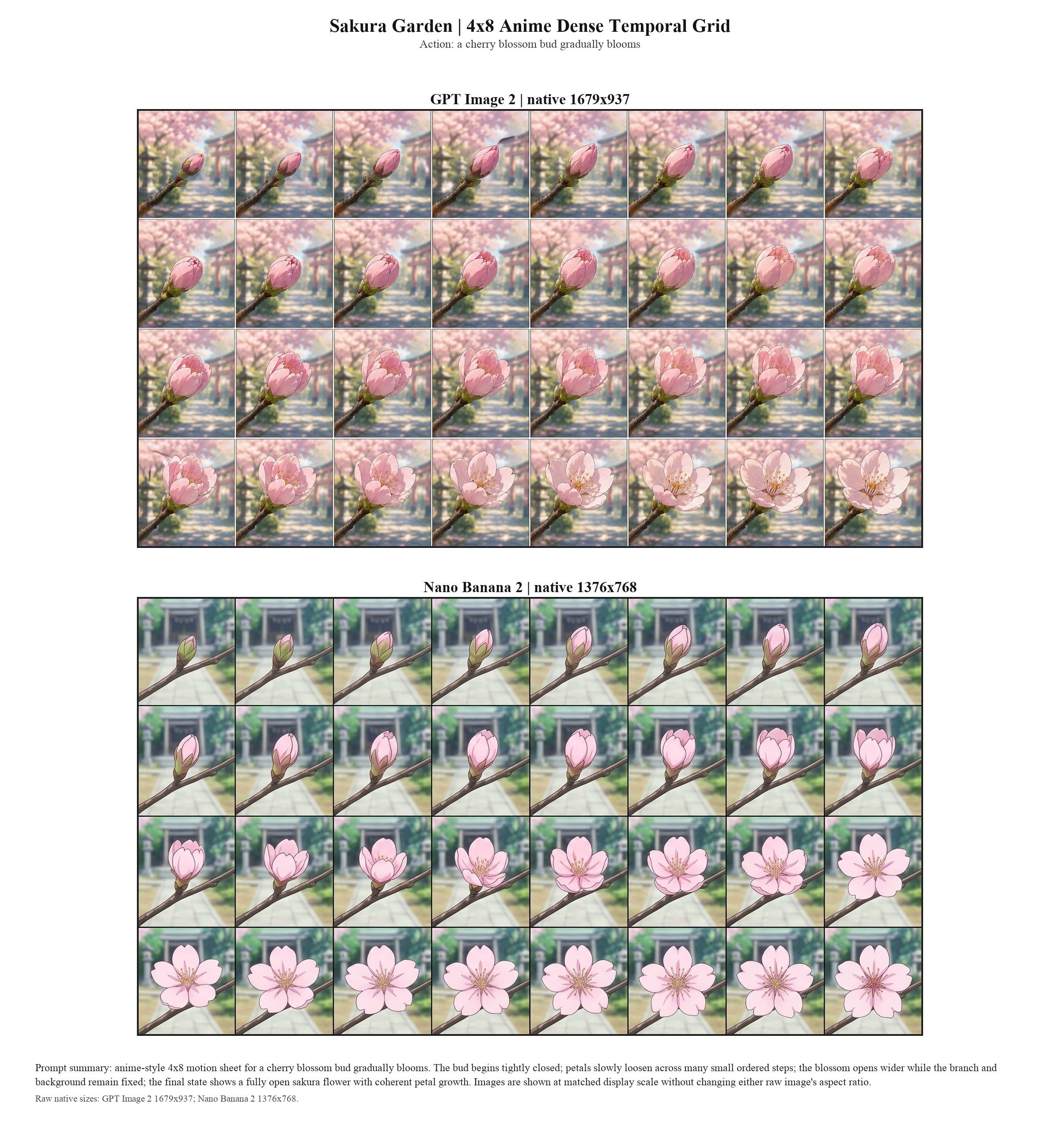}}\par
    \captionof{figure}{Anime-style dense-grid qualitative comparison for a Sakura Garden prompt using a 4x8 temporal grid, i.e., four rows with eight panels per row.}
    \label{fig:appendix-dense-grid-anime-sakura-garden}
\end{center}
\clearpage

\section{ImageTime Action Inventory}
\label{sec:appendix-action-inventory}
Table~\ref{tab:appendix-action-concepts} lists the complete ImageTime action-concept inventory. The benchmark contains 375 action concepts across 22 domains; each concept is instantiated with two semantically related but visually distinct variants, yielding the 750 prompt-only cases used in the main experiments. The table reports the domain, subcategory, action concept, and difficulty label for each action concept.

\begingroup
\scriptsize
\setlength{\tabcolsep}{3pt}
\renewcommand{\arraystretch}{1.06}
\begin{longtable}{@{}p{0.16\textwidth}p{0.23\textwidth}p{0.48\textwidth}p{0.08\textwidth}@{}}
\caption{Complete ImageTime action-concept inventory. Each row is one of the 375 action concepts; each concept is instantiated with two variants to construct the 750-case benchmark.}\label{tab:appendix-action-concepts}\\
\toprule
Domain & Subcategory & Action concept & Diff. \\
\midrule
\endfirsthead
\caption[]{Complete ImageTime action-concept inventory (continued).}\\
\toprule
Domain & Subcategory & Action concept & Diff. \\
\midrule
\endhead
\midrule
\multicolumn{4}{r@{}}{Continued on next page}\\
\endfoot
\bottomrule
\endlastfoot
\multicolumn{4}{@{}l}{\textbf{Animation (21 action concepts)}}\\
Animation & Character action & a cartoon character chases a balloon & easy \\
Animation & Narrative segment & a character chases another character & easy \\
Animation & Narrative segment & a character gives a map to a companion & easy \\
Animation & Character action & a character walks & easy \\
Animation & Object interaction & character open magic book & medium \\
Animation & Narrative segment & discover a hidden object & easy \\
Animation & Character action & eat food & easy \\
Animation & Narrative segment & exchange items & easy \\
Animation & Object interaction & hand over an item & hard \\
Animation & Character action & jump & easy \\
Animation & Object interaction & open door & medium \\
Animation & Object interaction & open treasure chest & medium \\
Animation & Object interaction & pick up an item & medium \\
Animation & Character action & pick up an object & easy \\
Animation & Object interaction & pick up magic wand & medium \\
Animation & Object interaction & pour potion & medium \\
Animation & Character action & put on clothes & easy \\
Animation & Character action & transform & easy \\
Animation & Narrative segment & transition between scenes & easy \\
Animation & Narrative segment & two people talk & easy \\
Animation & Character action & wave & easy \\
\multicolumn{4}{@{}l}{\textbf{Caregiving (13 action concepts)}}\\
Caregiving & Basic caregiving & apply a bandage & easy \\
Caregiving & Basic caregiving & cover a patient with a blanket & easy \\
Caregiving & Medicine and medical tool operation & disinfect a wound & medium \\
Caregiving & Basic caregiving & help a person sit down & medium \\
Caregiving & Basic caregiving & measure body temperature & easy \\
Caregiving & Medicine and medical tool operation & open bandage packaging & medium \\
Caregiving & Medicine and medical tool operation & open medicine box & medium \\
Caregiving & Basic caregiving & push wheelchair & easy \\
Caregiving & Medicine and medical tool operation & put medicine into a cup & medium \\
Caregiving & Basic caregiving & put on gloves & easy \\
Caregiving & Basic caregiving & remove mask & easy \\
Caregiving & Medicine and medical tool operation & take medicine & medium \\
Caregiving & Medicine and medical tool operation & use a stethoscope & medium \\
\multicolumn{4}{@{}l}{\textbf{Clothing (14 action concepts)}}\\
Clothing & Pose and appearance adjustment & close umbrella & easy \\
Clothing & Pose and appearance adjustment & open umbrella & easy \\
Clothing & Pose and appearance adjustment & organize collar & easy \\
Clothing & Pose and appearance adjustment & organize cuffs & easy \\
Clothing & Clothing change & put on coat & easy \\
Clothing & Clothing change & put on glasses & easy \\
Clothing & Clothing change & put on gloves & easy \\
Clothing & Clothing change & put on scarf & easy \\
Clothing & Clothing change & remove mask & easy \\
Clothing & Pose and appearance adjustment & roll up sleeves & easy \\
Clothing & Clothing change & take off hat & easy \\
Clothing & Pose and appearance adjustment & tie apron & easy \\
Clothing & Clothing change & tie shoelaces & easy \\
Clothing & Clothing change & wear backpack & easy \\
\multicolumn{4}{@{}l}{\textbf{Constraints (12 action concepts)}}\\
Constraints & Selective operation & do not touch cup on the table & easy \\
Constraints & Counterfactual change & if the ball does not hit the wall, its direction stays unchanged & easy \\
Constraints & Counterfactual change & if the cup is not picked up, it remains on the table & easy \\
Constraints & Counterfactual change & if the plant is not watered, it stays dry & easy \\
Constraints & Counterfactual change & if the switch is not pressed, the light stays off & easy \\
Constraints & Selective operation & open only left box & medium \\
Constraints & Selective operation & press only the green button and not the red button & easy \\
Constraints & Selective operation & take only the red ball and not the blue ball & easy \\
Constraints & Conditional blocking & the box cannot open so the object remains inside & medium \\
Constraints & Conditional blocking & the bridge is broken so the person stops & medium \\
Constraints & Conditional blocking & the cup lid is tightened so water cannot pour out & easy \\
Constraints & Conditional blocking & the door is locked so the person cannot enter & medium \\
\multicolumn{4}{@{}l}{\textbf{Outdoor Daily (13 action concepts)}}\\
Outdoor Daily & Environmental interaction & carry box & medium \\
Outdoor Daily & Environmental interaction & close umbrella & easy \\
Outdoor Daily & Path movement & cross road & medium \\
Outdoor Daily & Environmental interaction & fish with a fishing rod & easy \\
Outdoor Daily & Environmental interaction & open umbrella & easy \\
Outdoor Daily & Path movement & push stroller & medium \\
Outdoor Daily & Path movement & ride a bicycle & medium \\
Outdoor Daily & Path movement & run toward bus stop & medium \\
Outdoor Daily & Environmental interaction & set up a camping tent & easy \\
Outdoor Daily & Environmental interaction & spread a picnic mat on grass & easy \\
Outdoor Daily & Path movement & walk along a park path & medium \\
Outdoor Daily & Path movement & walk along a sidewalk & medium \\
Outdoor Daily & Environmental interaction & wash a car & easy \\
\multicolumn{4}{@{}l}{\textbf{Games (19 action concepts)}}\\
Games & Sandbox building & Minecraft place blocks & medium \\
Games & Action and parkour & a character picks up a key and opens a door & easy \\
Games & Strategy and multiplayer & a character protects a teammate while retreating & easy \\
Games & Action and parkour & a character runs & easy \\
Games & Strategy and multiplayer & attack a minion & easy \\
Games & Action and parkour & avoid an obstacle & easy \\
Games & Strategy and multiplayer & base build & easy \\
Games & Strategy and multiplayer & cast a skill & easy \\
Games & Strategy and multiplayer & character die or revive & easy \\
Games & Action and parkour & climb & easy \\
Games & Sandbox building & craft an item & medium \\
Games & Sandbox building & crops grow & medium \\
Games & Strategy and multiplayer & hero move & medium \\
Games & Action and parkour & jump & easy \\
Games & Sandbox building & open a chest & medium \\
Games & Action and parkour & open a door and enter a room & easy \\
Games & Sandbox building & open blocks & medium \\
Games & Action and parkour & pick up an item & easy \\
Games & Sandbox building & stack three blocks into a pillar & medium \\
\multicolumn{4}{@{}l}{\textbf{Gardening (12 action concepts)}}\\
Gardening & Environment arrangement & arrange flower pots in a row & medium \\
Gardening & Environment arrangement & clear fallen leaves & medium \\
Gardening & Plant operation & dig soil & medium \\
Gardening & Environment arrangement & organize flower pots & medium \\
Gardening & Plant operation & pick fruit from a plant & medium \\
Gardening & Plant operation & plant seeds & medium \\
Gardening & Plant operation & prune a branch & medium \\
Gardening & Environment arrangement & set up plant support & medium \\
Gardening & Environment arrangement & shovel snow & medium \\
Gardening & Plant operation & transplant a seedling into a flower pot & medium \\
Gardening & Plant operation & transplant plant & medium \\
Gardening & Plant operation & water flowers & medium \\
\multicolumn{4}{@{}l}{\textbf{Household (30 action concepts)}}\\
Household & Container opening and closing & close a door & easy \\
Household & Container opening and closing & close a suitcase & easy \\
Household & Cleaning and organizing & fold clothes & easy \\
Household & Cleaning and organizing & make a bed & easy \\
Household & Cleaning and organizing & mop the floor & easy \\
Household & Container opening and closing & open a box & easy \\
Household & Container opening and closing & open a cabinet door & easy \\
Household & Container opening and closing & open a drawer & easy \\
Household & Container opening and closing & open a refrigerator & easy \\
Household & Container opening and closing & open a wardrobe door & easy \\
Household & Object retrieval and placement & place a cup on a table & easy \\
Household & Switch and device control & press a doorbell & easy \\
Household & Container opening and closing & pull open curtains & easy \\
Household & Object retrieval and placement & put a book into a backpack & easy \\
Household & Object retrieval and placement & put a remote control back into a drawer & easy \\
Household & Object retrieval and placement & put a toy into a box & easy \\
Household & Cleaning and organizing & put scattered books back on a shelf & easy \\
Household & Cleaning and organizing & sweep the floor & easy \\
Household & Object retrieval and placement & take a book from a shelf & easy \\
Household & Object retrieval and placement & take an object from a drawer & easy \\
Household & Cleaning and organizing & take out trash & easy \\
Household & Cleaning and organizing & tidy up toys & easy \\
Household & Switch and device control & turn off a fan & easy \\
Household & Switch and device control & turn off a light & easy \\
Household & Switch and device control & turn off an air conditioner & easy \\
Household & Switch and device control & turn on a desk lamp switch & easy \\
Household & Switch and device control & turn on a faucet & easy \\
Household & Switch and device control & turn on a light & easy \\
Household & Switch and device control & turn on a television & easy \\
Household & Cleaning and organizing & wipe a table & easy \\
\multicolumn{4}{@{}l}{\textbf{Kitchen (29 action concepts)}}\\
Kitchen & Pouring and mixing & add seasoning to a bowl & easy \\
Kitchen & Cooking state change & boil noodles & easy \\
Kitchen & Cooking state change & bring water to a boil & easy \\
Kitchen & Tableware and container operation & close a pot lid & easy \\
Kitchen & Cutting and preparation & crack an egg & medium \\
Kitchen & Cutting and preparation & cut a carrot & medium \\
Kitchen & Cutting and preparation & cut fruit & medium \\
Kitchen & Cutting and preparation & cut vegetables & medium \\
Kitchen & Cooking state change & fry an egg & easy \\
Kitchen & Cooking state change & melt butter & easy \\
Kitchen & Cooking state change & melt cheese & easy \\
Kitchen & Cooking state change & melt ice & easy \\
Kitchen & Cutting and preparation & open a can & medium \\
Kitchen & Tableware and container operation & open a microwave & easy \\
Kitchen & Tableware and container operation & open a pot lid & easy \\
Kitchen & Cutting and preparation & peel an apple & medium \\
Kitchen & Pouring and mixing & pour beaten egg into a bowl & easy \\
Kitchen & Pouring and mixing & pour flour & easy \\
Kitchen & Pouring and mixing & pour ingredients into a pan & easy \\
Kitchen & Pouring and mixing & pour milk & easy \\
Kitchen & Pouring and mixing & pour oil & easy \\
Kitchen & Pouring and mixing & pour water & easy \\
Kitchen & Tableware and container operation & put a bowl into a cupboard & easy \\
Kitchen & Tableware and container operation & put a cup into a cupboard & easy \\
Kitchen & Tableware and container operation & put a plate into a sink & easy \\
Kitchen & Cutting and preparation & slice bread & medium \\
Kitchen & Pouring and mixing & stir ingredients & easy \\
Kitchen & Cooking state change & stir-fry vegetables & easy \\
Kitchen & Cooking state change & toast bread & easy \\
\multicolumn{4}{@{}l}{\textbf{Education \& Lab (12 action concepts)}}\\
Education \& Lab & Circuit and equipment operation & adjust a microscope focus & easy \\
Education \& Lab & Circuit and equipment operation & connect a battery to a light bulb & easy \\
Education \& Lab & Lab experiment operation & dropper drop liquid & easy \\
Education \& Lab & Circuit and equipment operation & insert wire & easy \\
Education \& Lab & Lab experiment operation & light alcohol lamp & easy \\
Education \& Lab & Circuit and equipment operation & measure with a ruler & easy \\
Education \& Lab & Lab experiment operation & mix colors & easy \\
Education \& Lab & Circuit and equipment operation & open a book & easy \\
Education \& Lab & Lab experiment operation & pick up a sample with tweezers & easy \\
Education \& Lab & Lab experiment operation & pour liquid enter beaker & easy \\
Education \& Lab & Circuit and equipment operation & turn on a switch & easy \\
Education \& Lab & Lab experiment operation & weigh object & easy \\
\multicolumn{4}{@{}l}{\textbf{Long-horizon (10 action concepts)}}\\
Long-horizon & Gradual change & a flower gradually blooms & easy \\
Long-horizon & Cycle and phase & a machine progresses through processing stages & easy \\
Long-horizon & Cycle and phase & a traffic light changes from red to green & easy \\
Long-horizon & Cycle and phase & a traffic light cycles through red, yellow, and green & easy \\
Long-horizon & Gradual change & dough gradually takes shape & easy \\
Long-horizon & Gradual change & foam on coffee gradually disappears & easy \\
Long-horizon & Gradual change & ice gradually melts & easy \\
Long-horizon & Cycle and phase & the moon phase changes & easy \\
Long-horizon & Gradual change & the sky changes from daytime to dusk & easy \\
Long-horizon & Cycle and phase & the tide rises and falls & easy \\
\multicolumn{4}{@{}l}{\textbf{Machines (19 action concepts)}}\\
Machines & Home appliance operation & a coffee machine dispenses coffee & easy \\
Machines & Industrial assembly line & a conveyor belt moves & medium \\
Machines & Industrial assembly line & a conveyor belt moves a box under a scanner & hard \\
Machines & Home appliance operation & a juicer starts dispensing juice & easy \\
Machines & Home appliance operation & a printer outputs paper & easy \\
Machines & Robot arm operation & a robot arm presses a button & medium \\
Machines & Robot arm operation & a robot opens a door & easy \\
Machines & Home appliance operation & a vacuum cleaner cleans the floor & easy \\
Machines & Home appliance operation & a vending machine dispenses an item & easy \\
Machines & Home appliance operation & a washing machine drum rotates & easy \\
Machines & Industrial assembly line & apply label & easy \\
Machines & Industrial assembly line & assemble parts & easy \\
Machines & Industrial assembly line & bottles fill & easy \\
Machines & Robot arm operation & carry a box & medium \\
Machines & Robot arm operation & grab take part & easy \\
Machines & Robot arm operation & place an object & easy \\
Machines & Industrial assembly line & product inspect & easy \\
Machines & Robot arm operation & robot pour water & easy \\
Machines & Industrial assembly line & sort parcel & medium \\
\multicolumn{4}{@{}l}{\textbf{Nature (29 action concepts)}}\\
Nature & Physical process & a ball rolls down a slope & medium \\
Nature & Plant and animal change & a bird takes off & easy \\
Nature & Physical process & a candle melts & easy \\
Nature & Plant and animal change & a fish swims & easy \\
Nature & Plant and animal change & a flower blooms & easy \\
Nature & Landscape and scene process & a stream flows around stones & easy \\
Nature & Plant and animal change & a vine grows along a support & easy \\
Nature & Landscape and scene process & a volcano erupts & easy \\
Nature & Plant and animal change & an insect crawls & easy \\
Nature & Weather and natural phenomena & clouds move & easy \\
Nature & Weather and natural phenomena & fog forms & easy \\
Nature & Physical process & glass break & easy \\
Nature & Weather and natural phenomena & lightning & easy \\
Nature & Physical process & melt ice & easy \\
Nature & Plant and animal change & plant grow & easy \\
Nature & Physical process & pour liquid into a container & easy \\
Nature & Weather and natural phenomena & rain begins falling & easy \\
Nature & Landscape and scene process & river flow & easy \\
Nature & Landscape and scene process & sand flow & easy \\
Nature & Physical process & smoke diffuse & easy \\
Nature & Plant and animal change & snake move & easy \\
Nature & Weather and natural phenomena & snow begins falling & easy \\
Nature & Landscape and scene process & snow melts on a mountain & easy \\
Nature & Weather and natural phenomena & sunrise changes into sunset & easy \\
Nature & Weather and natural phenomena & the sun emerges from behind clouds & easy \\
Nature & Physical process & water freezes & easy \\
Nature & Landscape and scene process & waterfall fall & easy \\
Nature & Landscape and scene process & waves wash ashore & easy \\
Nature & Weather and natural phenomena & wind blows leaves & easy \\
\multicolumn{4}{@{}l}{\textbf{Indoor Navigation (11 action concepts)}}\\
Indoor Navigation & Indoor room navigation & from bedroom to doorway & easy \\
Indoor Navigation & Indoor room navigation & go around table & easy \\
Indoor Navigation & Approach and arrival & pick up table object & easy \\
Indoor Navigation & Approach and arrival & sit down on a sofa & easy \\
Indoor Navigation & Indoor room navigation & walk from a kitchen to a dining room & easy \\
Indoor Navigation & Indoor room navigation & walk from a living room to a kitchen & easy \\
Indoor Navigation & Indoor room navigation & walk through a hallway & easy \\
Indoor Navigation & Approach and arrival & walk to a window & easy \\
Indoor Navigation & Approach and arrival & walk toward a desk and stop & medium \\
Indoor Navigation & Approach and arrival & walk toward door & medium \\
Indoor Navigation & Indoor room navigation & walk toward sofa & easy \\
\multicolumn{4}{@{}l}{\textbf{Occlusion (10 action concepts)}}\\
Occlusion & Entering occlusion & a ball rolls under a table & medium \\
Occlusion & Entering occlusion & a cat crawls into a box & easy \\
Occlusion & Exiting occlusion & a child takes a toy from under a blanket & easy \\
Occlusion & Entering occlusion & a person walks behind a pillar & easy \\
Occlusion & Entering occlusion & a toy car drives under a sofa & medium \\
Occlusion & Exiting occlusion & pick up a ball from under a table & easy \\
Occlusion & Entering occlusion & put an object into a bag & easy \\
Occlusion & Exiting occlusion & take a book out of a bag & easy \\
Occlusion & Exiting occlusion & take an object out of a cabinet & easy \\
Occlusion & Exiting occlusion & walk out from behind a door & easy \\
\multicolumn{4}{@{}l}{\textbf{Pets (16 action concepts)}}\\
Pets & Pet movement & a bird flies to a cage & easy \\
Pets & Pet movement & a cat crawls into a box & medium \\
Pets & Pet object retrieval & a cat grabs a toy & easy \\
Pets & Pet movement & a cat jumps from a sofa to a rug & easy \\
Pets & Pet movement & a cat jumps onto a table & easy \\
Pets & Pet object retrieval & a dog brings a toy to its owner & easy \\
Pets & Pet movement & a dog enters through a door & easy \\
Pets & Pet object retrieval & a dog picks up a ball in its mouth & easy \\
Pets & Pet movement & a dog runs toward a ball & easy \\
Pets & Pet object retrieval & a dog takes a toy from its owner & easy \\
Pets & Human-pet interaction & a person brushes a cat & easy \\
Pets & Human-pet interaction & a person feeds a dog & easy \\
Pets & Human-pet interaction & a person throws a ball to a dog & easy \\
Pets & Human-pet interaction & a person walks a dog on a leash & easy \\
Pets & Pet object retrieval & a pet eats food from a bowl & easy \\
Pets & Human-pet interaction & put a pet into a cage & easy \\
\multicolumn{4}{@{}l}{\textbf{Quantity-focused (12 action concepts)}}\\
Quantity-focused & Group state & a row of people moves forward one by one & easy \\
Quantity-focused & Multi-object quantity & from five cup take away two & easy \\
Quantity-focused & Single-object quantity & from three pen take away one blue pen & easy \\
Quantity-focused & Group state & multiple bottles are filled one by one & easy \\
Quantity-focused & Single-object quantity & one ball put into basket & easy \\
Quantity-focused & Group state & one of three birds flies away & medium \\
Quantity-focused & Multi-object quantity & separate four red blocks and three blue blocks & easy \\
Quantity-focused & Single-object quantity & take away one pen & easy \\
Quantity-focused & Single-object quantity & take one apple out of a box & medium \\
Quantity-focused & Group state & the first three of six bottles are filled in order & easy \\
Quantity-focused & Multi-object quantity & three blocks arrange into a row & easy \\
Quantity-focused & Multi-object quantity & two players pass one ball & medium \\
\multicolumn{4}{@{}l}{\textbf{Repair (13 action concepts)}}\\
Repair & Assembly process & assemble a model & medium \\
Repair & Assembly process & assemble chair & medium \\
Repair & Tool use & hammer nail & easy \\
Repair & Tool use & inflate a tire & easy \\
Repair & Assembly process & install a table leg & medium \\
Repair & Assembly process & install light bulb & medium \\
Repair & Assembly process & open toolbox take tool & medium \\
Repair & Tool use & paint wall & easy \\
Repair & Tool use & repair chain & easy \\
Repair & Tool use & saw wood & easy \\
Repair & Assembly process & set up blocks & medium \\
Repair & Tool use & tighten a nut with a wrench & easy \\
Repair & Tool use & tighten screw & easy \\
\multicolumn{4}{@{}l}{\textbf{Social (13 action concepts)}}\\
Social & Multi-person collaboration & a line of people moves forward & medium \\
Social & Handoff and receiving & a server brings food to a table & medium \\
Social & Handoff and receiving & a teacher hands out homework & medium \\
Social & Multi-person collaboration & a team passes a ball & medium \\
Social & Handoff and receiving & hand keys to a friend & medium \\
Social & Handoff and receiving & hand over book & medium \\
Social & Handoff and receiving & hand over cup & medium \\
Social & Multi-person collaboration & multiple people move household items & medium \\
Social & Multi-person collaboration & people set up a tent together & medium \\
Social & Handoff and receiving & shake hands & medium \\
Social & Handoff and receiving & two people carry a box & medium \\
Social & Multi-person collaboration & two people carry a sofa together & medium \\
Social & Multi-person collaboration & two people lift a table & medium \\
\multicolumn{4}{@{}l}{\textbf{Sports (30 action concepts)}}\\
Sports & Ball sports & a goalkeeper dives for a ball & medium \\
Sports & Team interaction & basketball teammates perform a screen-and-roll & easy \\
Sports & Ball sports & catch ball & medium \\
Sports & Team interaction & coordinate a volleyball play & easy \\
Sports & Water and snow sports & dive into water & easy \\
Sports & Water and snow sports & dive underwater & easy \\
Sports & Individual sports & do a push-up & easy \\
Sports & Individual sports & high jump & easy \\
Sports & Ball sports & hit ball & medium \\
Sports & Water and snow sports & ice skate & easy \\
Sports & Individual sports & jump rope & easy \\
Sports & Ball sports & kick ball & medium \\
Sports & Individual sports & long jump & easy \\
Sports & Team interaction & multiple players defend together & easy \\
Sports & Ball sports & pass ball & medium \\
Sports & Team interaction & pass catch ball & medium \\
Sports & Individual sports & perform a yoga pose & easy \\
Sports & Team interaction & play doubles badminton & easy \\
Sports & Individual sports & ride a skateboard & easy \\
Sports & Individual sports & roller skate & easy \\
Sports & Individual sports & run & easy \\
Sports & Team interaction & run a relay race & easy \\
Sports & Ball sports & shoot a basketball & medium \\
Sports & Water and snow sports & ski & easy \\
Sports & Ball sports & smash a badminton shuttlecock & medium \\
Sports & Water and snow sports & surf on a wave & easy \\
Sports & Water and snow sports & swim & easy \\
Sports & Ball sports & table tennis ball catch ball & medium \\
Sports & Individual sports & throw boxing punches & easy \\
Sports & Water and snow sports & turn while skiing & easy \\
\multicolumn{4}{@{}l}{\textbf{Motion Systems (19 action concepts)}}\\
Motion Systems & Vehicles and roads & a bus enters a bus stop & easy \\
Motion Systems & Vehicles and roads & a car drives into a parking space & easy \\
Motion Systems & Boats and aircraft & a small boat goes around a buoy & easy \\
Motion Systems & Public facilities & a train enters a station & medium \\
Motion Systems & Boats and aircraft & an airplane taxis on a runway & easy \\
Motion Systems & Public facilities & an escalator runs & medium \\
Motion Systems & Boats and aircraft & boat dock & easy \\
Motion Systems & Boats and aircraft & boat turn & easy \\
Motion Systems & Vehicles and roads & change lanes while driving & easy \\
Motion Systems & Boats and aircraft & drone land & easy \\
Motion Systems & Boats and aircraft & drone take off & easy \\
Motion Systems & Public facilities & elevator doors open & medium \\
Motion Systems & Boats and aircraft & helicopter hover & easy \\
Motion Systems & Vehicles and roads & park a car & easy \\
Motion Systems & Public facilities & pass through a turnstile & medium \\
Motion Systems & Public facilities & passengers exit an elevator after it arrives & medium \\
Motion Systems & Vehicles and roads & pedestrian cross road & easy \\
Motion Systems & Vehicles and roads & reverse a car into a parking space & easy \\
Motion Systems & Public facilities & subway train doors open & medium \\
\multicolumn{4}{@{}l}{\textbf{Unboxing (18 action concepts)}}\\
Unboxing & Assembly and installation & apply a screen protector & easy \\
Unboxing & Assembly and installation & assemble a stand & easy \\
Unboxing & Assembly and installation & connect a cable & easy \\
Unboxing & Object presentation & display a toy & medium \\
Unboxing & Object presentation & display accessories & medium \\
Unboxing & Assembly and installation & insert a battery & easy \\
Unboxing & Assembly and installation & install a part & easy \\
Unboxing & Assembly and installation & install a toy car wheel & easy \\
Unboxing & Object presentation & lay out an instruction manual & medium \\
Unboxing & Package opening & open a gift box & easy \\
Unboxing & Package opening & open a product box & easy \\
Unboxing & Package opening & open a shipping box & medium \\
Unboxing & Package opening & open a shoe box & easy \\
Unboxing & Package opening & remove plastic wrap & easy \\
Unboxing & Object presentation & take out a phone & medium \\
Unboxing & Object presentation & take out a product & medium \\
Unboxing & Object presentation & take out an earbud case & medium \\
Unboxing & Package opening & tear open packaging & easy \\
\end{longtable}
\endgroup

\FloatBarrier

\clearpage
\section{Example Case Gallery}
\label{sec:appendix-example-case-gallery}

\begin{center}
    \makebox[\textwidth][c]{\includegraphics[width=1.08\textwidth,height=0.385\textheight,keepaspectratio]{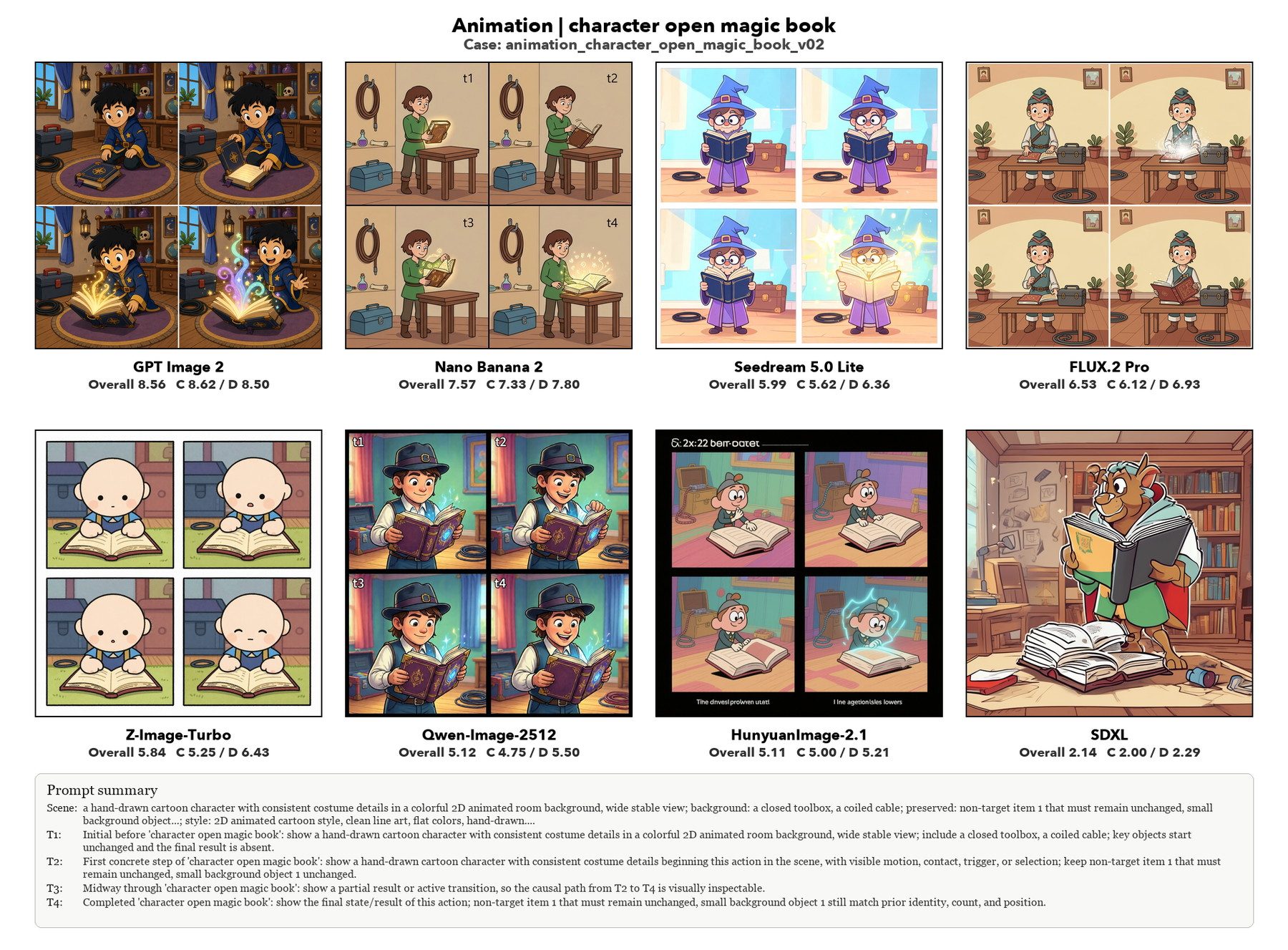}}\par
    \vspace{0.4em}
    \makebox[\textwidth][c]{\includegraphics[width=1.08\textwidth,height=0.385\textheight,keepaspectratio]{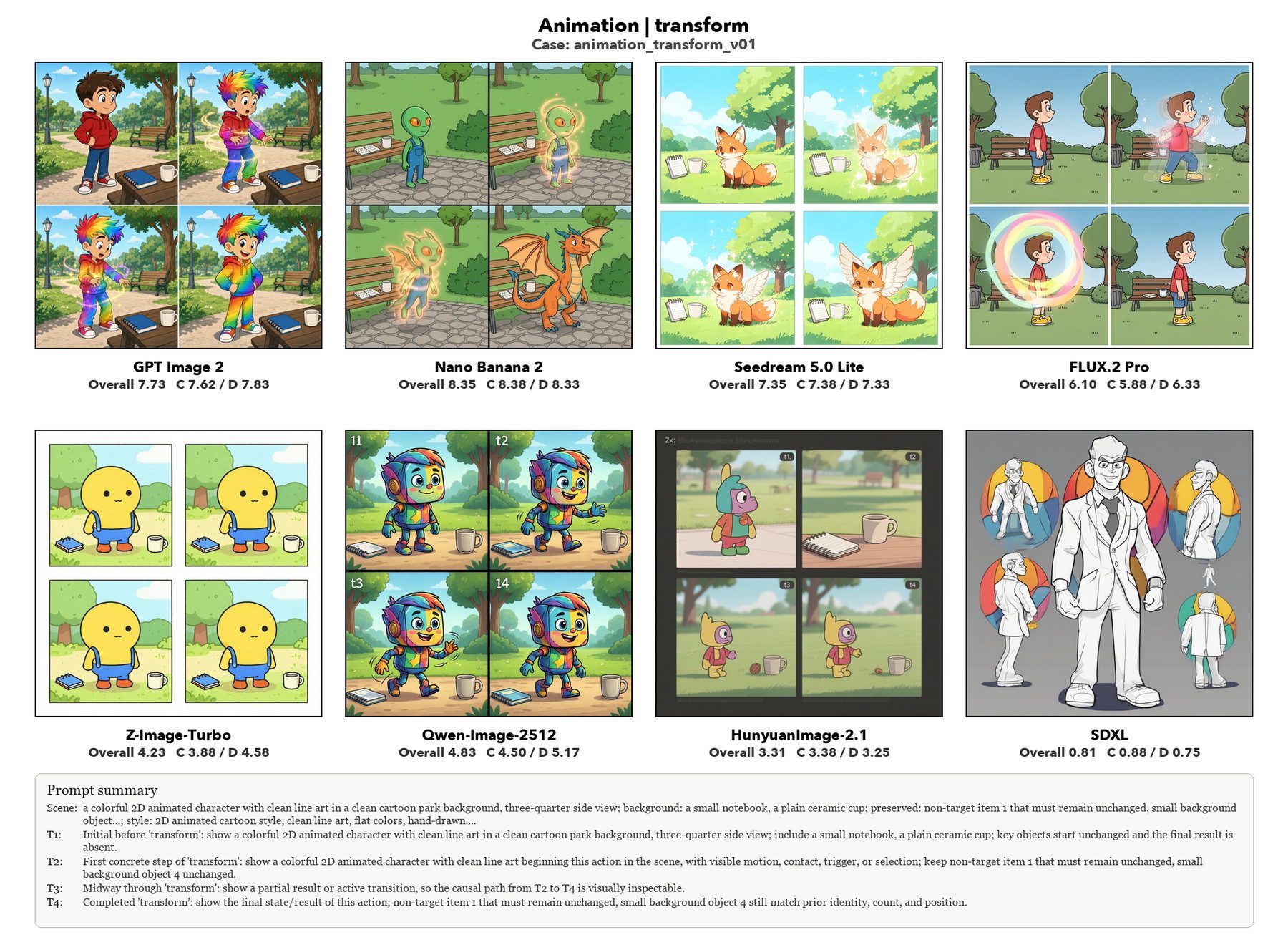}}\par
    \captionof{figure}{Example summary grids from the Animation category. Each grid compares the same prompt across eight image-generation models and reports the overall mean score below each generated motion sheet.}
    \label{fig:appendix-examples-animation}
\end{center}
\clearpage

\begin{center}
    \makebox[\textwidth][c]{\includegraphics[width=1.08\textwidth,height=0.435\textheight,keepaspectratio]{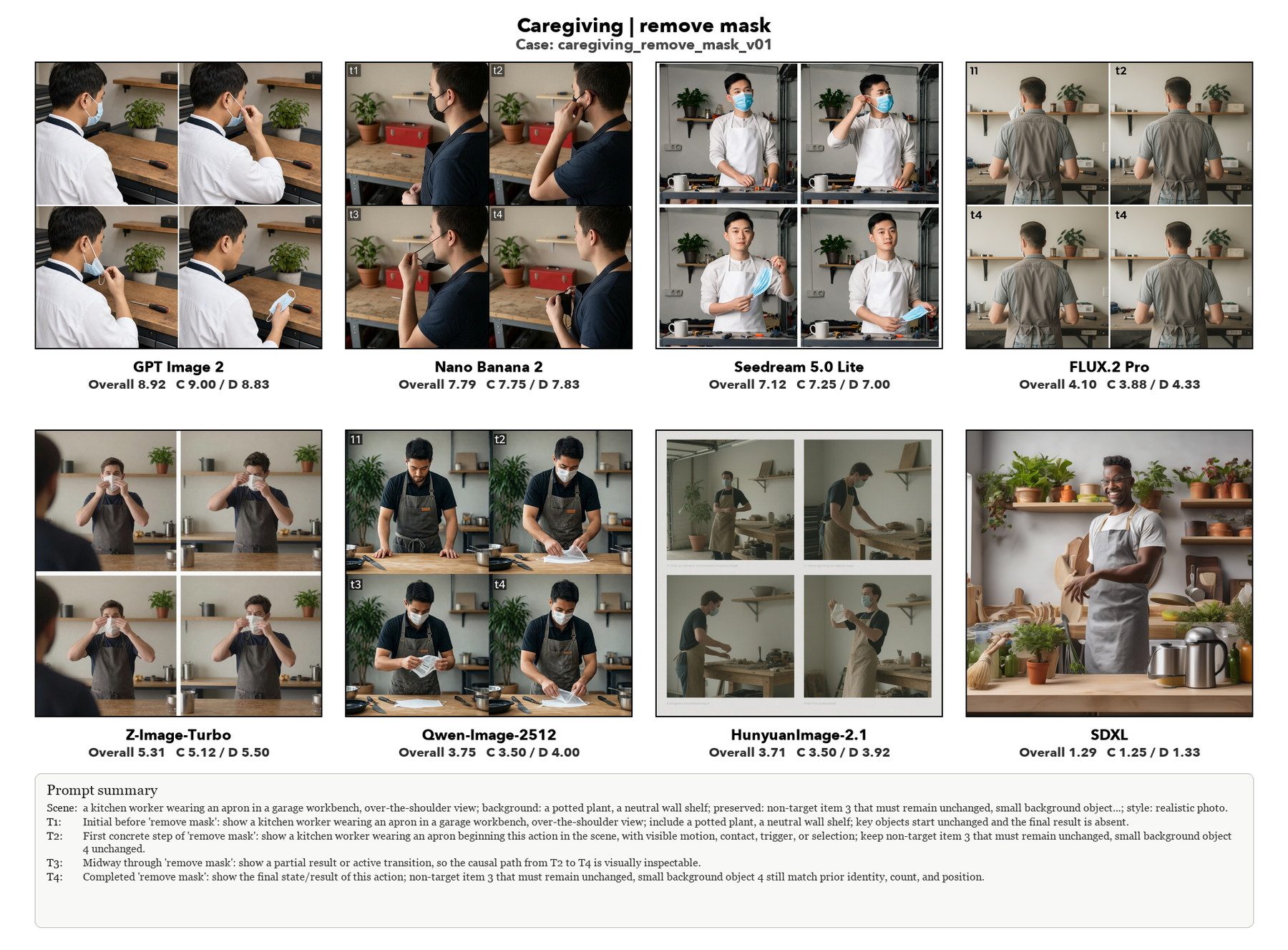}}\par
    \vspace{0.4em}
    \makebox[\textwidth][c]{\includegraphics[width=1.08\textwidth,height=0.435\textheight,keepaspectratio]{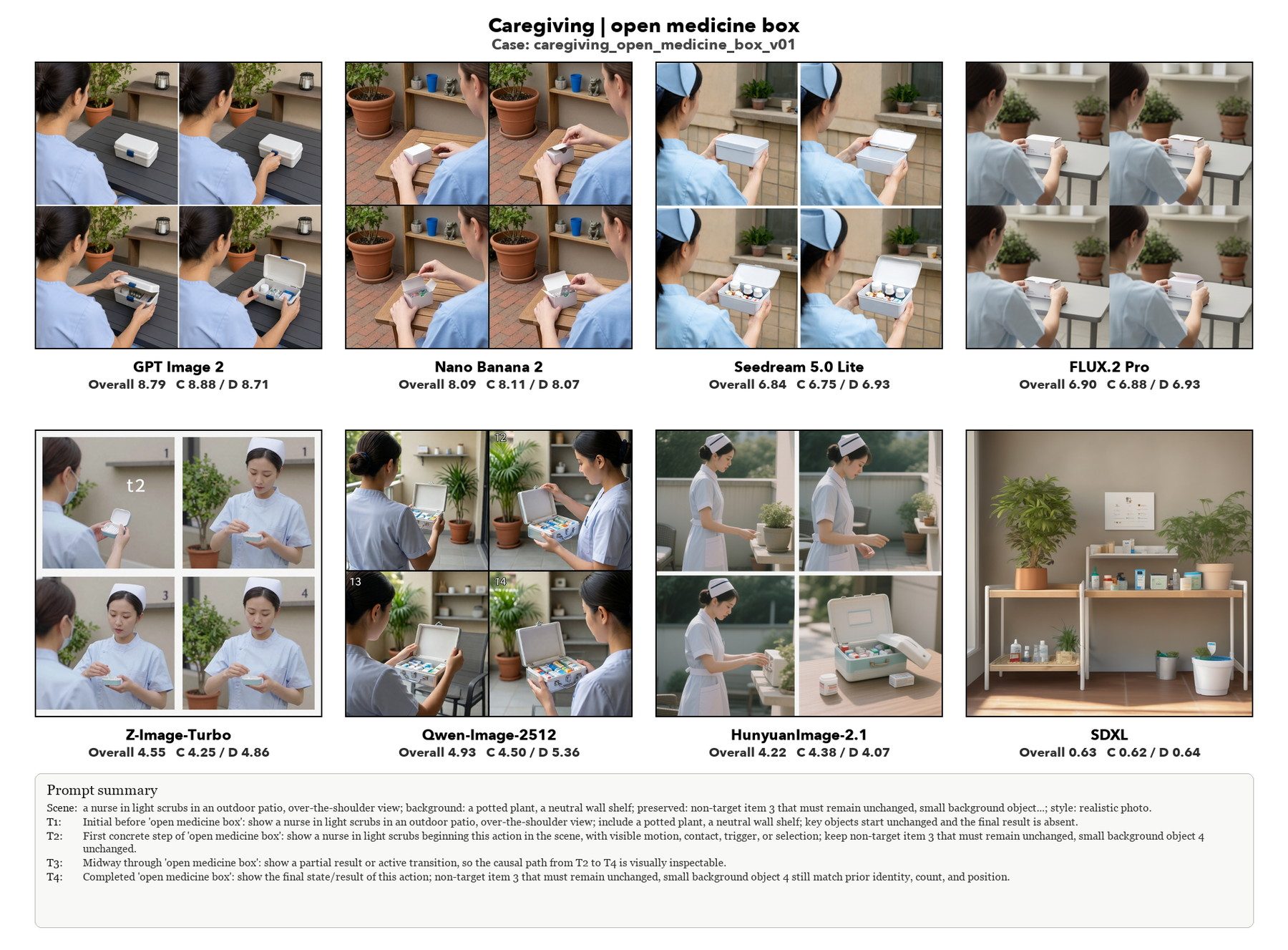}}\par
    \captionof{figure}{Example summary grids from the Caregiving category. Each grid compares the same prompt across eight image-generation models and reports the overall mean score below each generated motion sheet.}
    \label{fig:appendix-examples-caregiving}
\end{center}
\clearpage

\begin{center}
    \makebox[\textwidth][c]{\includegraphics[width=1.08\textwidth,height=0.435\textheight,keepaspectratio]{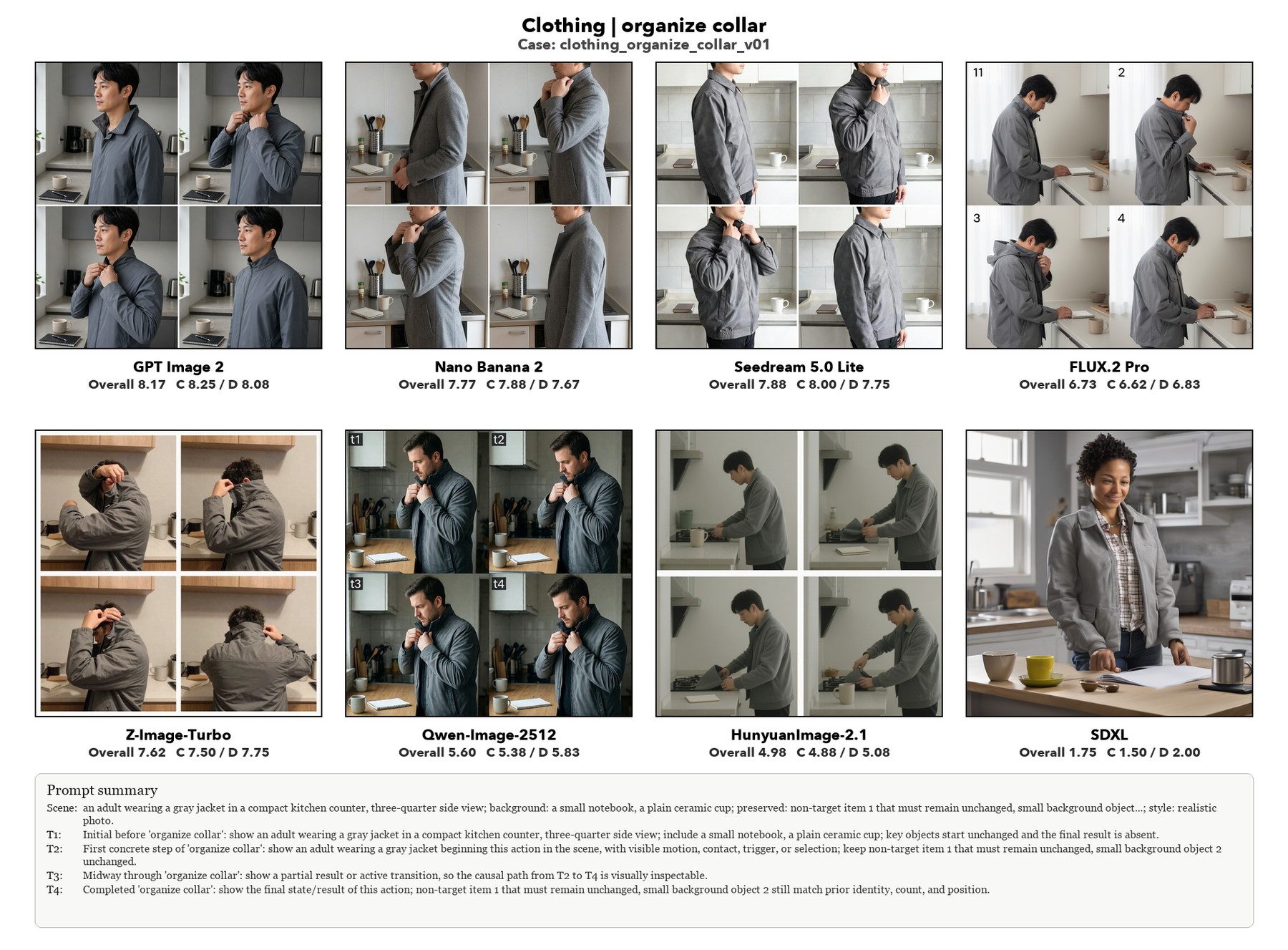}}\par
    \vspace{0.4em}
    \makebox[\textwidth][c]{\includegraphics[width=1.08\textwidth,height=0.435\textheight,keepaspectratio]{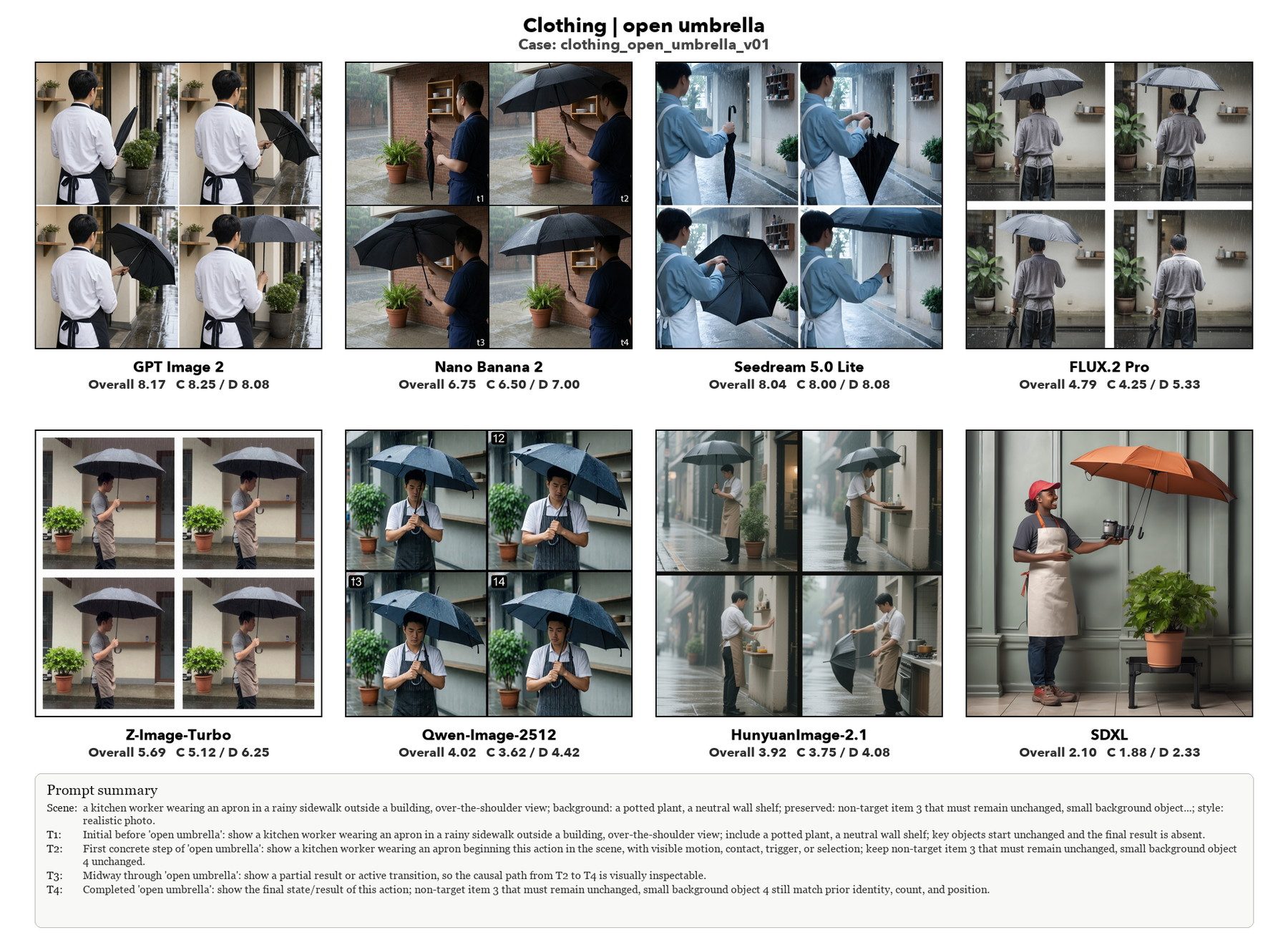}}\par
    \captionof{figure}{Example summary grids from the Clothing category. Each grid compares the same prompt across eight image-generation models and reports the overall mean score below each generated motion sheet.}
    \label{fig:appendix-examples-clothing}
\end{center}
\clearpage

\begin{center}
    \makebox[\textwidth][c]{\includegraphics[width=1.08\textwidth,height=0.435\textheight,keepaspectratio]{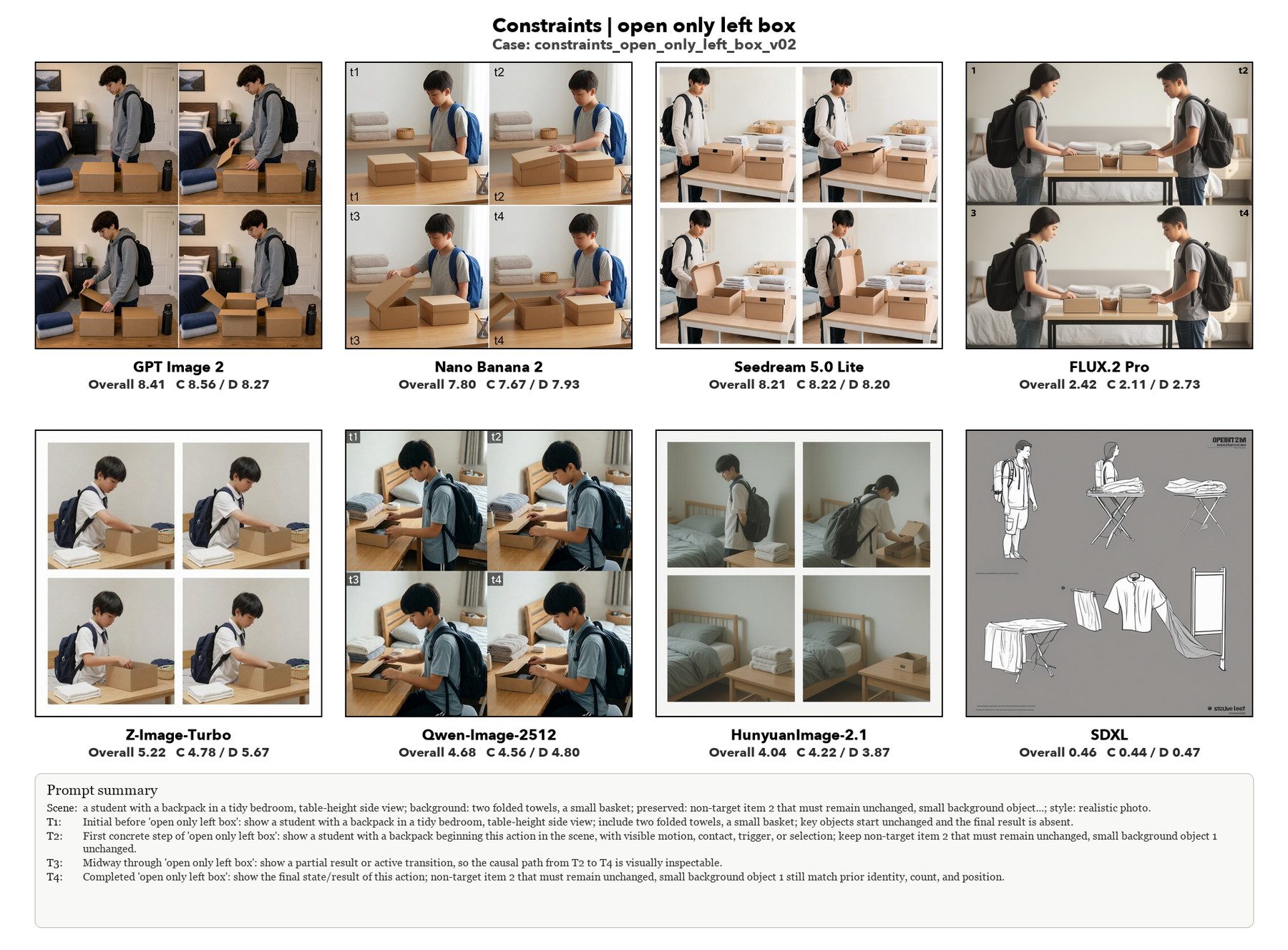}}\par
    \vspace{0.4em}
    \makebox[\textwidth][c]{\includegraphics[width=1.08\textwidth,height=0.435\textheight,keepaspectratio]{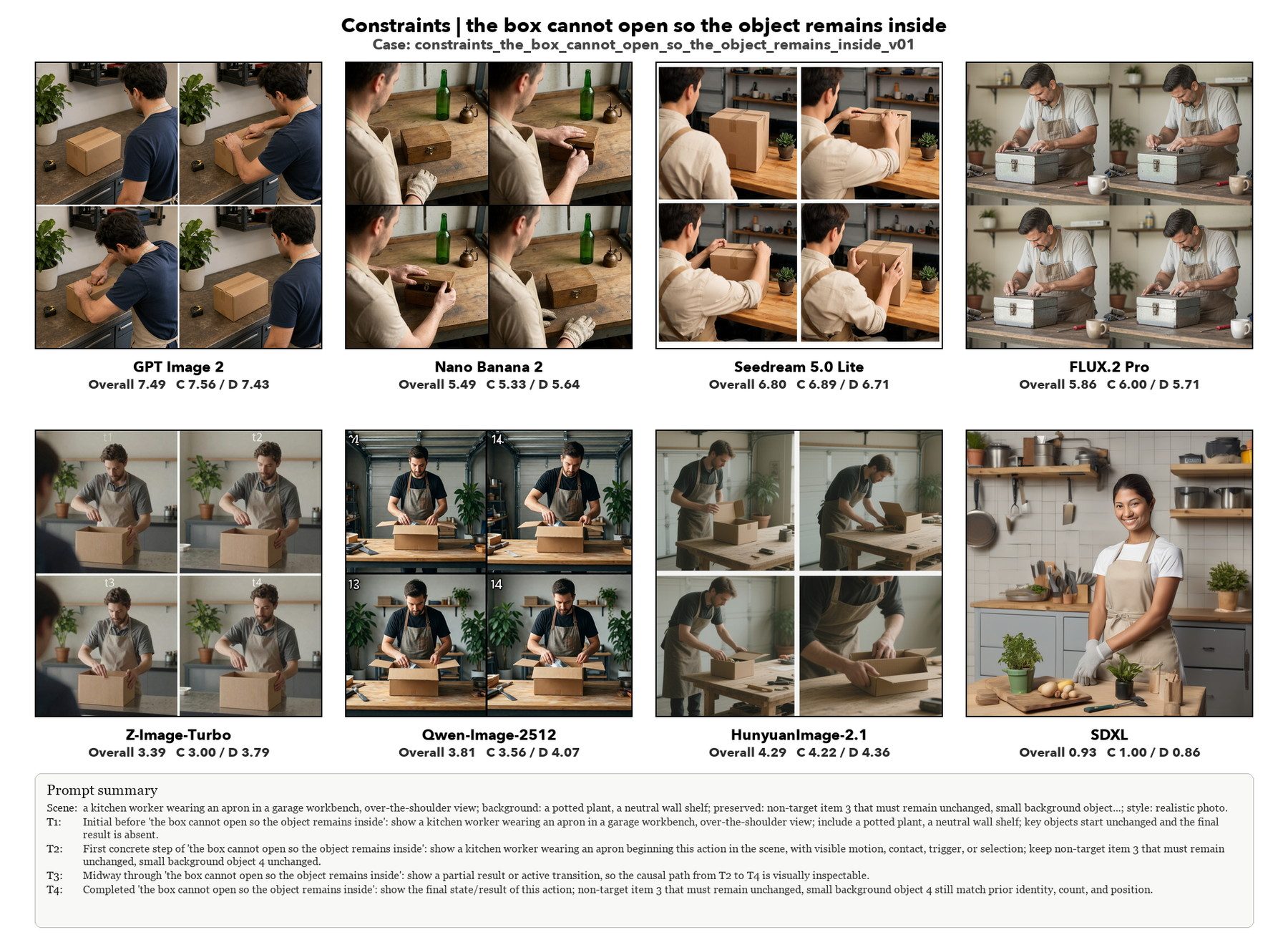}}\par
    \captionof{figure}{Example summary grids from the Constraints category. Each grid compares the same prompt across eight image-generation models and reports the overall mean score below each generated motion sheet.}
    \label{fig:appendix-examples-constraints}
\end{center}
\clearpage

\begin{center}
    \makebox[\textwidth][c]{\includegraphics[width=1.08\textwidth,height=0.435\textheight,keepaspectratio]{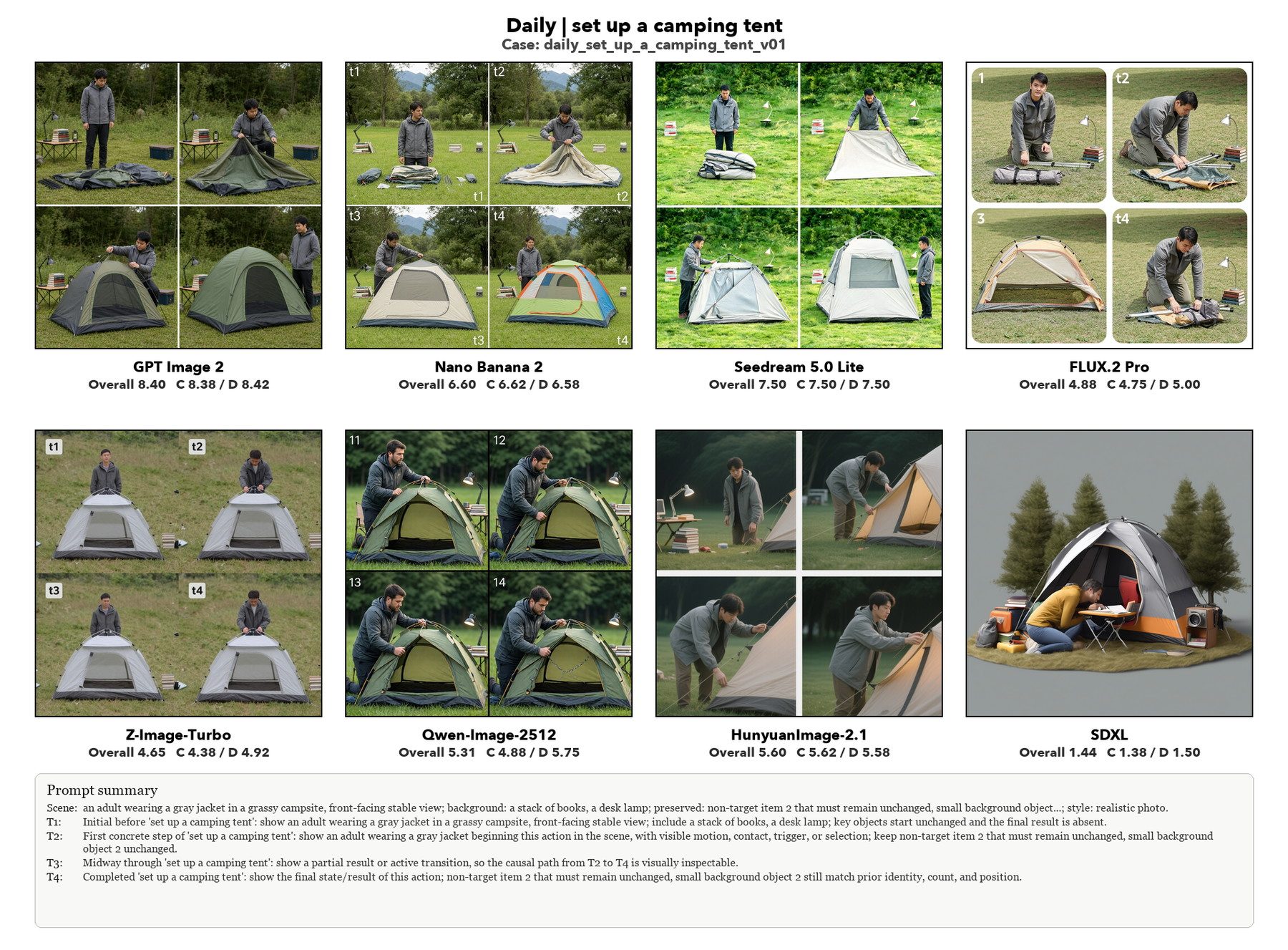}}\par
    \vspace{0.4em}
    \makebox[\textwidth][c]{\includegraphics[width=1.08\textwidth,height=0.435\textheight,keepaspectratio]{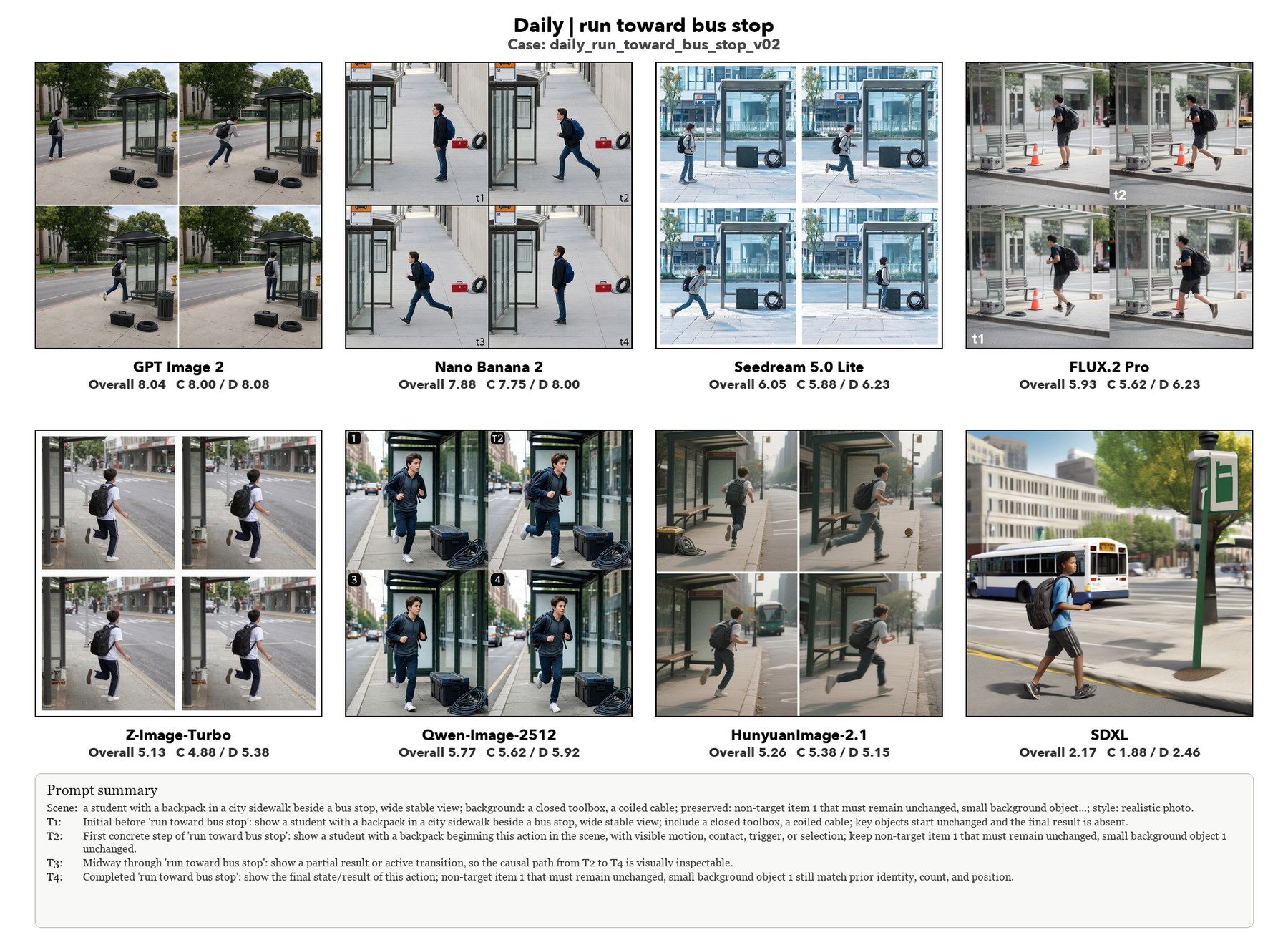}}\par
    \captionof{figure}{Example summary grids from the Daily category. Each grid compares the same prompt across eight image-generation models and reports the overall mean score below each generated motion sheet.}
    \label{fig:appendix-examples-daily}
\end{center}
\clearpage

\begin{center}
    \makebox[\textwidth][c]{\includegraphics[width=1.08\textwidth,height=0.435\textheight,keepaspectratio]{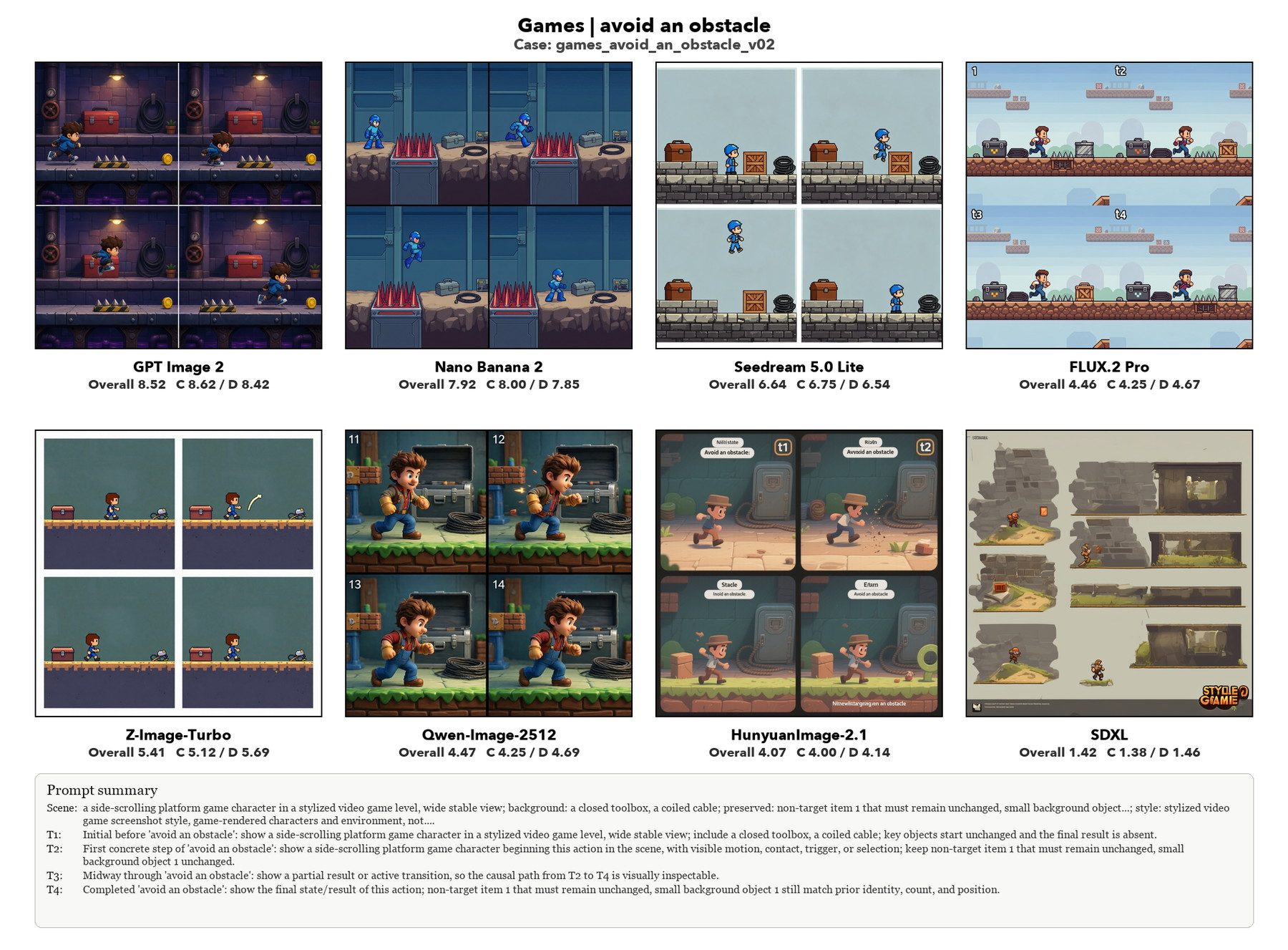}}\par
    \vspace{0.4em}
    \makebox[\textwidth][c]{\includegraphics[width=1.08\textwidth,height=0.435\textheight,keepaspectratio]{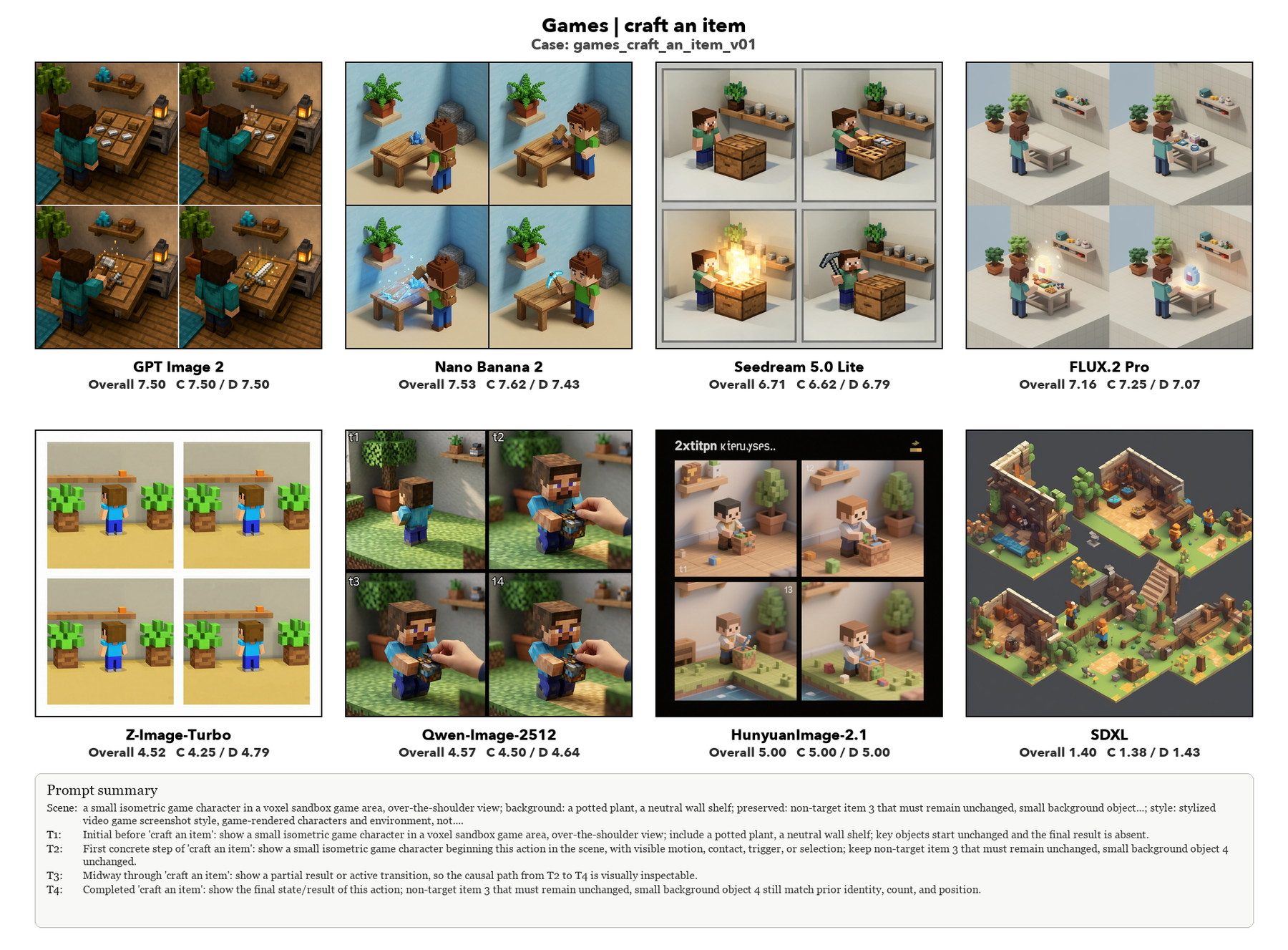}}\par
    \captionof{figure}{Example summary grids from the Games category. Each grid compares the same prompt across eight image-generation models and reports the overall mean score below each generated motion sheet.}
    \label{fig:appendix-examples-games}
\end{center}
\clearpage

\begin{center}
    \makebox[\textwidth][c]{\includegraphics[width=1.08\textwidth,height=0.435\textheight,keepaspectratio]{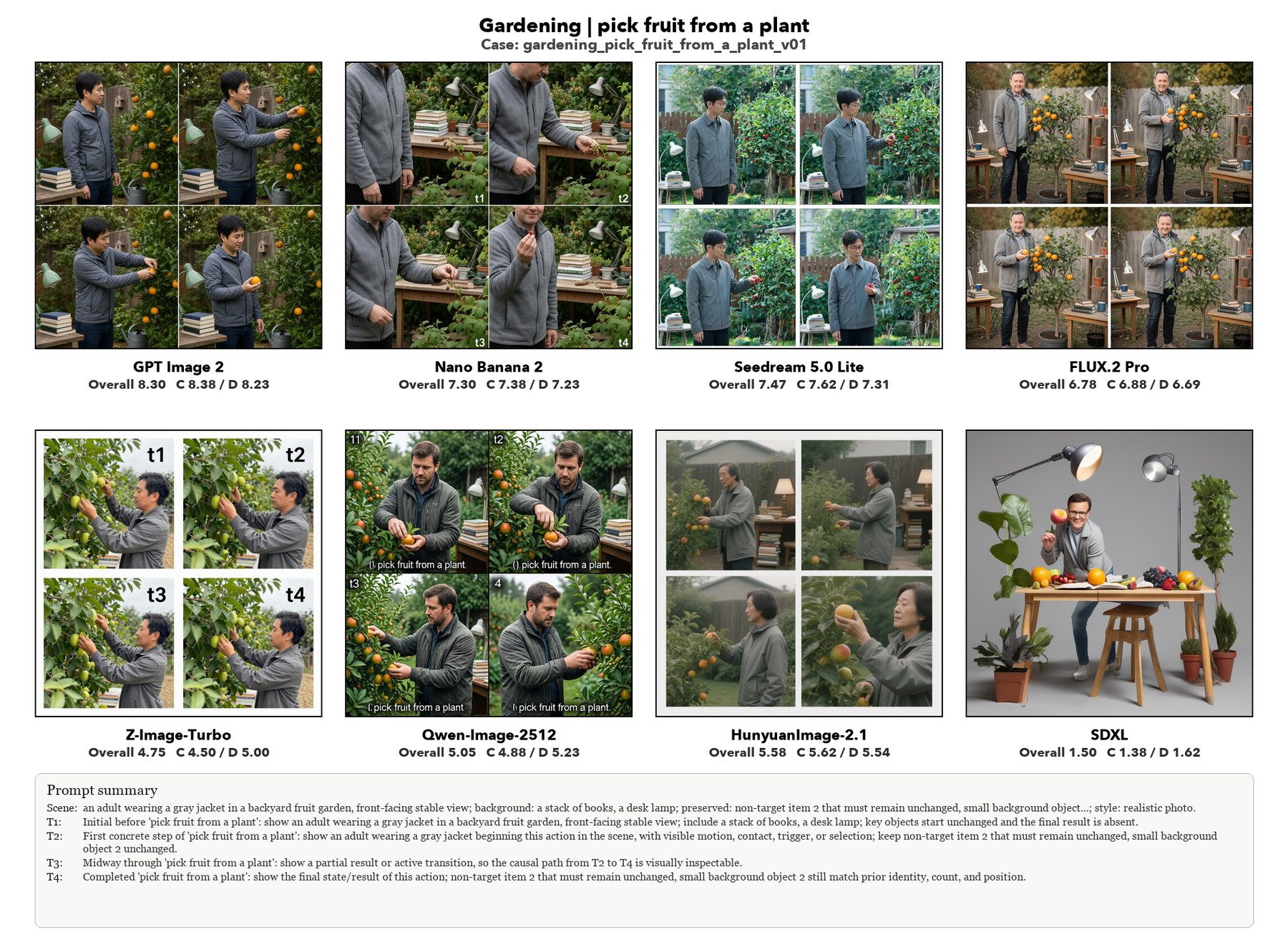}}\par
    \vspace{0.4em}
    \makebox[\textwidth][c]{\includegraphics[width=1.08\textwidth,height=0.435\textheight,keepaspectratio]{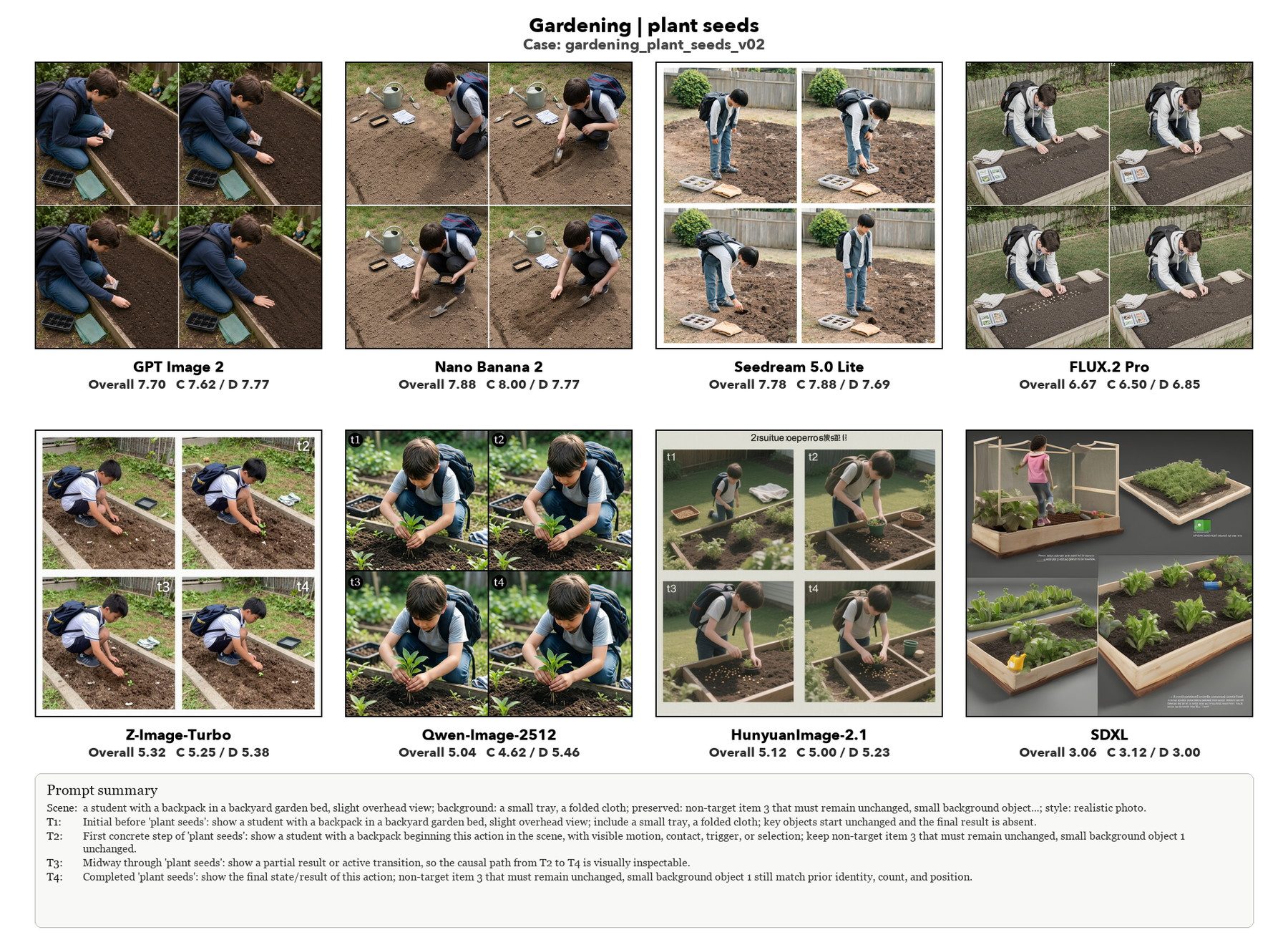}}\par
    \captionof{figure}{Example summary grids from the Gardening category. Each grid compares the same prompt across eight image-generation models and reports the overall mean score below each generated motion sheet.}
    \label{fig:appendix-examples-gardening}
\end{center}
\clearpage

\begin{center}
    \makebox[\textwidth][c]{\includegraphics[width=1.08\textwidth,height=0.435\textheight,keepaspectratio]{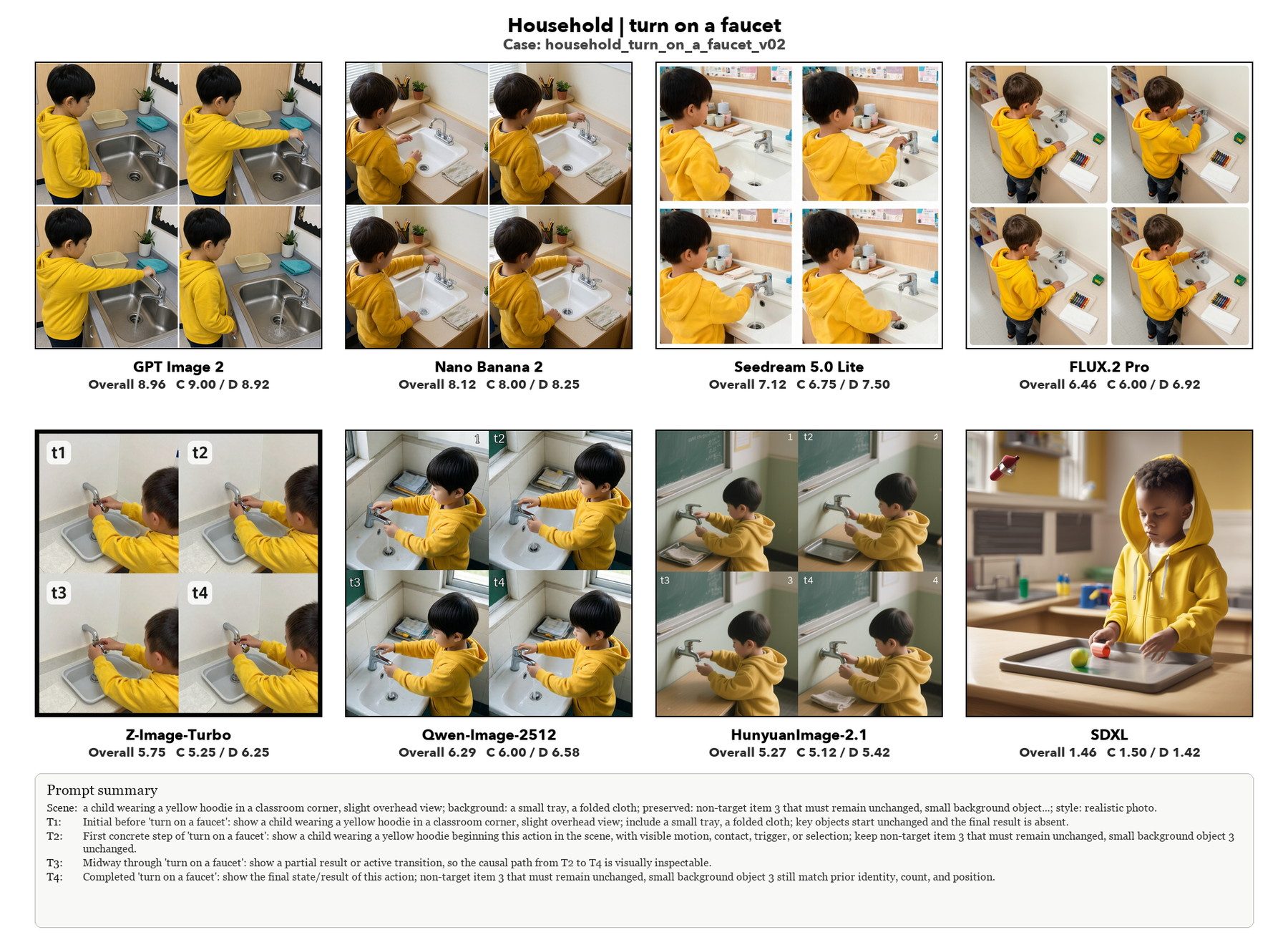}}\par
    \vspace{0.4em}
    \makebox[\textwidth][c]{\includegraphics[width=1.08\textwidth,height=0.435\textheight,keepaspectratio]{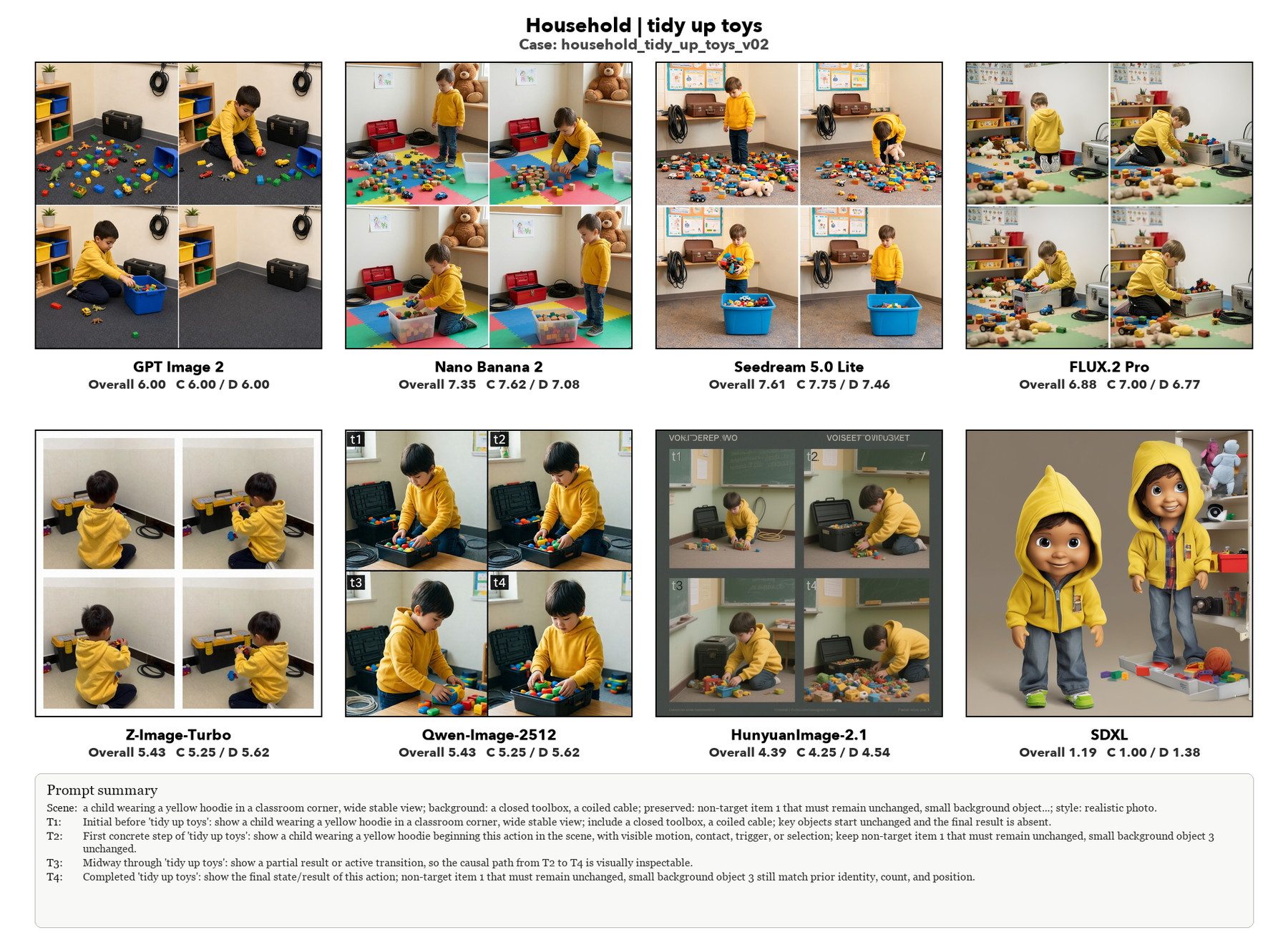}}\par
    \captionof{figure}{Example summary grids from the Household category. Each grid compares the same prompt across eight image-generation models and reports the overall mean score below each generated motion sheet.}
    \label{fig:appendix-examples-household}
\end{center}
\clearpage

\begin{center}
    \makebox[\textwidth][c]{\includegraphics[width=1.08\textwidth,height=0.435\textheight,keepaspectratio]{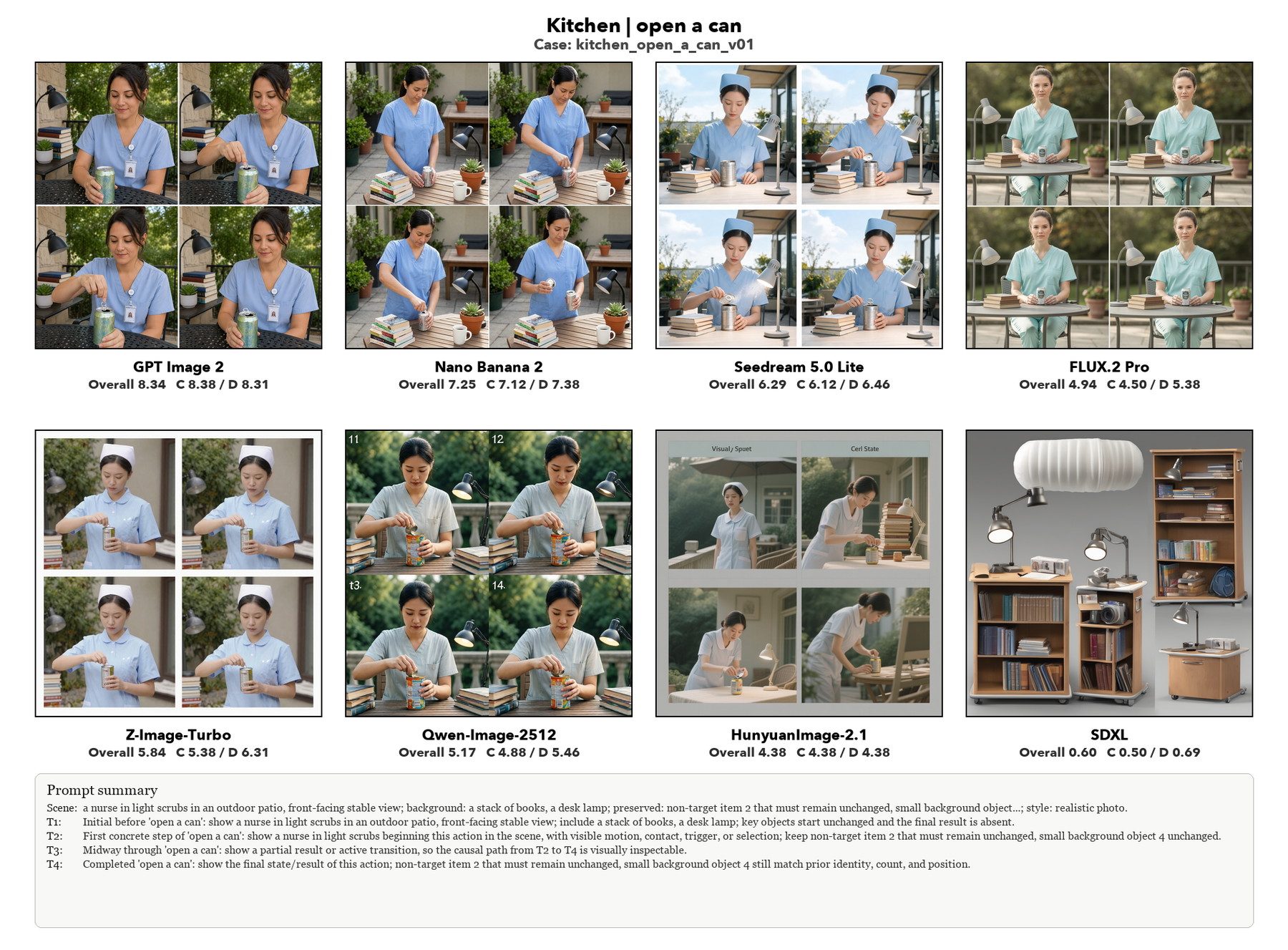}}\par
    \vspace{0.4em}
    \makebox[\textwidth][c]{\includegraphics[width=1.08\textwidth,height=0.435\textheight,keepaspectratio]{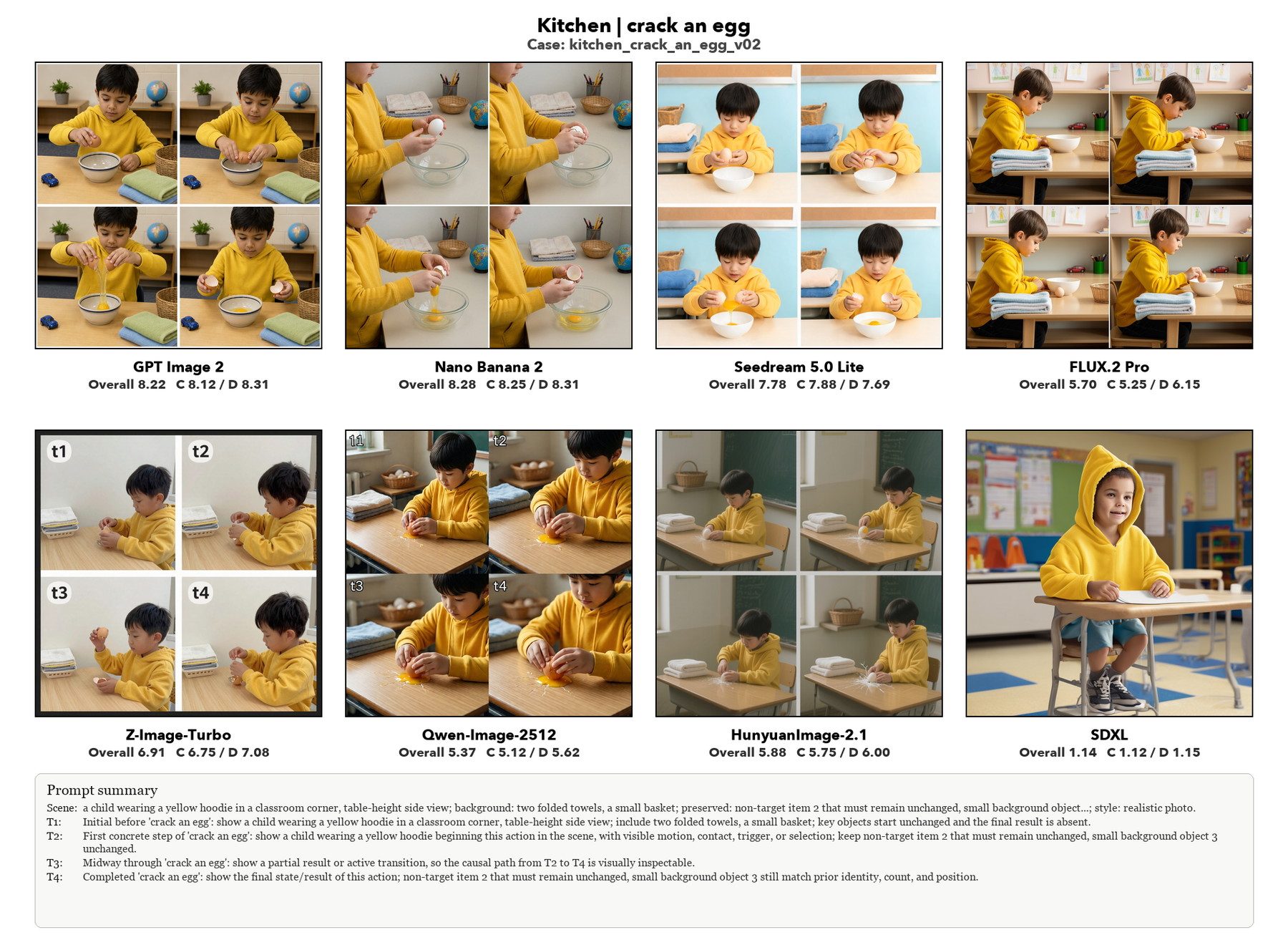}}\par
    \captionof{figure}{Example summary grids from the Kitchen category. Each grid compares the same prompt across eight image-generation models and reports the overall mean score below each generated motion sheet.}
    \label{fig:appendix-examples-kitchen}
\end{center}
\clearpage

\begin{center}
    \makebox[\textwidth][c]{\includegraphics[width=1.08\textwidth,height=0.435\textheight,keepaspectratio]{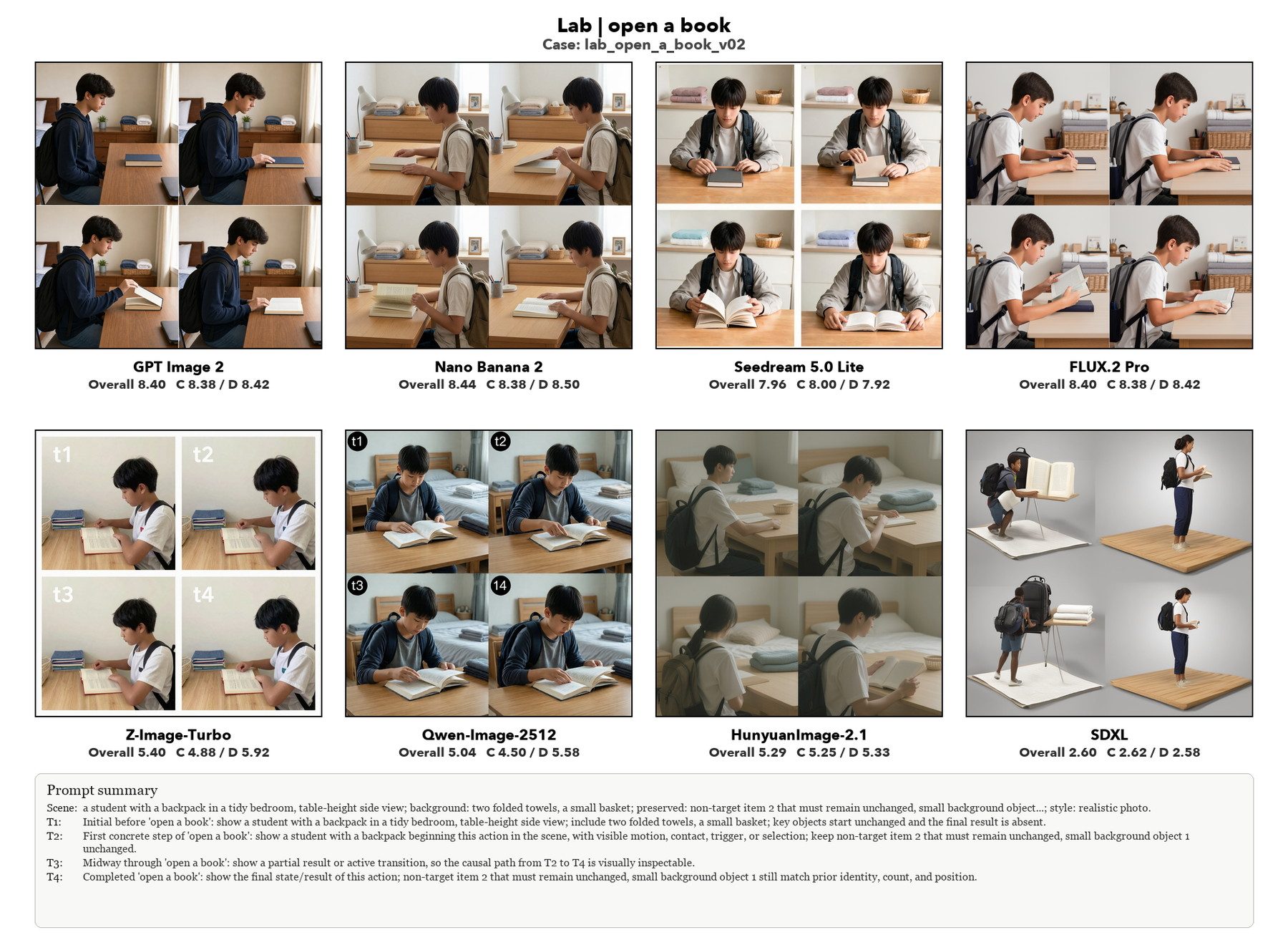}}\par
    \vspace{0.4em}
    \makebox[\textwidth][c]{\includegraphics[width=1.08\textwidth,height=0.435\textheight,keepaspectratio]{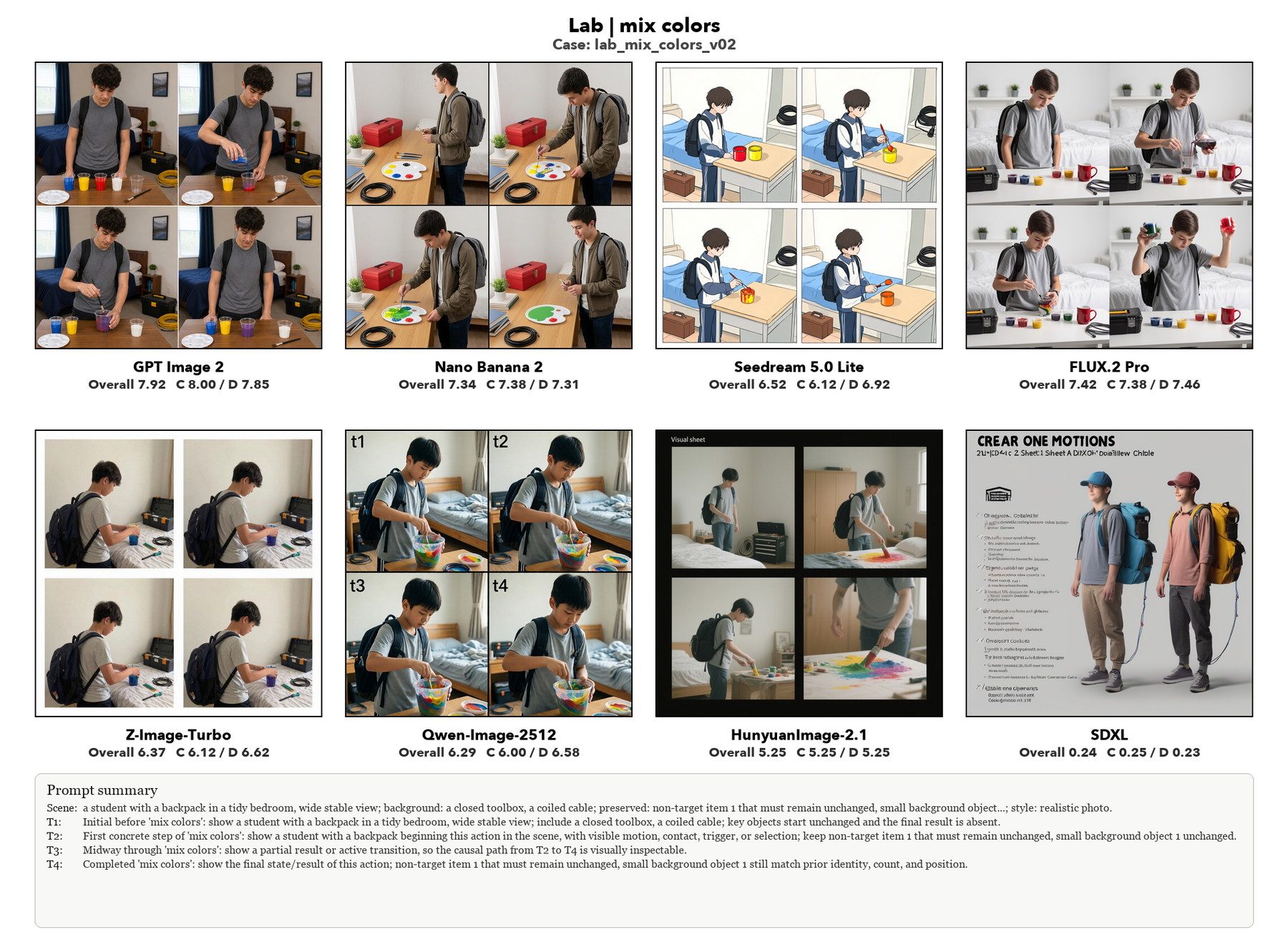}}\par
    \captionof{figure}{Example summary grids from the Lab category. Each grid compares the same prompt across eight image-generation models and reports the overall mean score below each generated motion sheet.}
    \label{fig:appendix-examples-lab}
\end{center}
\clearpage

\begin{center}
    \makebox[\textwidth][c]{\includegraphics[width=1.08\textwidth,height=0.435\textheight,keepaspectratio]{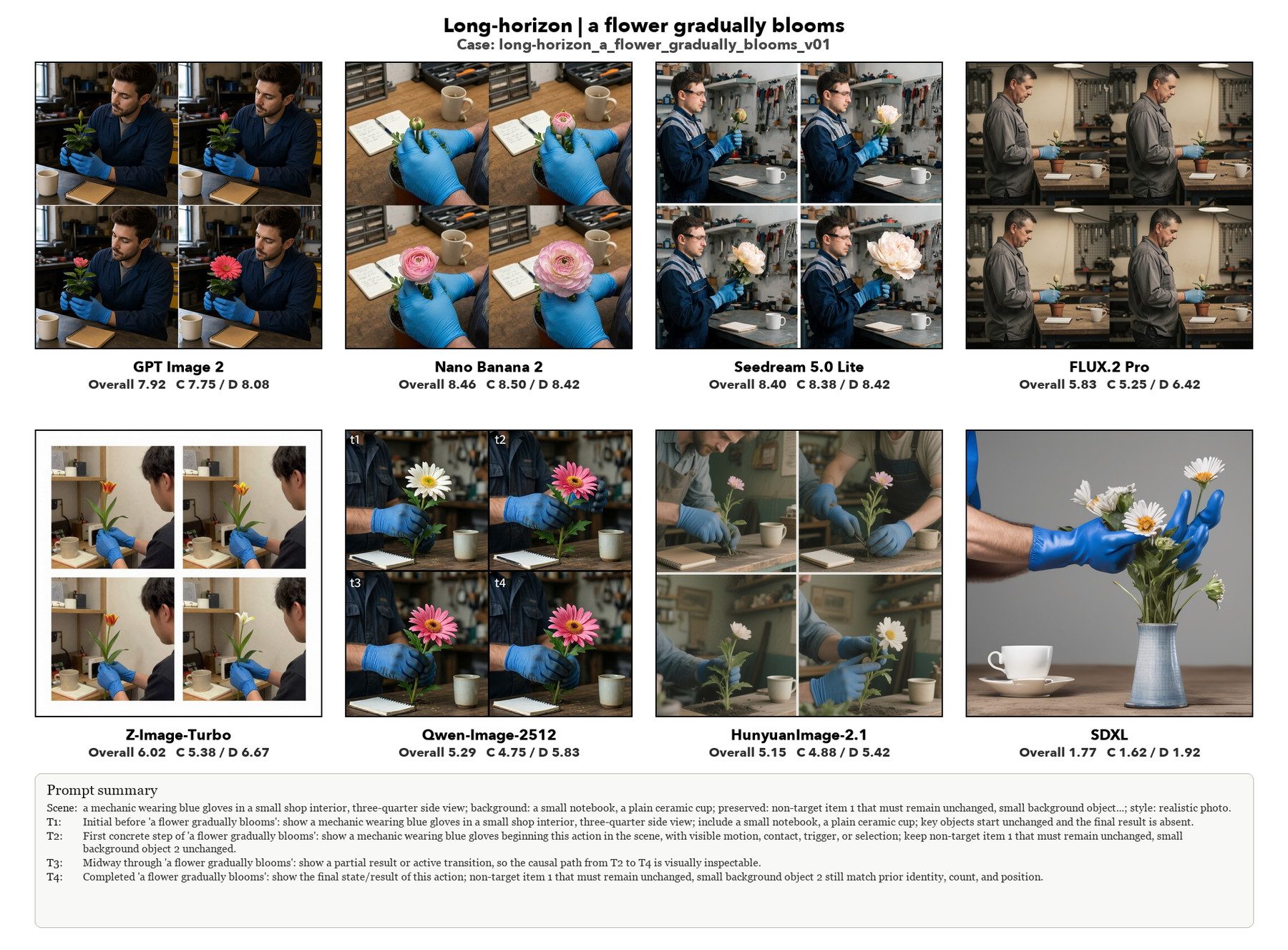}}\par
    \vspace{0.4em}
    \makebox[\textwidth][c]{\includegraphics[width=1.08\textwidth,height=0.435\textheight,keepaspectratio]{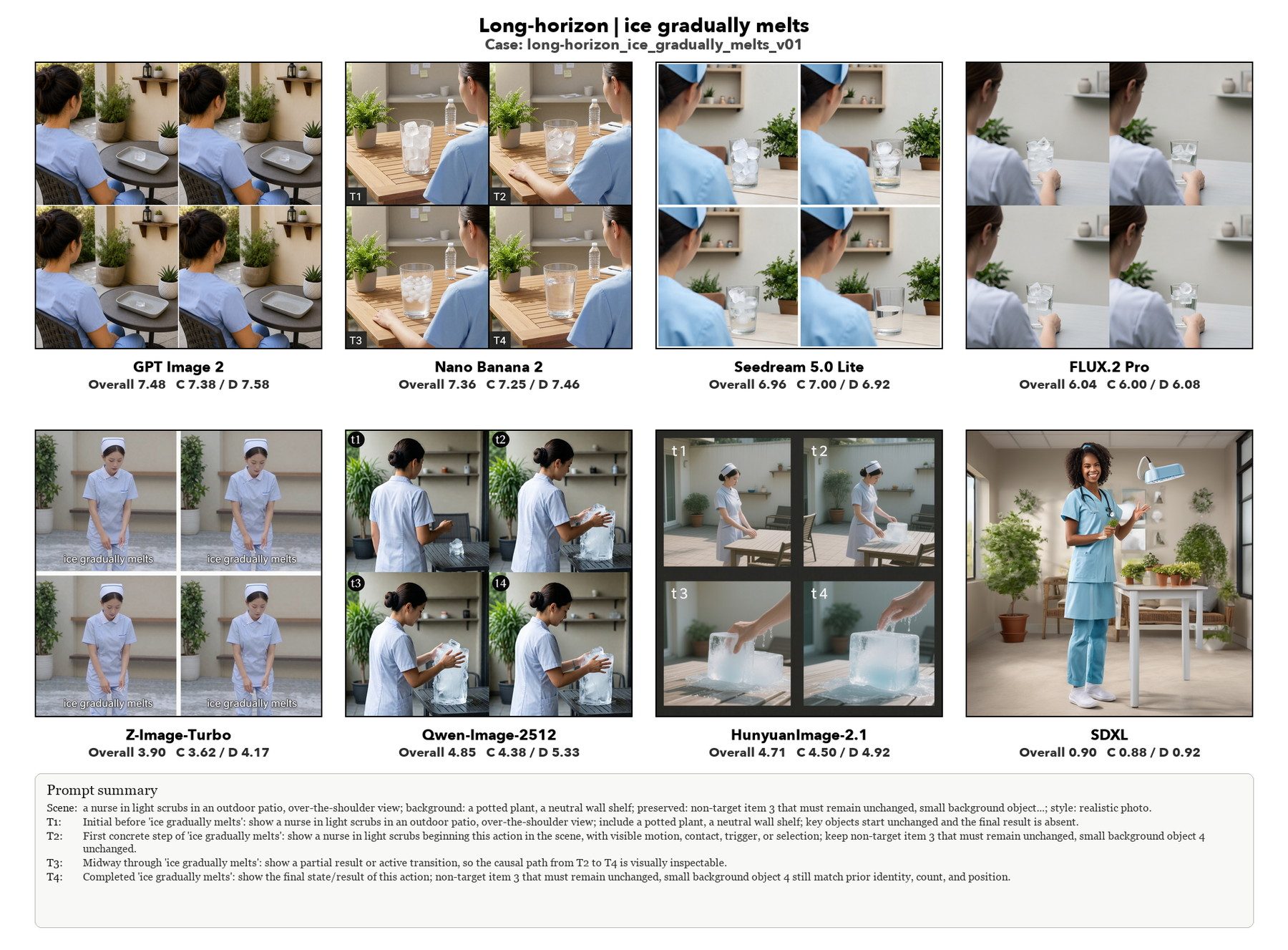}}\par
    \captionof{figure}{Example summary grids from the Long Horizon category. Each grid compares the same prompt across eight image-generation models and reports the overall mean score below each generated motion sheet.}
    \label{fig:appendix-examples-long-horizon}
\end{center}
\clearpage

\begin{center}
    \makebox[\textwidth][c]{\includegraphics[width=1.08\textwidth,height=0.435\textheight,keepaspectratio]{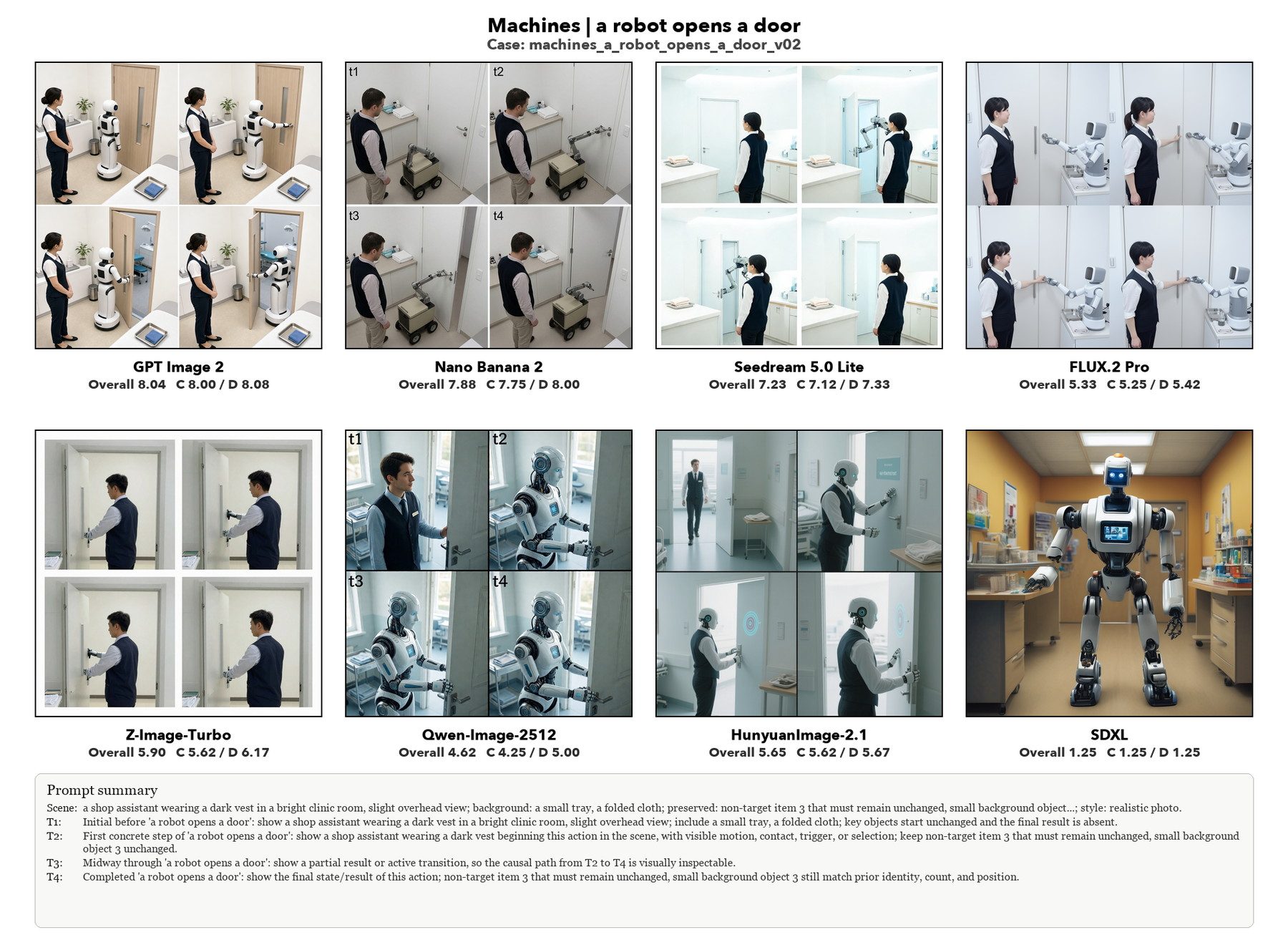}}\par
    \vspace{0.4em}
    \makebox[\textwidth][c]{\includegraphics[width=1.08\textwidth,height=0.435\textheight,keepaspectratio]{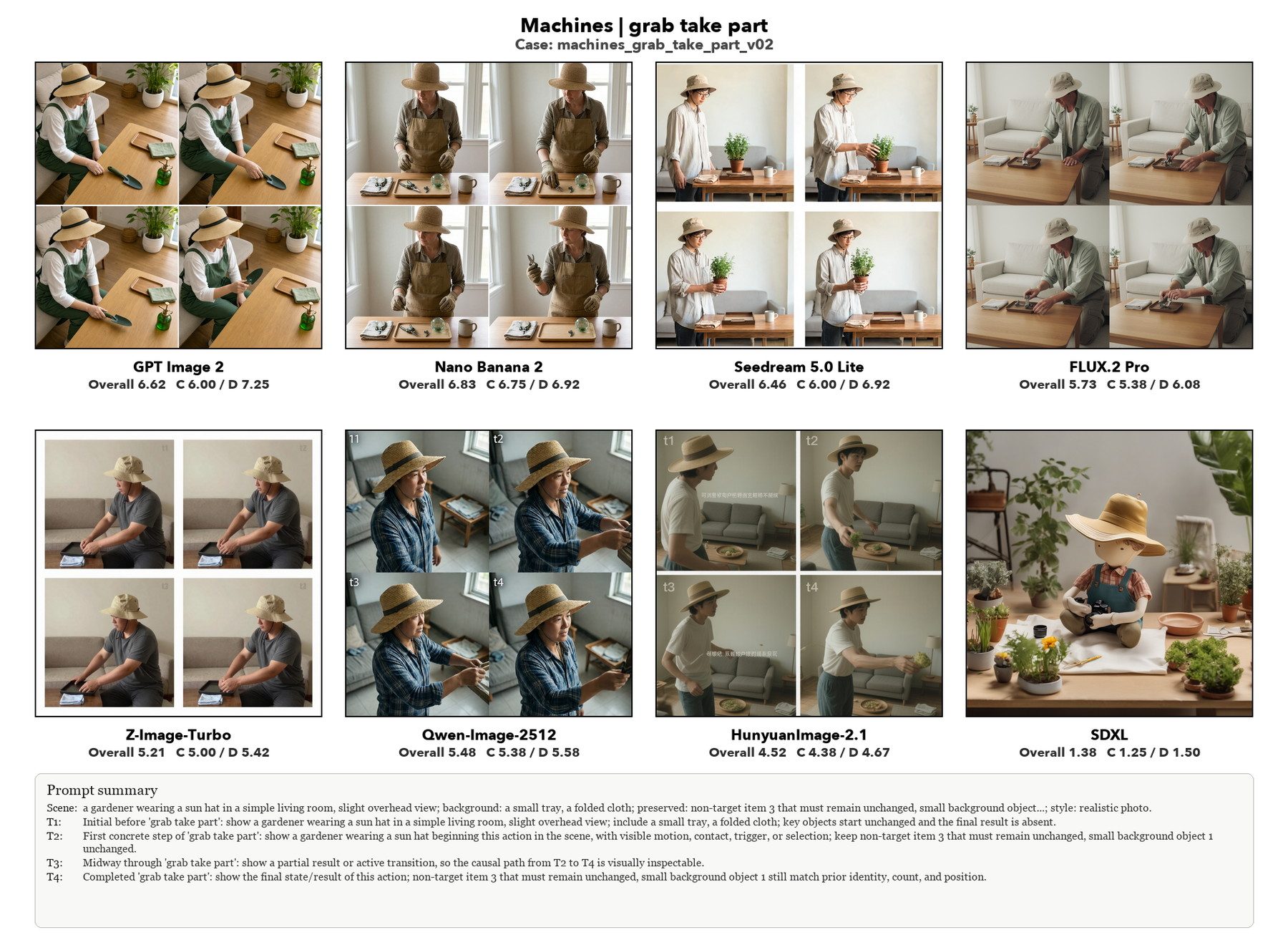}}\par
    \captionof{figure}{Example summary grids from the Machines category. Each grid compares the same prompt across eight image-generation models and reports the overall mean score below each generated motion sheet.}
    \label{fig:appendix-examples-machines}
\end{center}
\clearpage

\begin{center}
    \makebox[\textwidth][c]{\includegraphics[width=1.08\textwidth,height=0.435\textheight,keepaspectratio]{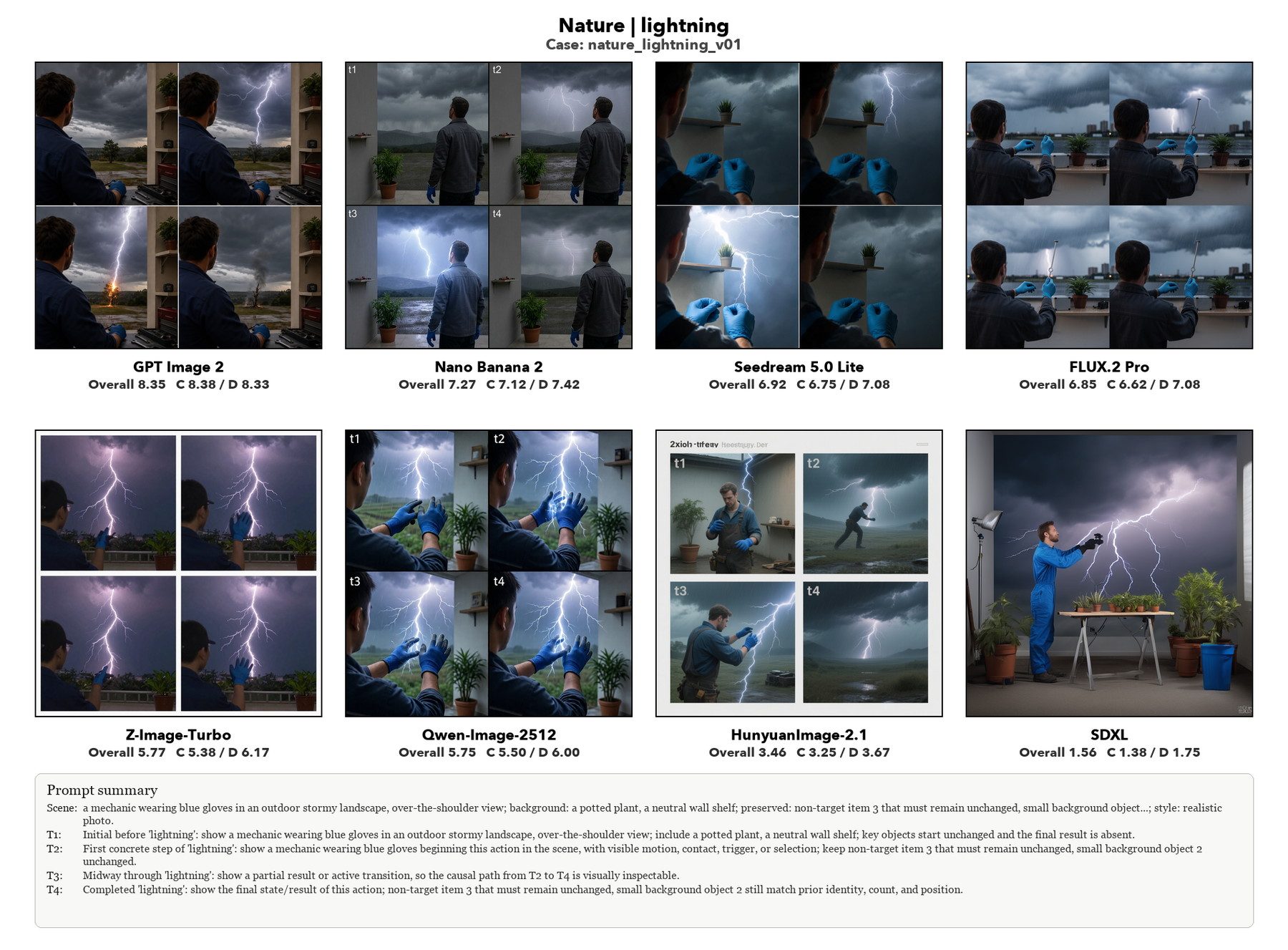}}\par
    \vspace{0.4em}
    \makebox[\textwidth][c]{\includegraphics[width=1.08\textwidth,height=0.435\textheight,keepaspectratio]{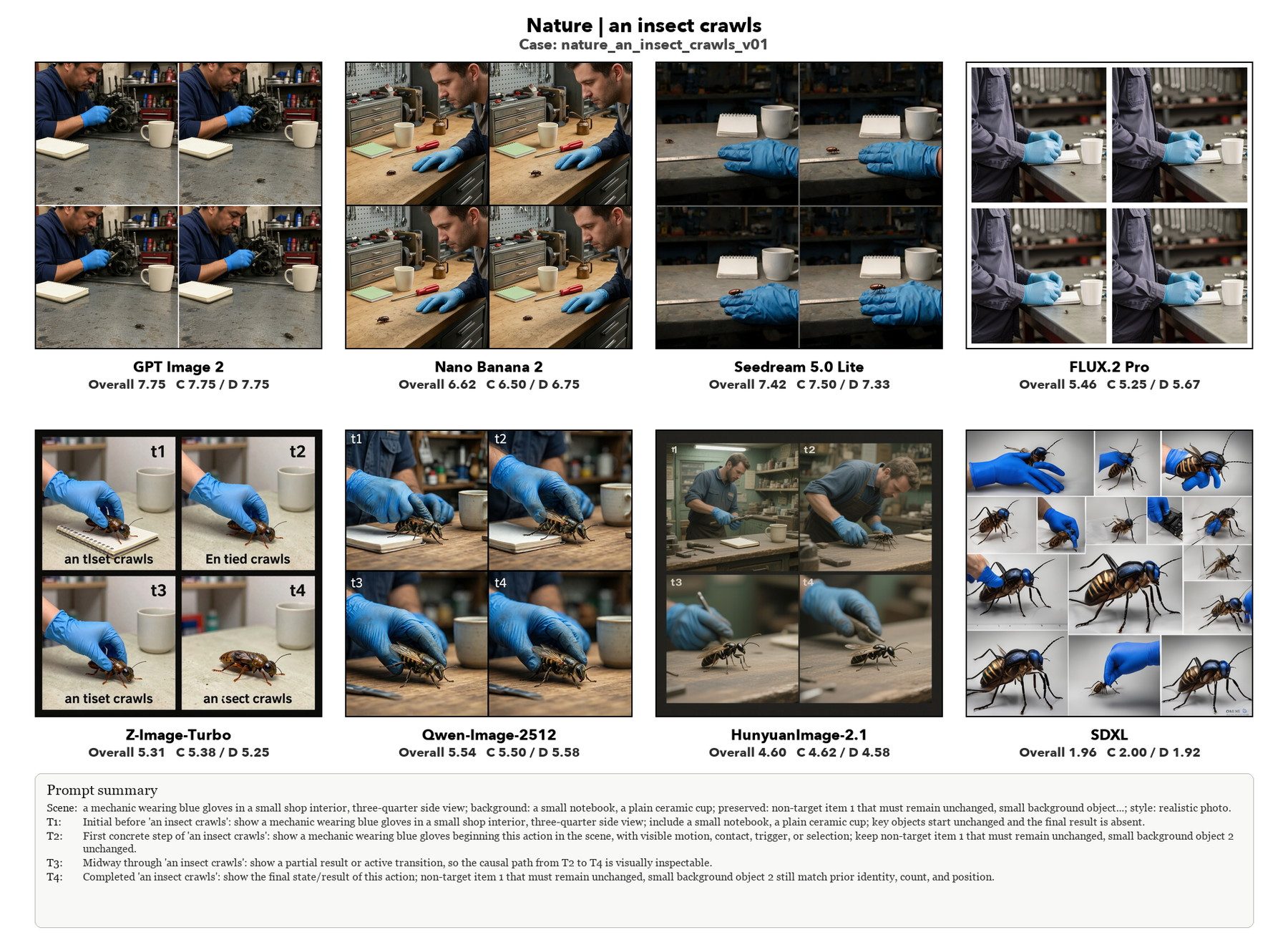}}\par
    \captionof{figure}{Example summary grids from the Nature category. Each grid compares the same prompt across eight image-generation models and reports the overall mean score below each generated motion sheet.}
    \label{fig:appendix-examples-nature}
\end{center}
\clearpage

\begin{center}
    \makebox[\textwidth][c]{\includegraphics[width=1.08\textwidth,height=0.435\textheight,keepaspectratio]{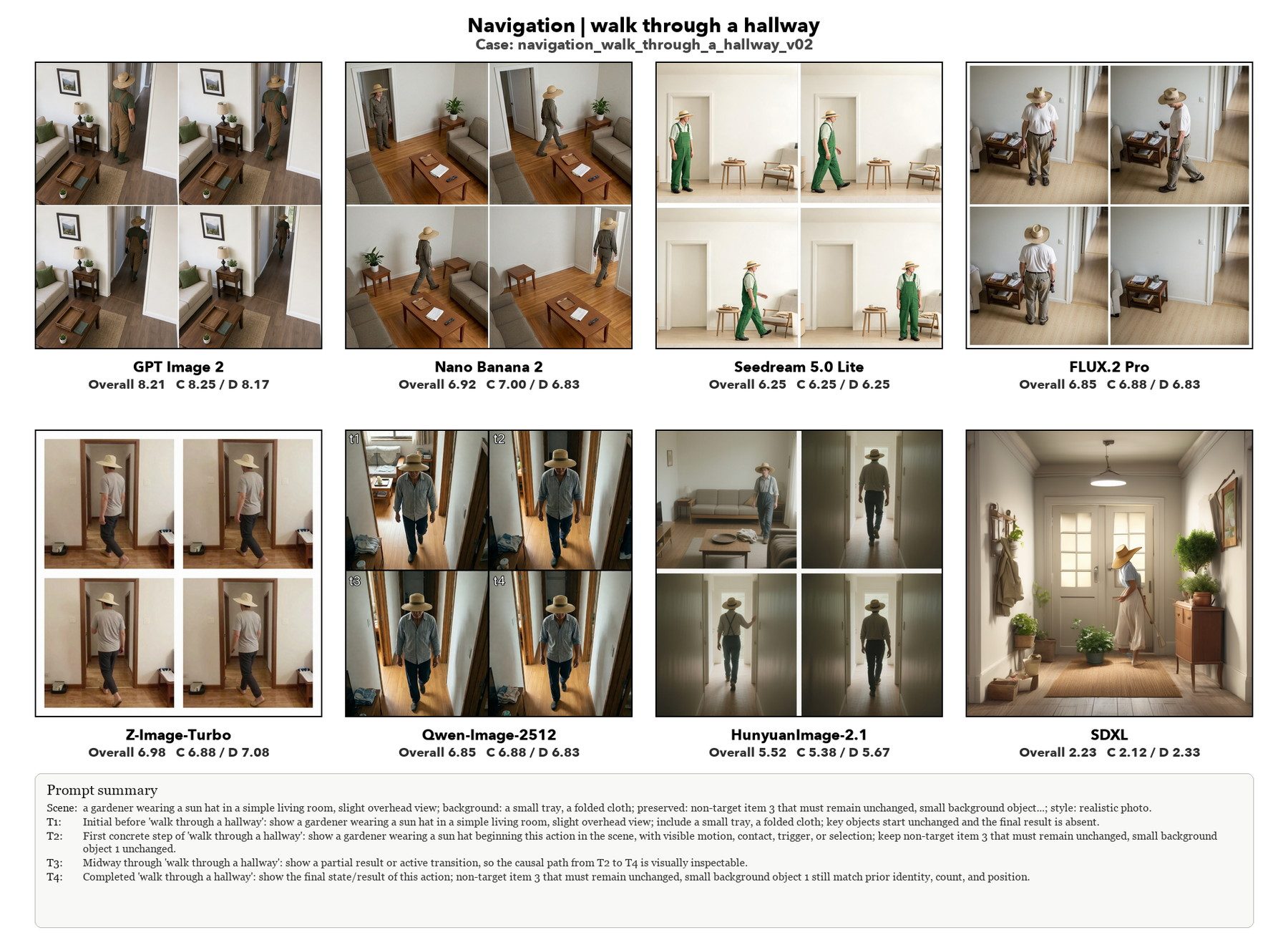}}\par
    \vspace{0.4em}
    \makebox[\textwidth][c]{\includegraphics[width=1.08\textwidth,height=0.435\textheight,keepaspectratio]{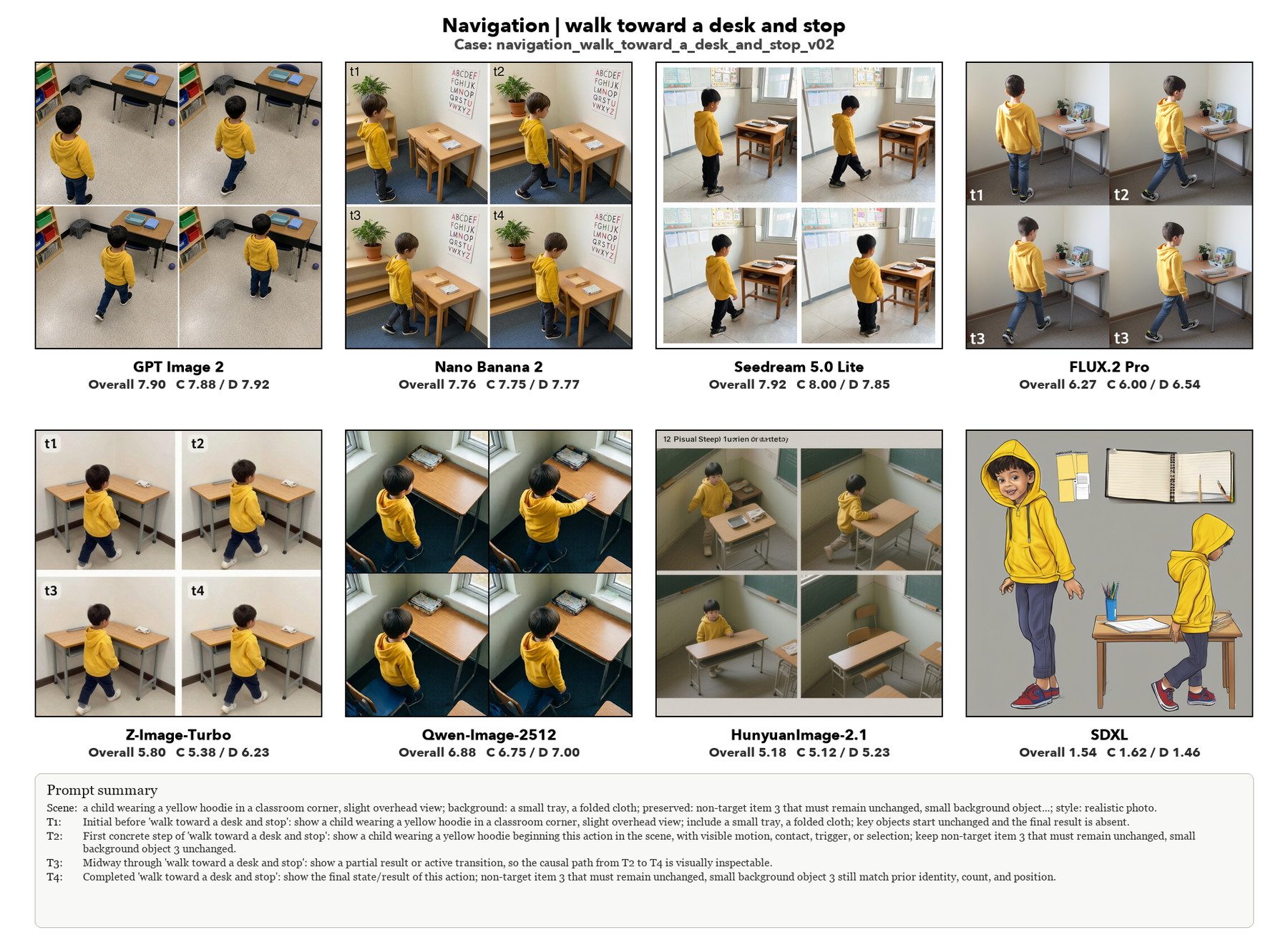}}\par
    \captionof{figure}{Example summary grids from the Navigation category. Each grid compares the same prompt across eight image-generation models and reports the overall mean score below each generated motion sheet.}
    \label{fig:appendix-examples-navigation}
\end{center}
\clearpage

\begin{center}
    \makebox[\textwidth][c]{\includegraphics[width=1.08\textwidth,height=0.435\textheight,keepaspectratio]{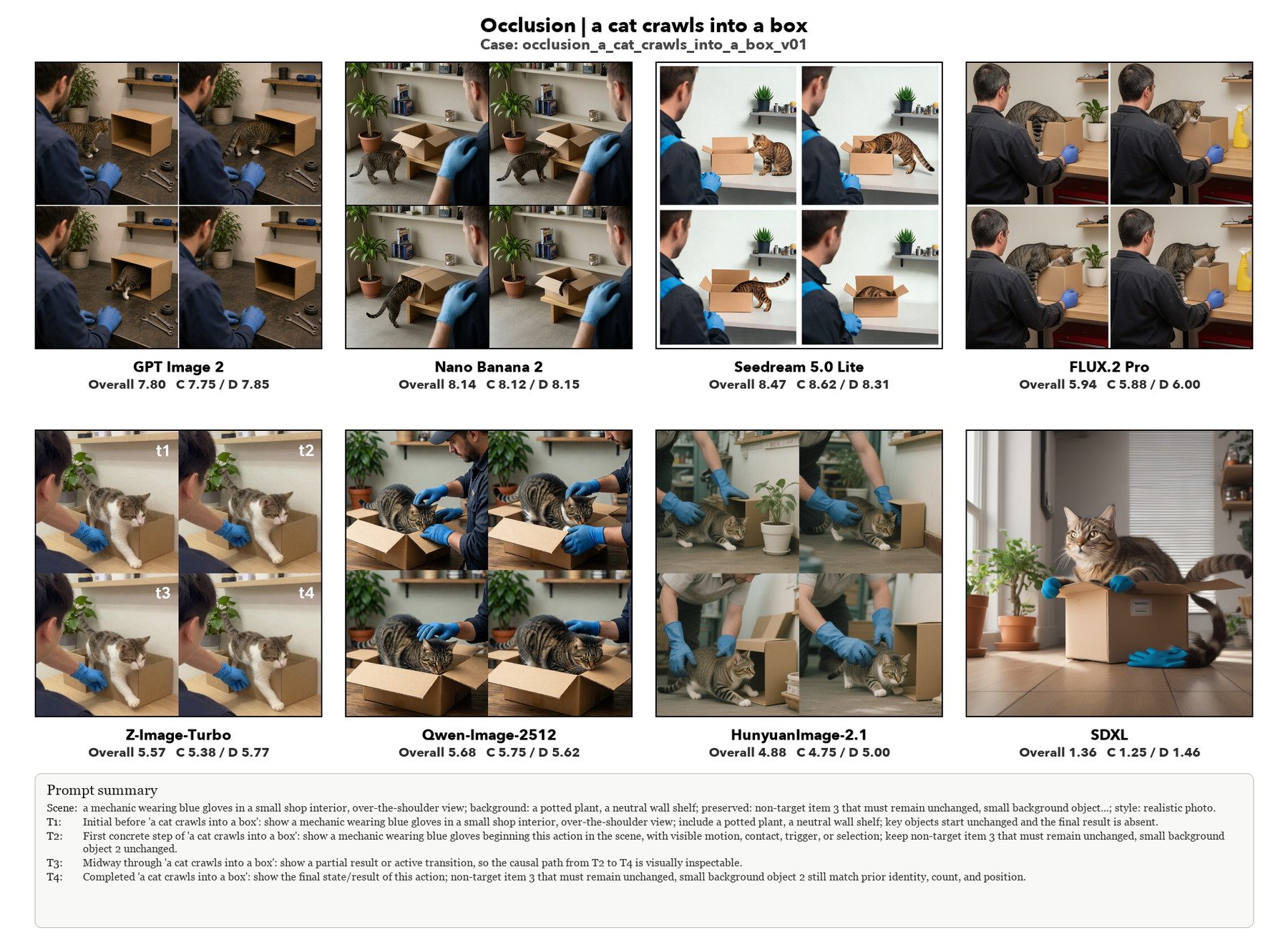}}\par
    \vspace{0.4em}
    \makebox[\textwidth][c]{\includegraphics[width=1.08\textwidth,height=0.435\textheight,keepaspectratio]{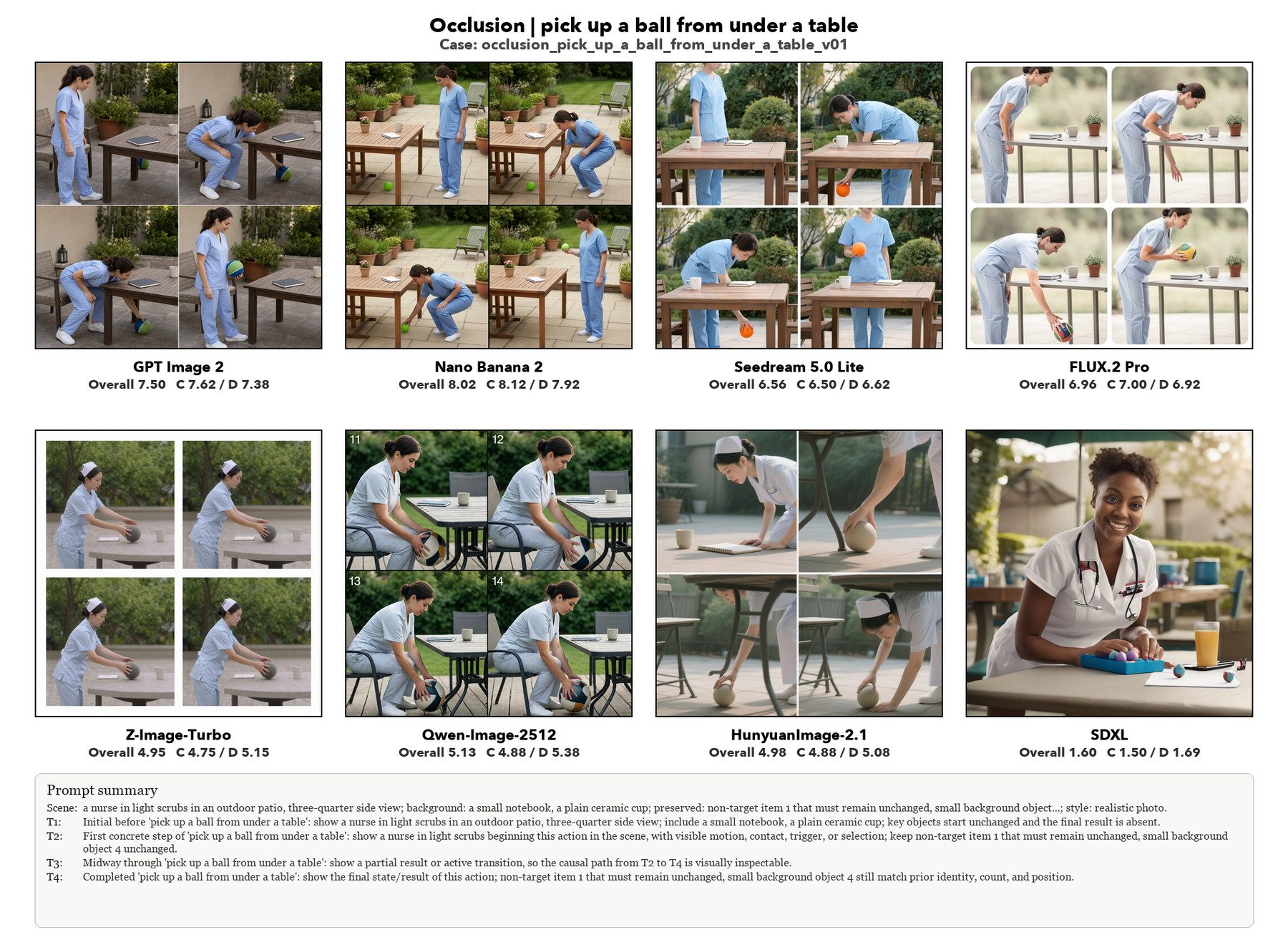}}\par
    \captionof{figure}{Example summary grids from the Occlusion category. Each grid compares the same prompt across eight image-generation models and reports the overall mean score below each generated motion sheet.}
    \label{fig:appendix-examples-occlusion}
\end{center}
\clearpage

\begin{center}
    \makebox[\textwidth][c]{\includegraphics[width=1.08\textwidth,height=0.435\textheight,keepaspectratio]{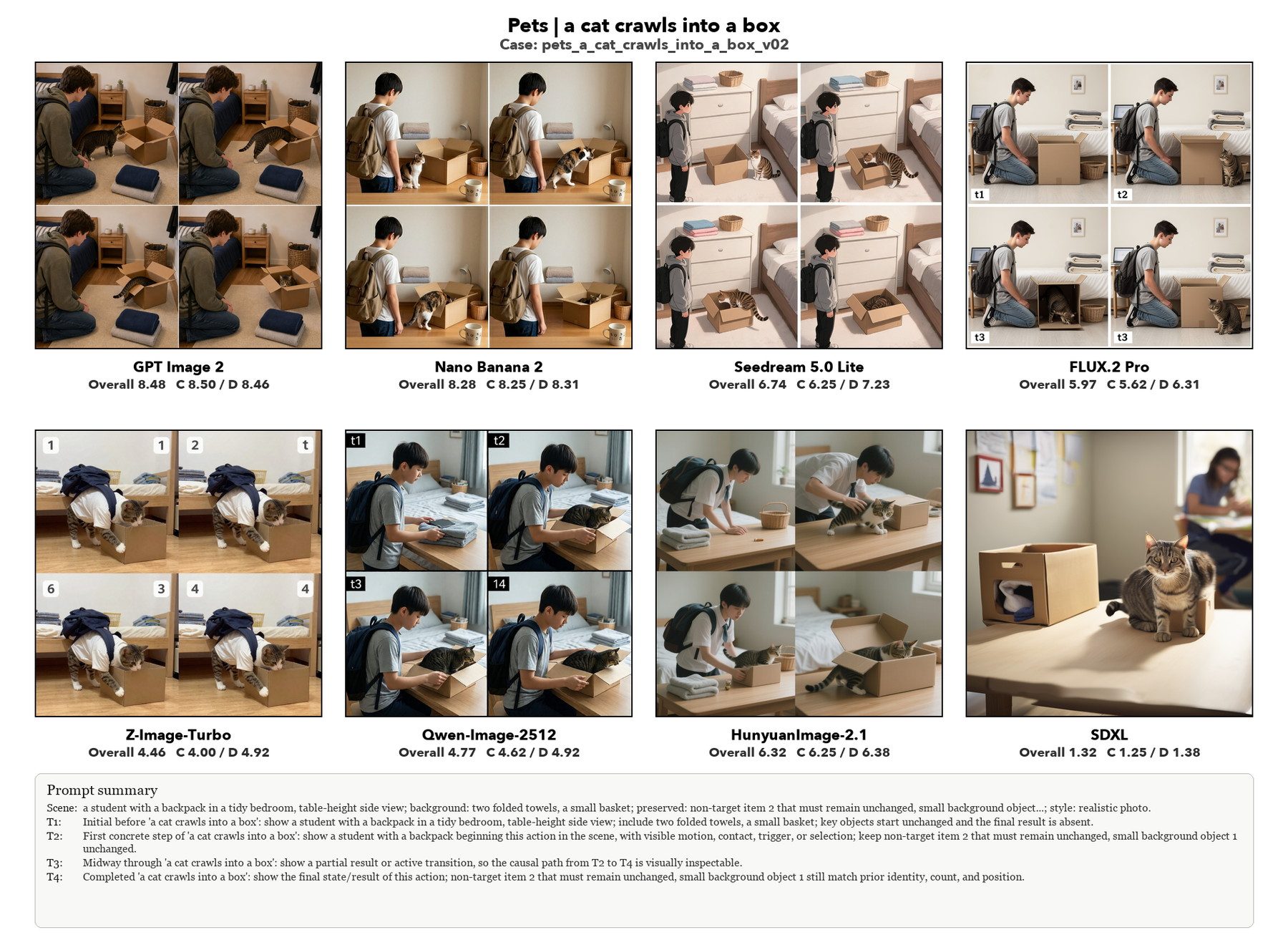}}\par
    \vspace{0.4em}
    \makebox[\textwidth][c]{\includegraphics[width=1.08\textwidth,height=0.435\textheight,keepaspectratio]{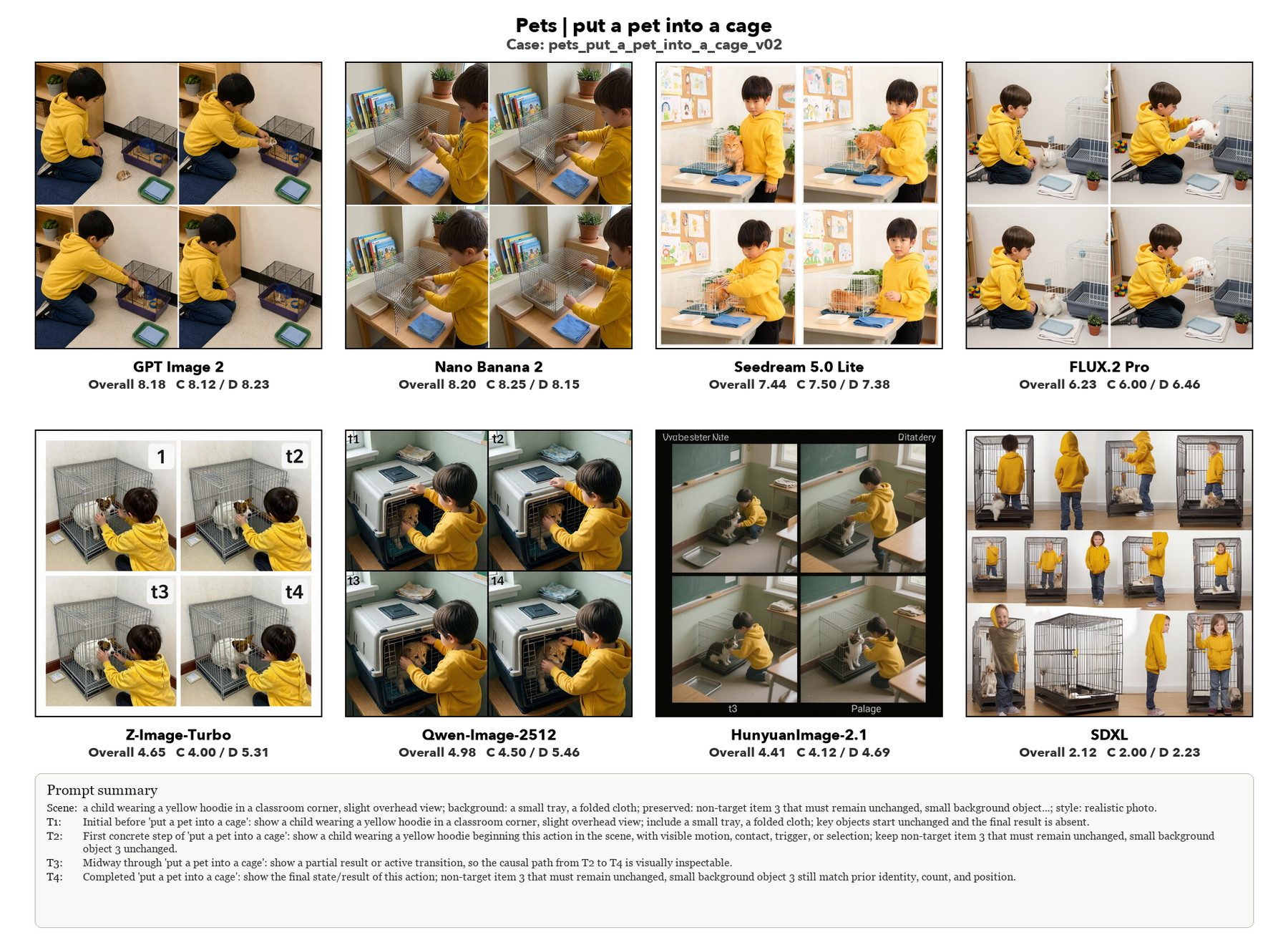}}\par
    \captionof{figure}{Example summary grids from the Pets category. Each grid compares the same prompt across eight image-generation models and reports the overall mean score below each generated motion sheet.}
    \label{fig:appendix-examples-pets}
\end{center}
\clearpage

\begin{center}
    \makebox[\textwidth][c]{\includegraphics[width=1.08\textwidth,height=0.435\textheight,keepaspectratio]{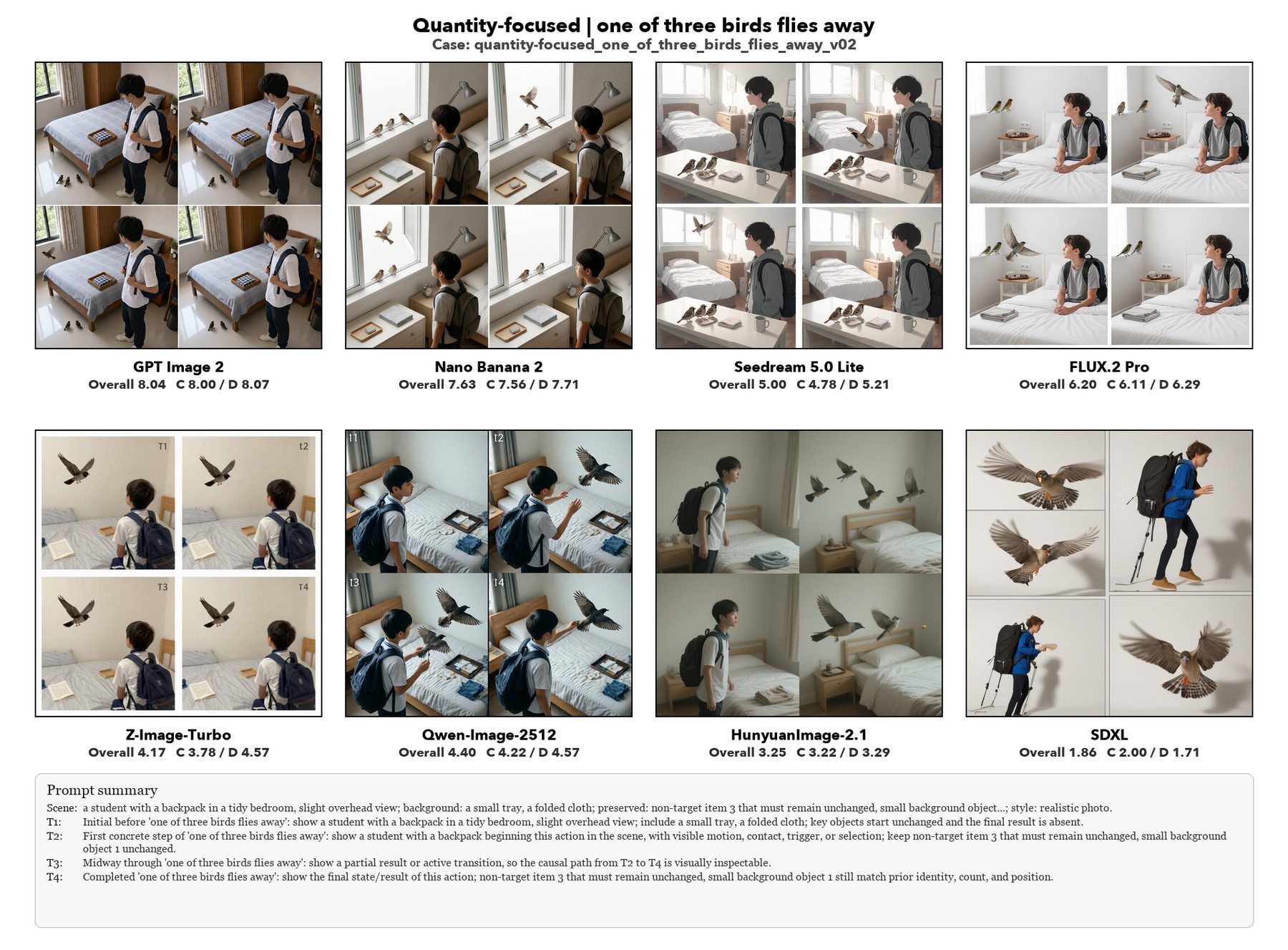}}\par
    \vspace{0.4em}
    \makebox[\textwidth][c]{\includegraphics[width=1.08\textwidth,height=0.435\textheight,keepaspectratio]{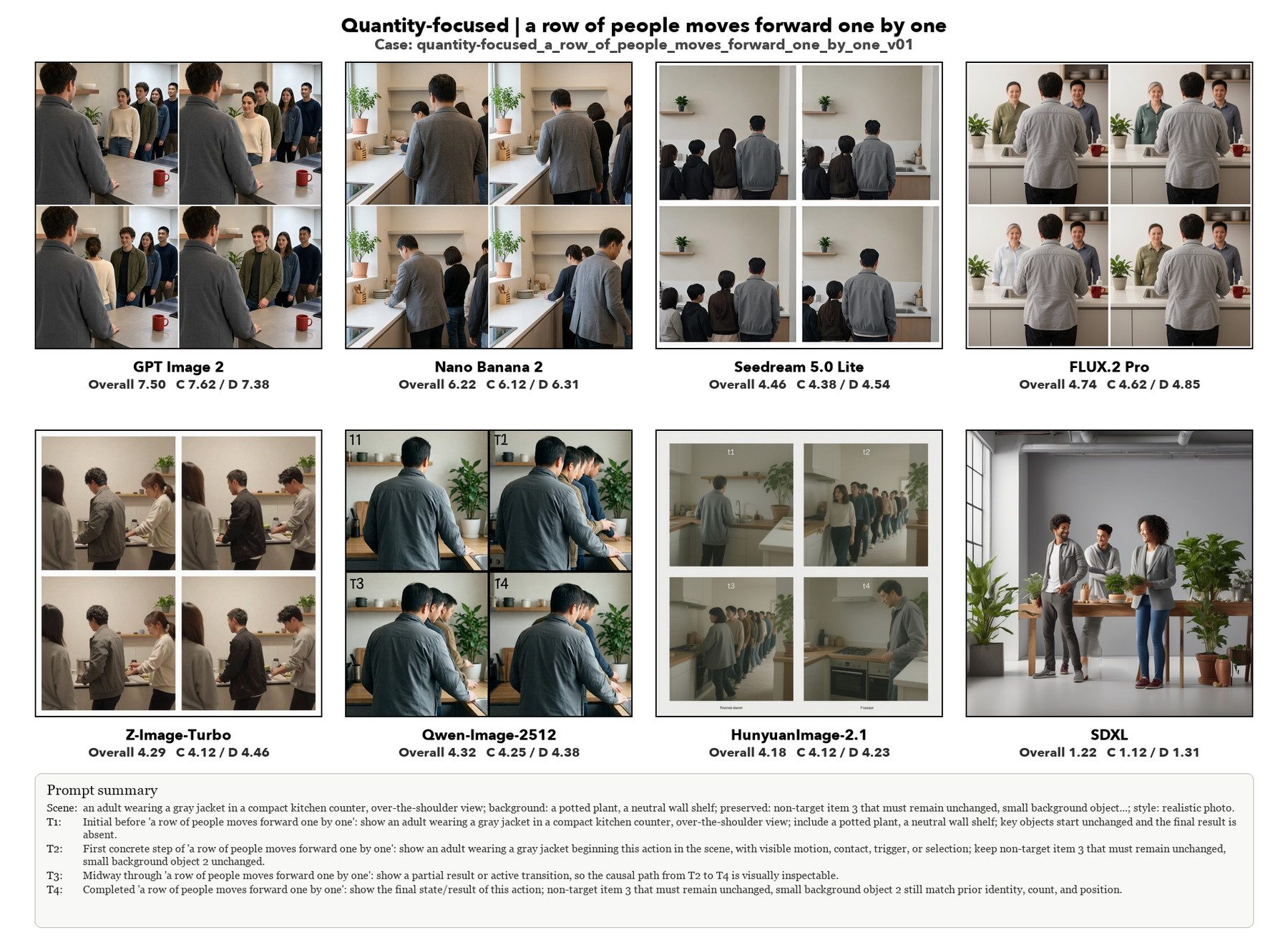}}\par
    \captionof{figure}{Example summary grids from the Quantity Focused category. Each grid compares the same prompt across eight image-generation models and reports the overall mean score below each generated motion sheet.}
    \label{fig:appendix-examples-quantity-focused}
\end{center}
\clearpage

\begin{center}
    \makebox[\textwidth][c]{\includegraphics[width=1.08\textwidth,height=0.435\textheight,keepaspectratio]{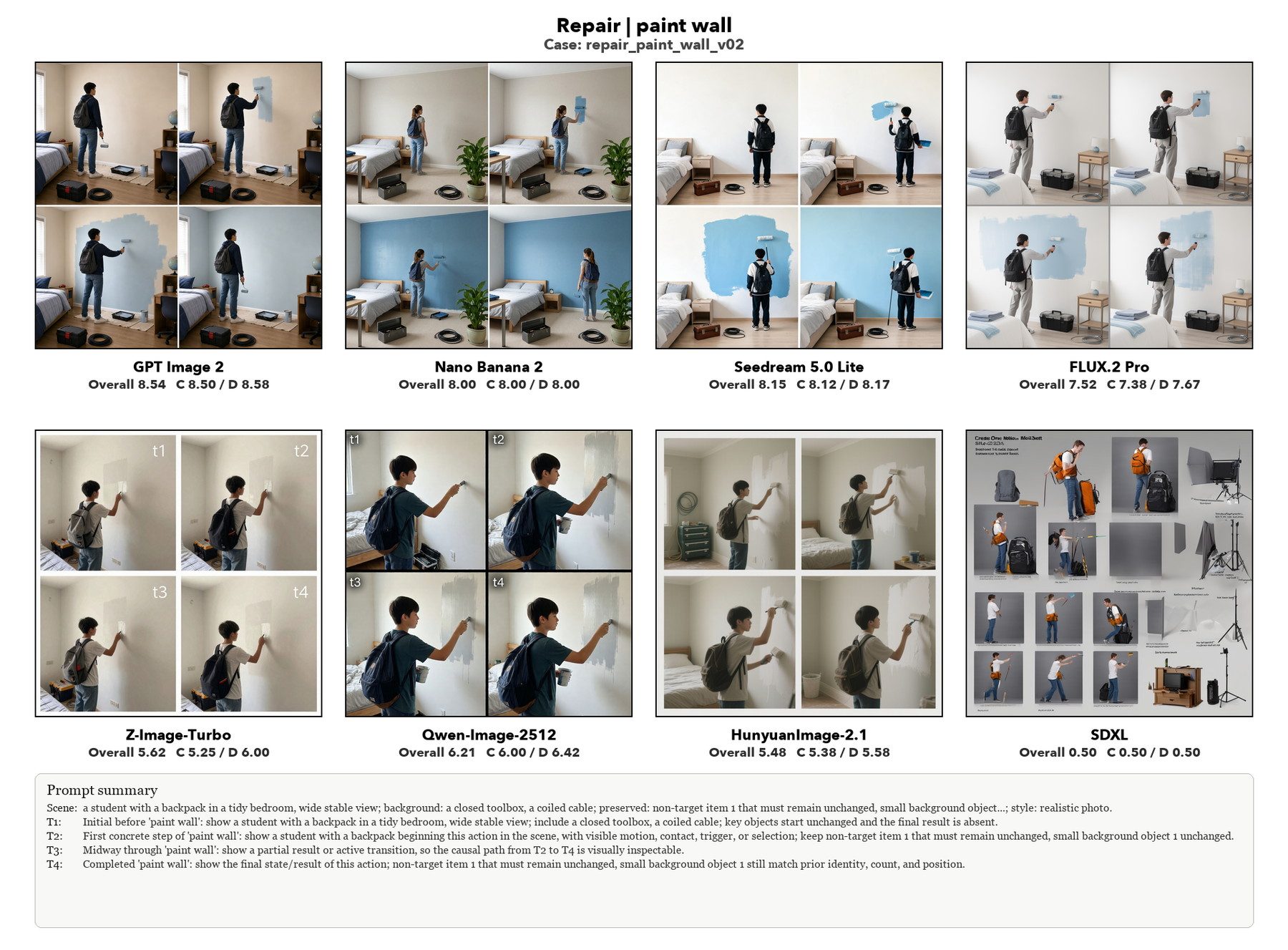}}\par
    \vspace{0.4em}
    \makebox[\textwidth][c]{\includegraphics[width=1.08\textwidth,height=0.435\textheight,keepaspectratio]{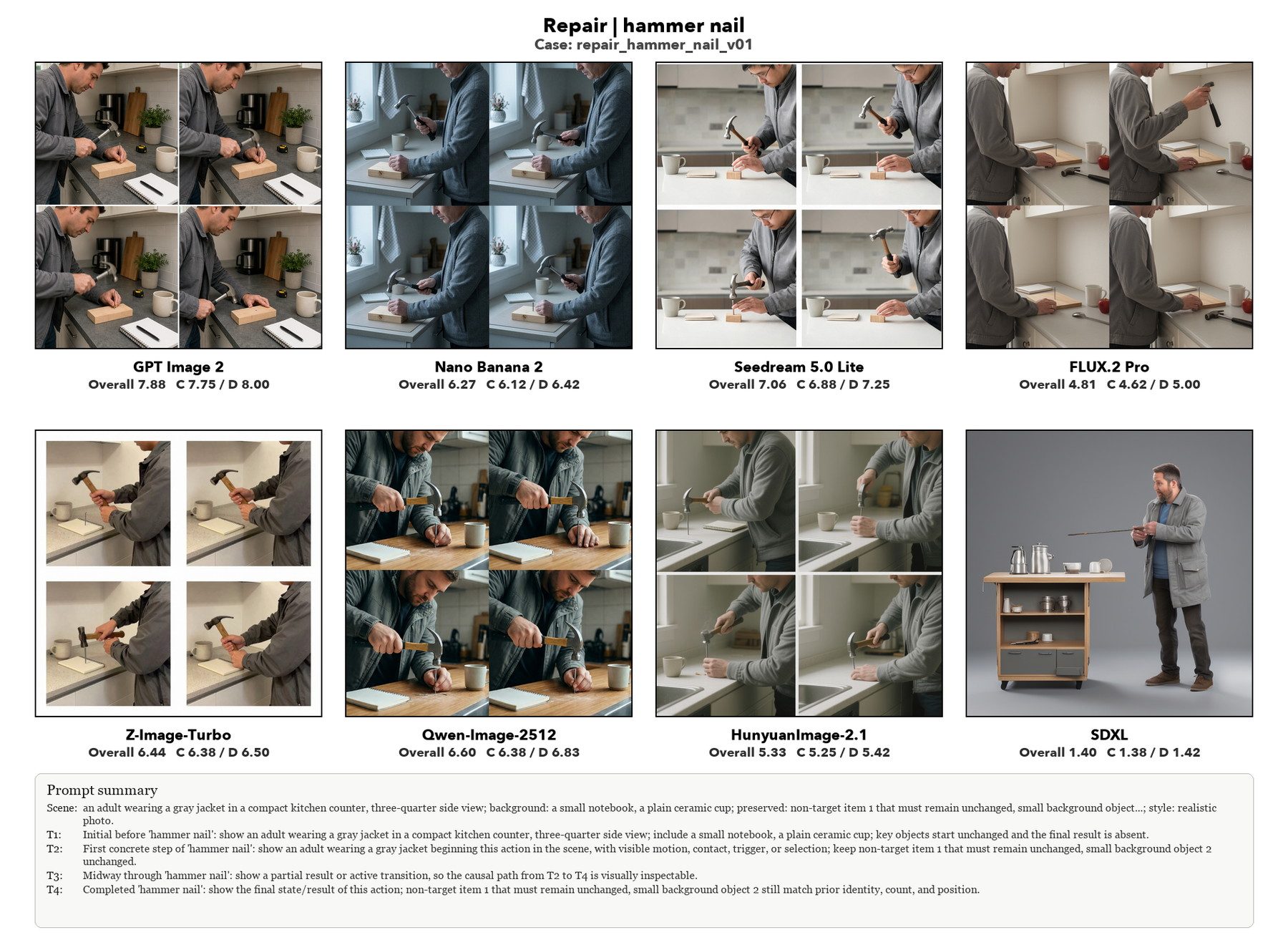}}\par
    \captionof{figure}{Example summary grids from the Repair category. Each grid compares the same prompt across eight image-generation models and reports the overall mean score below each generated motion sheet.}
    \label{fig:appendix-examples-repair}
\end{center}
\clearpage

\begin{center}
    \makebox[\textwidth][c]{\includegraphics[width=1.08\textwidth,height=0.435\textheight,keepaspectratio]{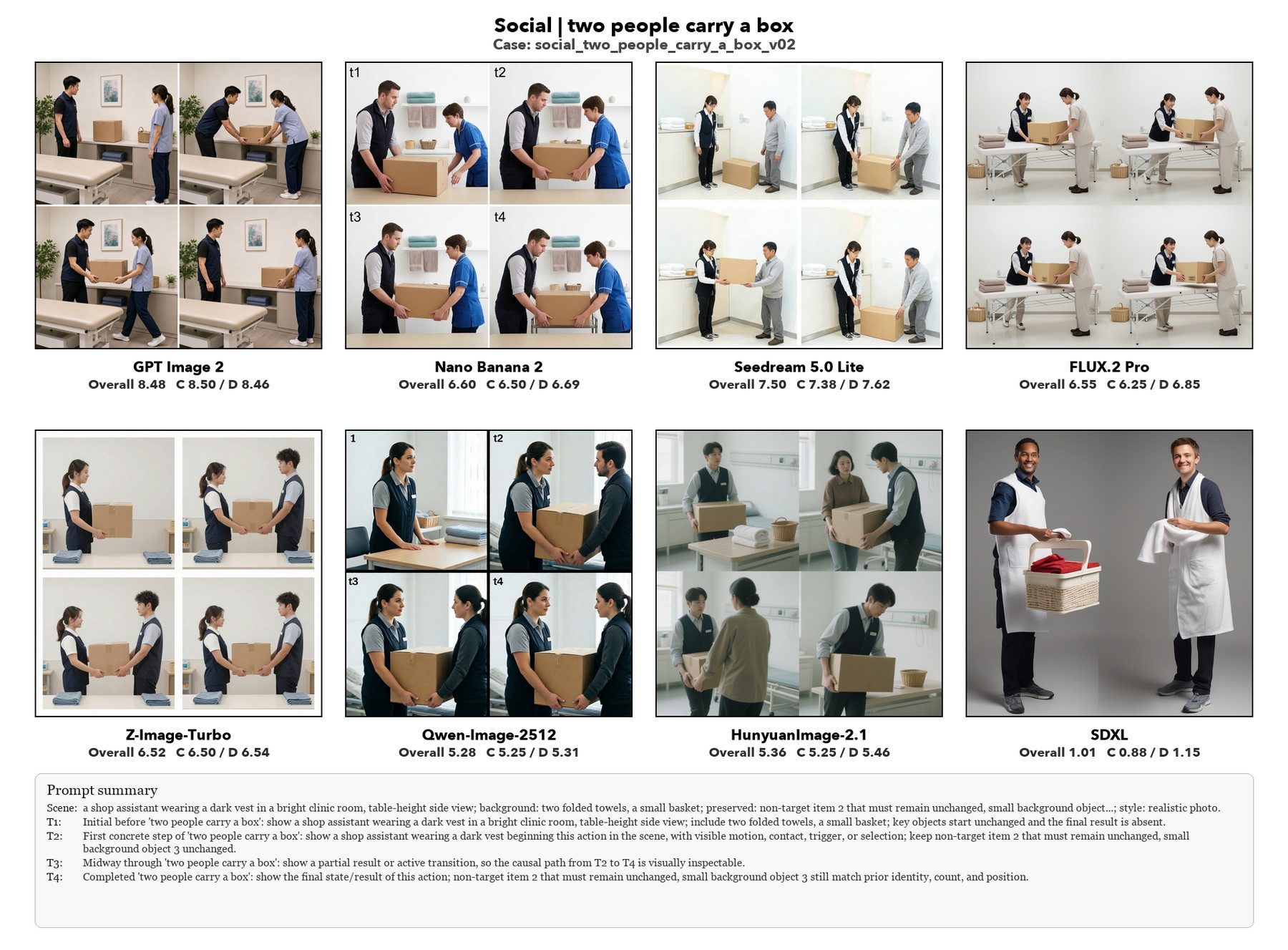}}\par
    \vspace{0.4em}
    \makebox[\textwidth][c]{\includegraphics[width=1.08\textwidth,height=0.435\textheight,keepaspectratio]{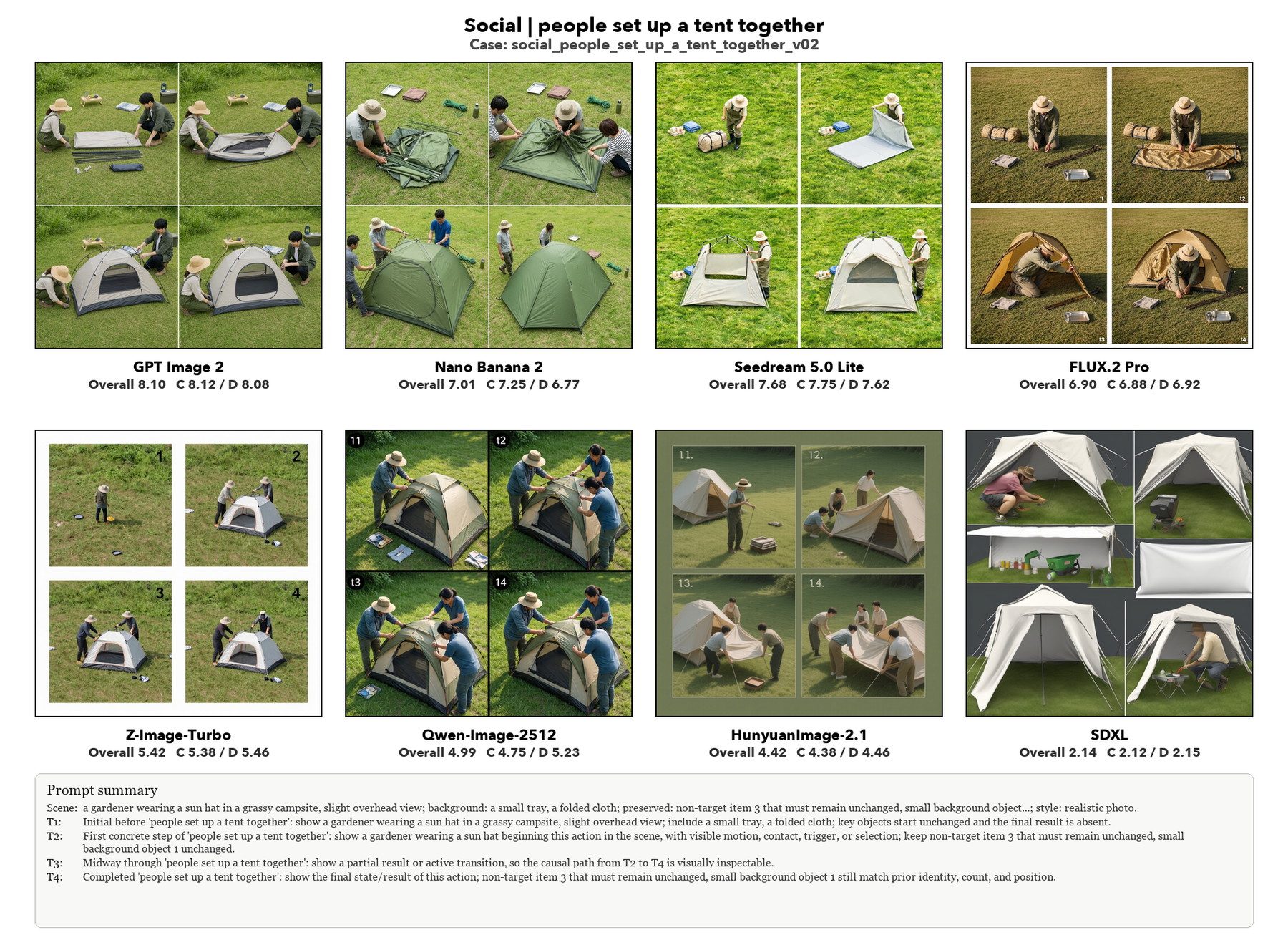}}\par
    \captionof{figure}{Example summary grids from the Social category. Each grid compares the same prompt across eight image-generation models and reports the overall mean score below each generated motion sheet.}
    \label{fig:appendix-examples-social}
\end{center}
\clearpage

\begin{center}
    \makebox[\textwidth][c]{\includegraphics[width=1.08\textwidth,height=0.435\textheight,keepaspectratio]{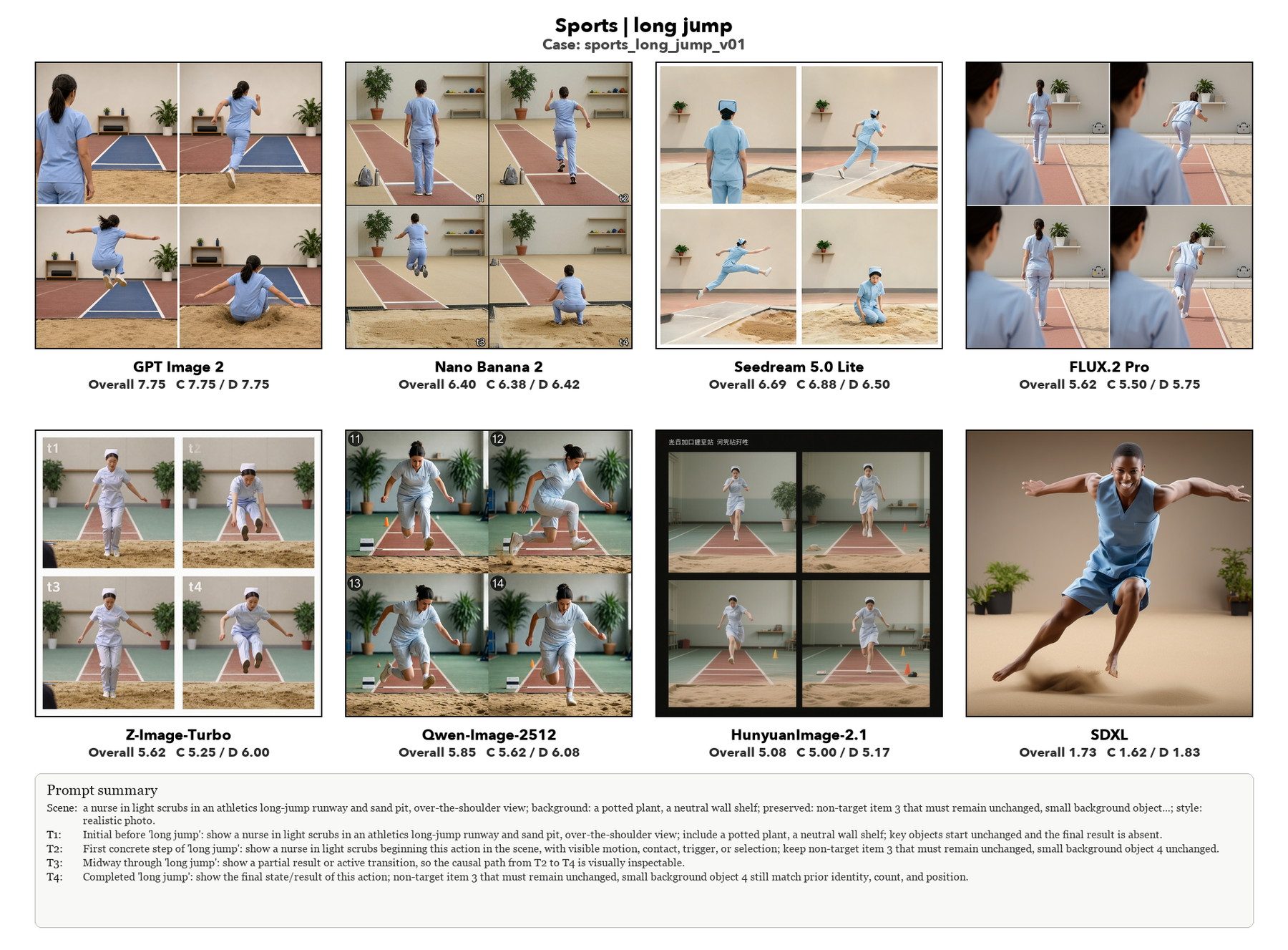}}\par
    \vspace{0.4em}
    \makebox[\textwidth][c]{\includegraphics[width=1.08\textwidth,height=0.435\textheight,keepaspectratio]{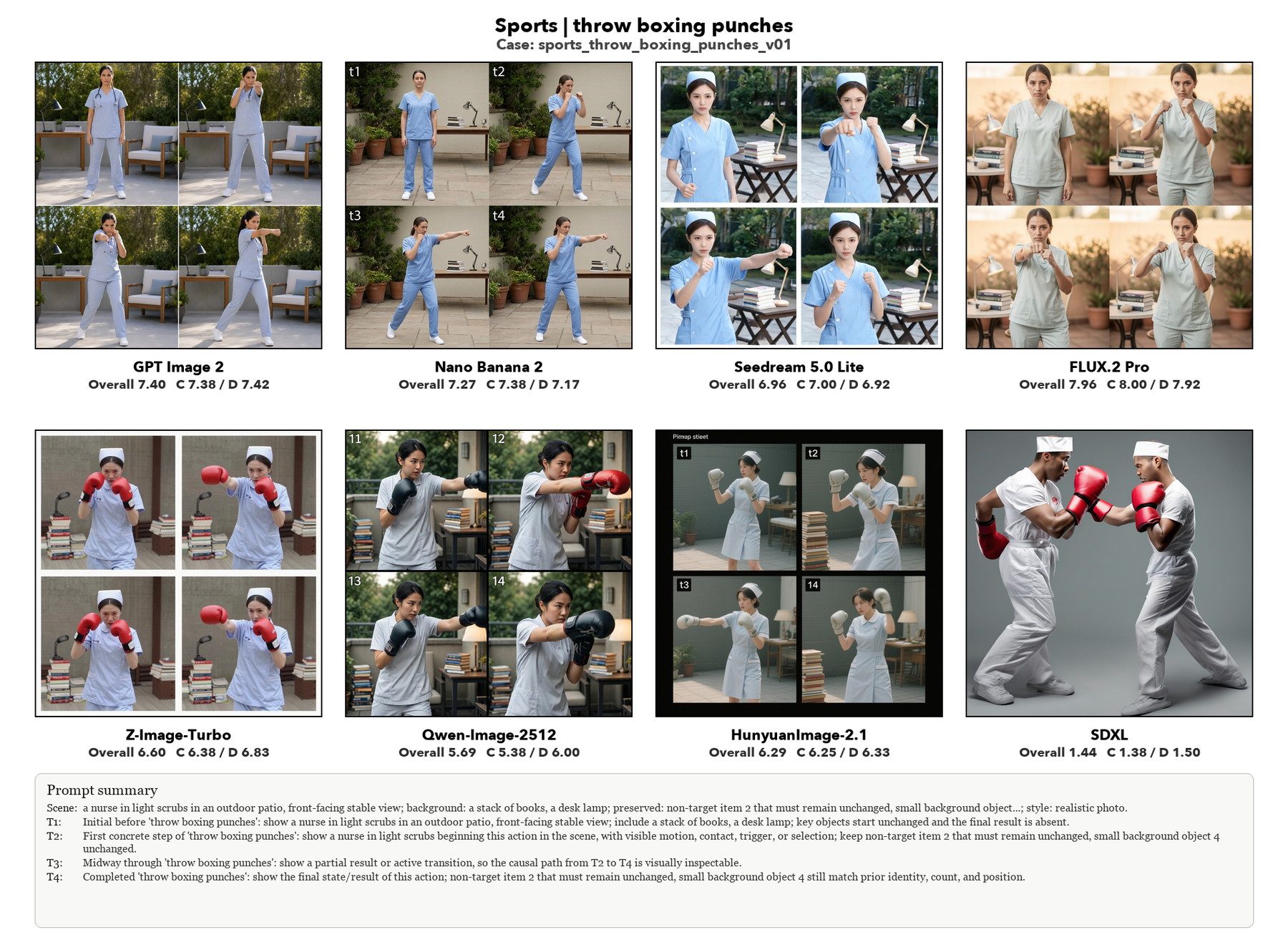}}\par
    \captionof{figure}{Example summary grids from the Sports category. Each grid compares the same prompt across eight image-generation models and reports the overall mean score below each generated motion sheet.}
    \label{fig:appendix-examples-sports}
\end{center}
\clearpage

\begin{center}
    \makebox[\textwidth][c]{\includegraphics[width=1.08\textwidth,height=0.435\textheight,keepaspectratio]{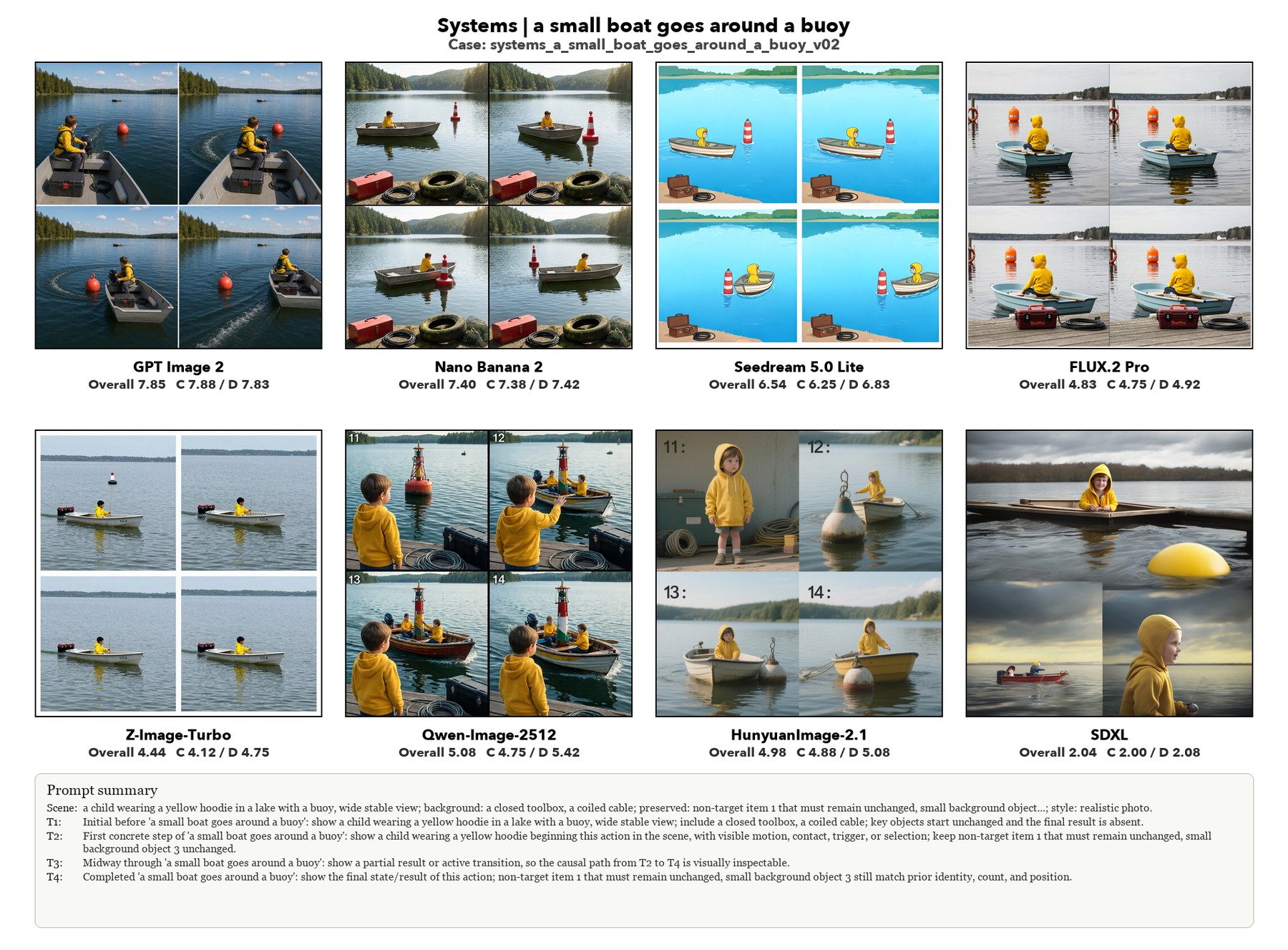}}\par
    \vspace{0.4em}
    \makebox[\textwidth][c]{\includegraphics[width=1.08\textwidth,height=0.435\textheight,keepaspectratio]{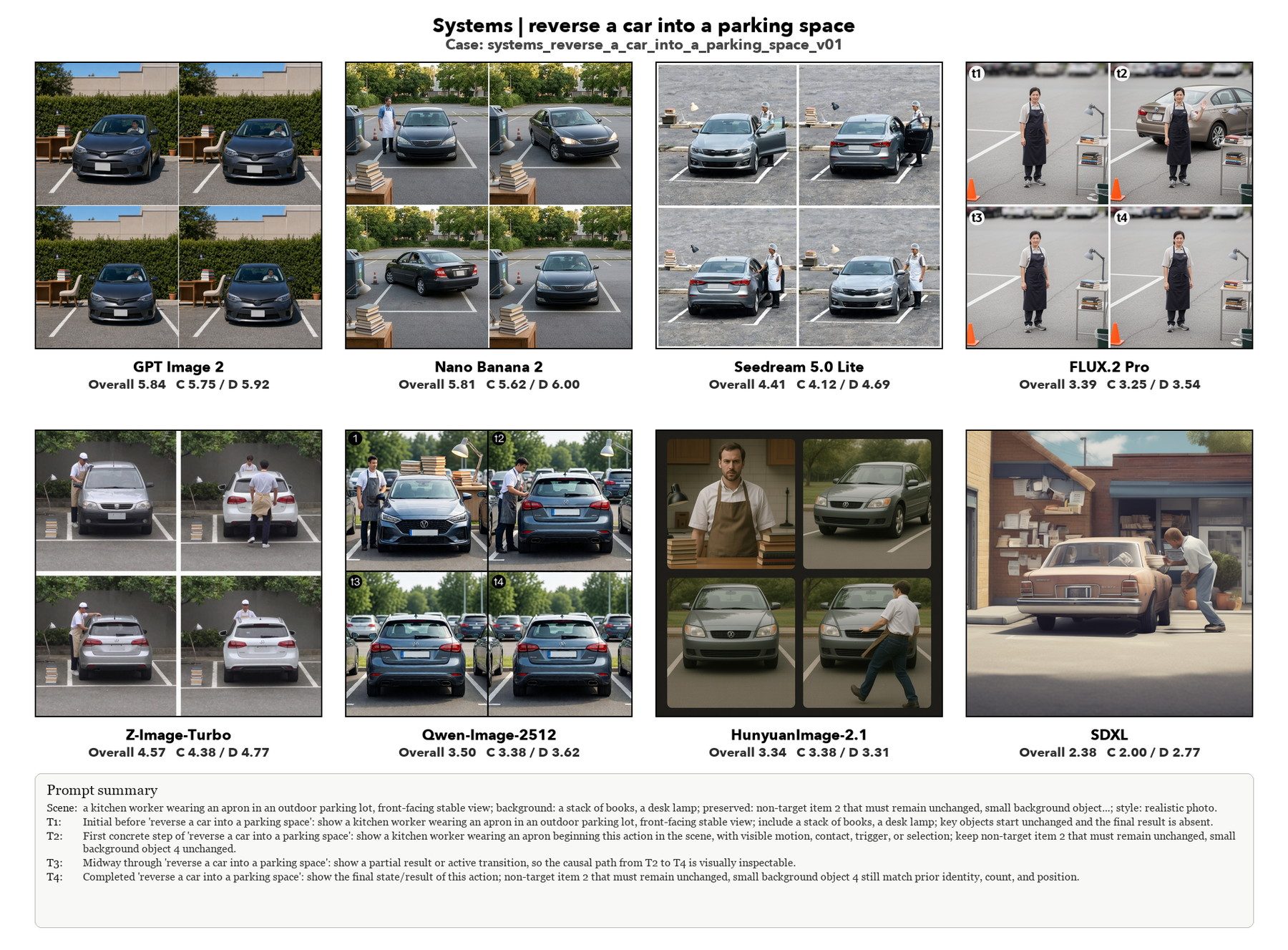}}\par
    \captionof{figure}{Example summary grids from the Systems category. Each grid compares the same prompt across eight image-generation models and reports the overall mean score below each generated motion sheet.}
    \label{fig:appendix-examples-systems}
\end{center}
\clearpage

\begin{center}
    \makebox[\textwidth][c]{\includegraphics[width=1.08\textwidth,height=0.435\textheight,keepaspectratio]{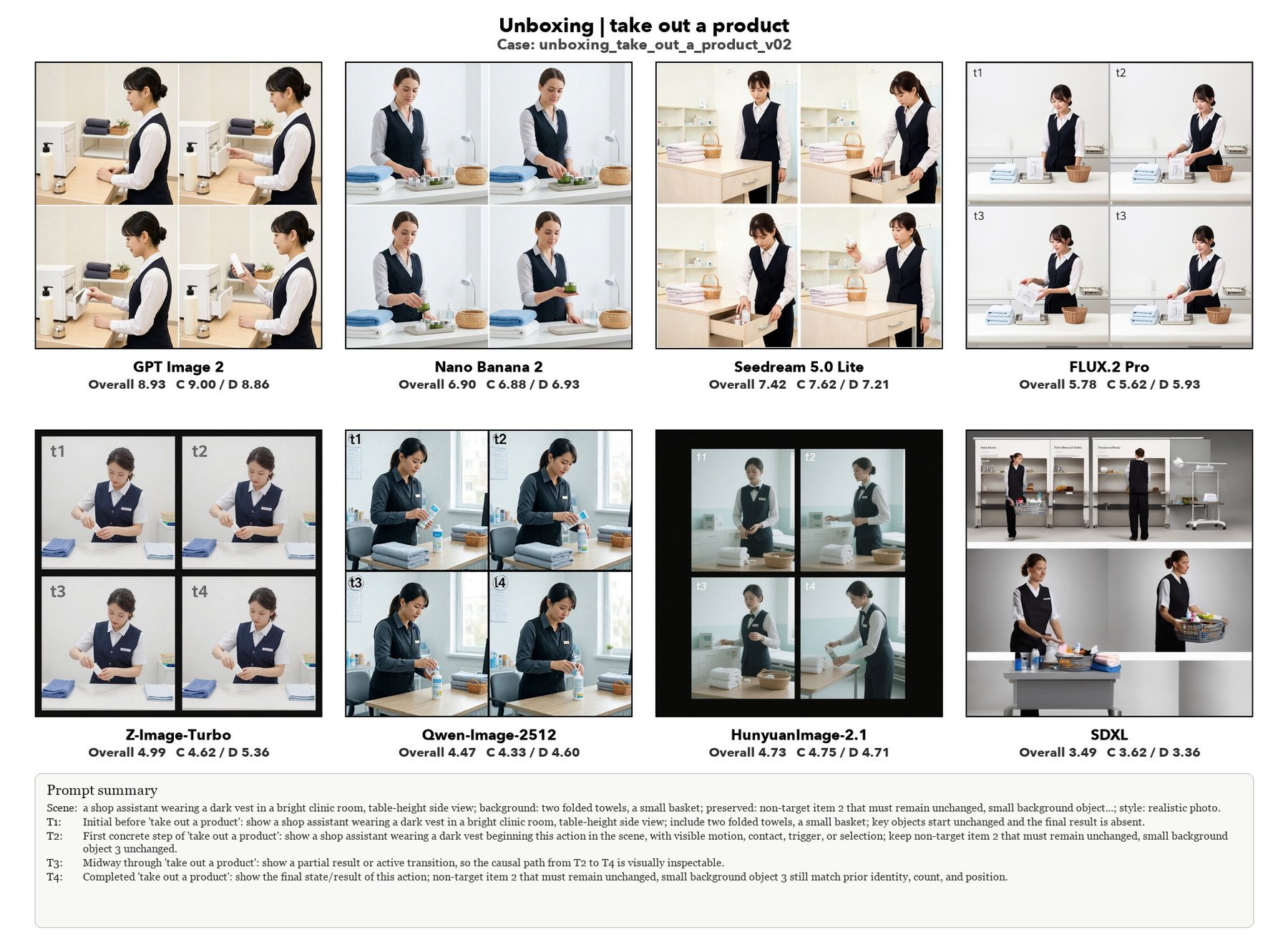}}\par
    \vspace{0.4em}
    \makebox[\textwidth][c]{\includegraphics[width=1.08\textwidth,height=0.435\textheight,keepaspectratio]{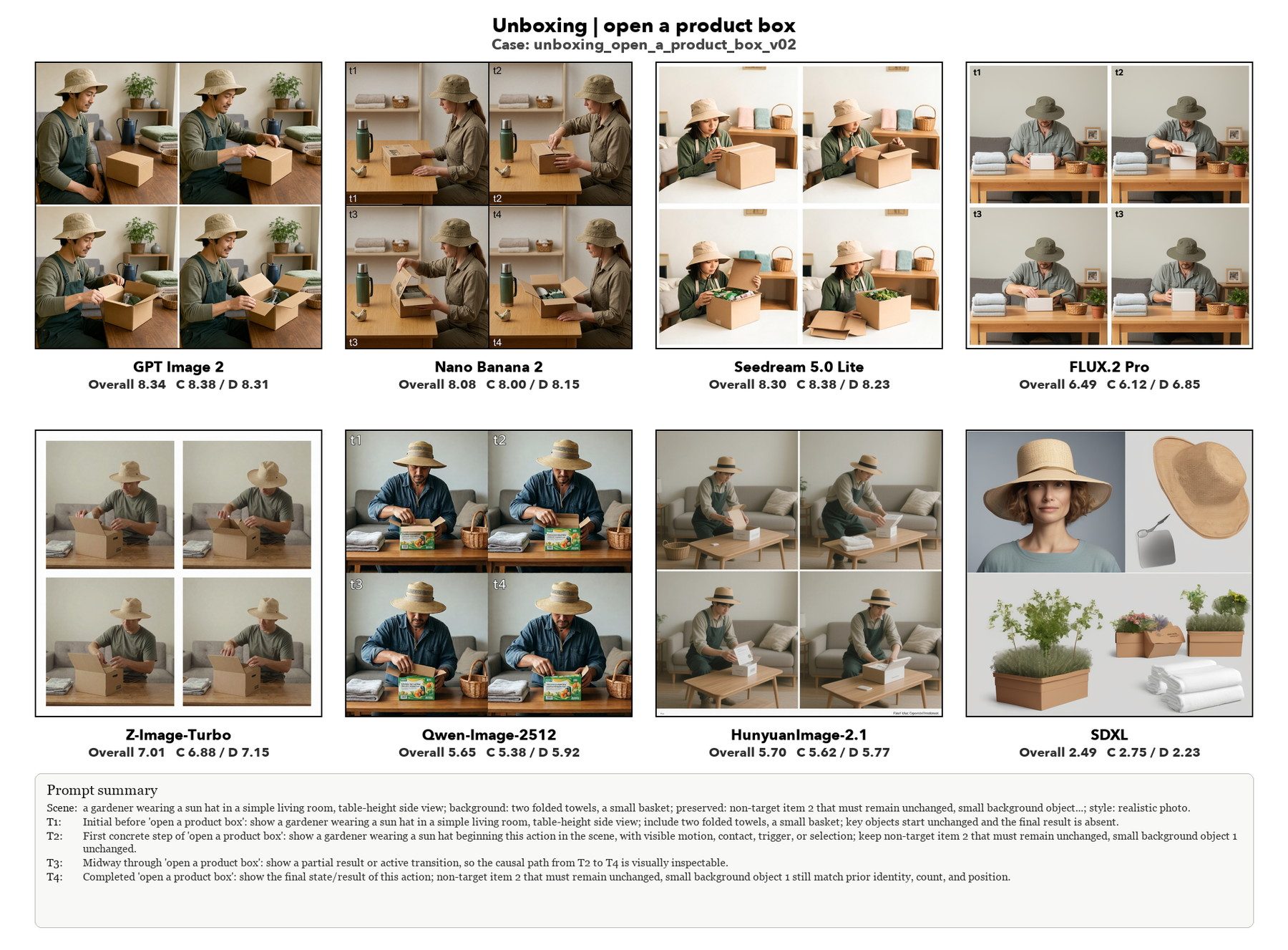}}\par
    \captionof{figure}{Example summary grids from the Unboxing category. Each grid compares the same prompt across eight image-generation models and reports the overall mean score below each generated motion sheet.}
    \label{fig:appendix-examples-unboxing}
\end{center}

\clearpage
\section{Reproducibility and Ethics}
\label{sec:appendix-reproducibility-ethics}

\textbf{Reproducibility.}
We will release the ImageTime case specifications, generation prompts, reference prompts, prompt-only and scaffold prompts, state predicates, forbidden-state annotations, GPT-5.5 judge prompts, scoring scripts, generated-output manifests, score JSON files, and CSV summaries where licensing terms permit. In the reported experiments, GPT-5.5 is used to score all images generated by all evaluated models. Each evaluated image is produced by a single generation request from the corresponding model; the score files do not include multi-turn dialogue outputs, iterative fixes, manually selected retries, or judge-guided regenerations. For each generated motion sheet, the scoring input contains the image, original prompt, structured process specification, capability rubric, diagnostic rubric, and conservative score-capping rules. The final analysis CSV records every GPT-5.5-produced C-score, D-score, confidence value, overall score, model name, setting, and case identifier used in the paper figures.

The generated image set and GPT-5.5 score files should be treated as the primary reproducible artifact. The released scripts support independent re-scoring, aggregation, and figure generation from these artifacts, and the structured case specifications allow future work to regenerate or extend the benchmark under comparable prompt and evaluation settings. Because hosted model behavior can change over time, we will record the judge model name, scoring date, prompt template, and raw structured judge response wherever possible; future re-scoring with a different judge model should be reported separately rather than mixed with the main GPT-5.5 scores.

\textbf{Ethics.}
ImageTime is designed as a diagnostic benchmark for generation capability, not as a tool for high-stakes decisions. The prompts and reference materials are synthetic, benchmark-oriented, or generated for authorized evaluation use. The dataset avoids tasks that require identifying private individuals, inferring sensitive attributes, or evaluating real people. GPT-5.5-assisted quality checks were used to remove or repair prompts with obvious factual or common-sense mismatches, such as actions placed in physically inappropriate scenes.

The benchmark can still encode the biases of its task design, model outputs, and GPT-5.5 judge. Some actions, settings, objects, and visual styles may be overrepresented, and GPT-5.5 can miss subtle physical errors or over-reward polished render quality. We therefore report diagnostic subscores and failure labels rather than only a single ranking. Generated images should be used as benchmark artifacts and should not be presented as human-created visual evidence.


\end{document}